\documentclass[review]{elsarticle}

\usepackage{lineno,hyperref}
\journal{Journal of Precision Agriculture}
\usepackage{graphicx}
\hypersetup{colorlinks=true,linkcolor=black}
\usepackage{subfig}
\usepackage{floatrow}
\usepackage{setspace}
\usepackage[labelsep=period]{caption}
\usepackage[british]{babel} 
\usepackage[hang]{footmisc}

\usepackage{multirow}
\usepackage[binary-units = true]{siunitx}
\DeclareSIUnit\px{px}

\usepackage{csquotes}
\usepackage{nameref}% references by name da keine nummerierung der Sections
\usepackage{tabu}
\usepackage{array} %for thickness of table line
\usepackage{mathtools}

\usepackage{pifont}%
\newcommand{\xmark}{\ding{55}}%

\def\figref#1{Fig.~\ref{#1}}
\def\tabref#1{Tab.~\ref{#1}}
\def\eqref#1{Eq.~(\ref{#1})}

\newcommand{\ical}[1]{{\mbox{\usefont{OT1}{pzc}{m}{it}{#1}}}}
\newcommand{\m}[1]{{\mbox{{\fontencoding{T1}\sffamily\slshape{#1\/}}}}}
\renewcommand{\d}[1]{{\mbox{\boldmath$#1$}}}

\begin{document}

\begin{frontmatter}

\title{Behind the leaves -- Estimation of occluded grapevine berries with conditional generative adversarial networks.}
\author[igg]{Jana Kierdorf\corref{mycorrespondingauthor}}
\ead{jkierdorf@uni-bonn.de}

\author[kob]{Immanuel Weber}
\ead{immanuel.weber@hs-koblenz.de}

\author[jki]{Anna Kicherer}
\ead{anna.kicherer@julius-kuehn.de}

\author[igg]{Laura Zabawa}
\ead{l.zabawa@igg.uni-bonn.de}

\author[igg]{Lukas Drees}
\ead{ldrees@uni-bonn.de}

\author[igg]{Ribana Roscher}
\ead{ribana.roscher@uni-bonn.de}

% \address[igg-rs]{Institute of Geodesy and Geoinformation, Remote Sensing Group, University of Bonn, Germany}
\address[igg]{Institute of Geodesy and Geoinformation, University of Bonn, Germany}
\address[kob]{Application Center for Machine Learning and Sensor Technology, University of Applied Sciences Koblenz, Germany}
% \address[igg-gd]{Institute of Geodesy and Geoinformation, Department of Geodesy, University of Bonn, Germany}
\address[jki]{Julius Kühn-Institut, Federal Research Centre of Cultivated Plants, Institute for
Grapevine Breeding Geilweilerhof, Germany}

\cortext[mycorrespondingauthor]{Corresponding author}

\begin{abstract}
The need for accurate yield estimates for viticulture is becoming more important due to increasing competition in the wine market worldwide. 
One of the most promising methods to estimate the harvest is berry counting, as it can be approached non-destructively, and its process can be automated. 
In this article, we present a method that addresses the challenge of occluded berries with leaves to obtain a more accurate estimate of the number of berries that will enable a better estimate of the harvest.
We use generative adversarial networks, a deep learning-based approach that generates a likely scenario behind the leaves exploiting learned patterns from images with non-occluded berries. 
Our experiments show that the estimate of the number of berries after applying our method is closer to the manually counted reference. 
In contrast to applying a factor to the berry count, our approach better adapts to local conditions by directly involving the appearance of the visible berries.
Furthermore, we show that our approach can identify which areas in the image should be changed by adding new berries without explicitly requiring information about hidden areas. 
\end{abstract}


\end{frontmatter}

% %%%%%%%%%%%%%%%%%%%%%%%%%%%%%%%%%%%%%%%%%%%%%%%%%%%%%%%%%%%%%%%%%%%%%%%%%%%%%%%%%%%%%%%%%%%%%%%%%%%%%%%%%%%%%%%%%%%%
% %%%%%%%%%%%%%%%%%%%%%%%%%%%%%%%%%%%%%%%%%%%%%%%%%%%%%%%%%%%%%%%%%%%%%%%%%%%%%%%%%%%%%%%%%%%%%%%%%%%%%%%%%%%%%%%%%%%%
% KAO: Sloppy spacing ensures non-overfull lines. Can be removed if this is not an issue.
% \sloppy

\section*{Introduction}
\label{sec:introduction}
%% WHY: First, answer the WHY question. Why is that relevant? Why should I be motivated to read the paper?
With increasing competition on the wine market worldwide, the need for accurate yield estimations has been getting more and more important for viticulture. 
The variation of yield over the years is mainly caused by the berry number per vine (90\%), while the remaining 10\% are caused by the average berry weight \cite{clingeleffer2001crop}, which is generally collected manually and averaged over many years.
Traditionally, yield estimations in viticulture can be done at three phenological timepoints by (1) counting the number of bunches 4-6 weeks after budburst, (2) counting the number of berries after fruit set \cite{may1972forecasting} or (3) destructively sampling vines or segments of vines close to harvest.
Considering that yield estimation can be more accurately and reliably determined as harvest approaches, a berry count is a promising option that can be approached non-destructively and whose process can be automated.

%% WHICH PROBLEM: Second, explain WHICH problem you are solving/address to solve.
Several papers show that machine learning-based methods for analyzing data from imaging sensors provide an objective and fast method for counting visible berries \cite{aquino2017new,roscher2014automated,nuske2014automated,kicherer2014image,diago2012grapevine,zabawa2020counting,Coviello20}, and thus for automated yield predictions in the field.
One of the main challenges in deriving berry counts from image data taken in the field is occlusions, which generally causes an underestimation of the number of berries and yield \cite{Zabawa2021}.
%% HOW & WHAT: Third, explain briefly how one can address/model/solve the problem and mention briefly what others/we before have done. Prepare the reader for your contribution that comes in the next section.
First, occlusions of berries by other berries make it difficult to distinguish or count individual berries. 
Therefore, approaches that perform a %semantic 
segmentation of regions of berries and regions without berries is not sufficient, and more advanced methods that recognize individual instances of berries must be applied \cite{zabawa2020counting}.
Second, occlusions by leaves play a major role in underestimating the number of berries.
Zabawa et al. \cite{Zabawa2021} perform leaf occlusion experiments over two years and show that the yield estimation is highly dependent on the number of visible berries. 
With vines defoliated (i.e., with manually removed leaves) at pea size, they report an average error of total yield estimation of 27\%, whereas Nuske \cite{nuske2014automated} observed average errors between 3\% to 11\% using images of entirely defoliated fruit zones.

In order to overcome the challenge of leaf occlusions, defoliation can be performed in the grape fruit zone, but this is immensely time-consuming and labor intensive.
Partial defoliation is carried out in viticulture, for example, for ventilation and rapid drying of the grape zone to avoid fungal infections of the grapes or yield and quality regulation \cite{diago2009early}. 
However, complete defoliation is not feasible on a large scale or may lead to negative effects such as increased sunburn on the berries \cite{feng2015influence} or generally have an undesirable impact on yield results.
Alternatively, machine-learning-based approaches can be used to obtain a more accurate estimation of the berry number.
Numerous approaches rely on information where occlusions are present, which is generally provided as a manual input \cite{bertalmio2003simultaneous,barnes2009patchmatch,iizuka2017globally,liu2018image,dekel2018sparse}. 
In contrast to this, two-step approaches first detect occlusions and then fill the corresponding regions with information according to the environment \cite{yan2019visualizing,ostyakov2018seigan}.

%% MAIN CONTRIBUTION & WHAT FOLLOWS FROM THAT: Explain your contribution in one paragraph. Start that paragraph with: The main contribution of this paper is a \dots  We achieve this by \dots This allows us to \dots See \figref{fig:motivation} for an example. (something like this)
This article addresses the challenge of occlusions caused by leaves by generating images that reveal a likely situation behind the leaves, exploiting learned patterns from a carefully designed dataset. 
The generated images can then be used to count berries in a post-processing step.
Our approach generates potential berries behind leaves based on RGB information obtained by visible light imaging, as this is an efficient, cheap and non-harmful approach in contrast to data from material-penetrating sensors.
In order to train our machine-learning method, we use aligned image pairs showing plants with leaves and the same plants after defoliation.
In detail, we model this problem as a domain-transfer task and regard the aligned images containing occluded berries as one domain and images with revealed berries as a second domain.
We resort to methods like Pix2Pix~\cite{isola2017image}, that uses a conditional generative adversarial network (cGAN)~\cite{mirza2014conditional} and can learn the described domain-transfer.
In contrast to other works, we present a one-step approach that is end-to-end trainable, meaning the positions of the occlusions are identified, and patterns that need to be filled are learned simultaneously.
Through the experience the model gains during training, it learns patterns such as grape instances with their appearing shapes, their environment, and where they occur in the image. 
This knowledge is exploited during the generation step, in which the learned domain-transfer model is applied to images of vines that have not been defoliated to obtain a high-probability and realistic impression of the scene behind the leaves.
In order to obtain a berry count, the generated images are further processed with the berry counting algorithm of Zabawa et al.~\cite{zabawa2020counting}.
In this way, we provide a more accurate count of grape berries since both visible berries and berries potentially occluded by leaves are taken into account.

A major challenge for training is that there is no large dataset of aligned natural images that includes both images with occluded berries and images with berries exposed by defoliation.
In addition, in our case, the spatial alignment between the image pairs is not accurate enough since defoliation leads to a resulting movement of branches, grape bunches, and other objects in the non-occluded domain patches.
As a result, the natural data is not sufficient to train a model that matches our requirements of a reliable model. 
Due to this, we propose the use of a synthetically generated dataset that contains paired data of both domains.
Our main contributions of this paper are:
\begin{itemize}
    \item The true scenario behind the leaves without defoliation is unlikely to be identified. Therefore, our approach estimates a likely scenario based on visible information in the image, especially of the surroundings of the occlusion, and learned patterns during the training process.
    
    \item We present a one-step approach, which can implicitly identify which image areas contain visible berries and which areas are occluded without supervision regarding occluded and non-occluded areas. This differs from approaches such as inpainting \cite{bertalmio2003simultaneous,barnes2009patchmatch,iizuka2017globally,liu2018image,dekel2018sparse}, in which the occluded areas must be known a priori.
    
    \item In addition to the acquired images, we use so-called berry masks obtained by the approach presented in \cite{zabawa2020counting}, which uses semantic segmentation to indicate in the image which pixels belong to \texttt{berry}, \texttt{berry-edge} and \texttt{background}. During training, this leads to a more stable and easier optimization process. During testing, the berry mask is only needed for the input image since our GAN-based method simultaneously generates the berry mask in which the berries are counted, in addition to the visually generated image.
    
    \item Since a direct comparison of the true scenario behind the leaves and our generated scenario is not appropriate using standard evaluation methods such as a pixel-by-pixel comparison, we perform a comprehensive evaluation using alternative evaluation metrics, such as generation maps and correlation, that assesses the performance of our approach. 
    
    \item We show that the application of our approach minimizes the offset compared to the manual reference berry count and the variance, which is not achieved by applying a factor.
    
    \item We create various synthetic datasets and show that our approach trained on synthetic data also works on natural data.
\end{itemize}

The paper is structured as follows:
After surveying related works, we start by introducing our domain-transfer framework and describe the different components, such as the conditional generative adversarial network, that are used in our approach. 
%In Sec. \enquote{\nameref{sec:data}}, %\secref{sec:data}, 
We explain the data acquisition and post-processing of the natural and synthetic datasets we use in our work. 
We explain the evaluation metrics we use %, in Sec. \enquote{\nameref{sec:evalMetrics}} %\secref{sec:evalMetrics}
and then describe our experiments in which we analyse the generation quality of different synthetic input data, %(Sec. \hyperref[sec:Experiment1]{\enquote{Experiment 1}}), %\secref{sec:Experiment1}),
compare generated results with real results in the occluded %(Sec. \hyperref[sec:Experiment2]{\enquote{Experiment 2}}) %(Exp.2, \secref{sec:Experiment2}) 
as well as the non-occluded domain %(Sec. \hyperref[sec:Experiment3]{\enquote{Experiment 3}}) %(Exp.3, \secref{sec:Experiment3}) 
and analyze the berry counting based on the generated results. %(Sec. \hyperref[sec:Experiment4]{\enquote{Experiment 4}}). %(Exp.4, \secref{sec:Experiment4}). 
Finally, we investigate the application of the synthetically learned models to natural data. %(Sec. \hyperref[sec:Experiment5]{\enquote{Experiment 5}}). %(Exp.5, \secref{sec:Experiment5}). 
We end our paper with the conclusion and future directions.

% %%%%%%%%%%%%%%%%%%%%%%%%%%%%%%%%%%%%%%%%%%%%%%%%%%%%%%%%%%%%%%%%%%%%%%%%%%%%%%%%%%%%%%%%%%%%%%%%%%%%%%%%%%%%%%%%%%%%
% %%%%%%%%%%%%%%%%%%%%%%%%%%%%%%%%%%%%%%%%%%%%%%%%%%%%%%%%%%%%%%%%%%%%%%%%%%%%%%%%%%%%%%%%%%%%%%%%%%%%%%%%%%%%%%%%%%%%

\section*{Related Work}

\paragraph{Yield estimation and counting.}
Since an accurate yield estimation is one of the major needs in viticulture, especially on a large scale, there is a strong demand for objective, fast, and non-destructive methods for yield forecasts in the field.
For many plants, including grapevines, the derivation of phenotypic traits is essential for estimating future yields.
Besides 3D-reconstruction \cite{scholer2015automated,mack2017high,mack2018semantic}, 2D-image processing is also a widely used method \cite{hacking2019investigating} for the derivation of such traits. % Volumenschätzung von vinegrapes
For vine, one plant trait that strongly correlates with yield is the number of bearing fruits, that means the amount of berries. %, die eine Pflanze trägt. 
This correlation is underlined by the study of Clingeleffer et al. \cite{clingeleffer2001crop} in which it is shown that the variation of grapevine yield over the years is mainly caused by the berry number per vine (90\%).%, while the remaining 10\% are caused by the average berry weight.

The task of object counting can be divided into two main approaches: 1) regression \cite{lempitsky2010learning,xie2018microscopy,paul2017count,arteta2016counting} which directly quantifies the number of objects for a given input, and 2) detection and instance segmentation approaches which identify objects as an intermediate step for counting \cite{nyarko2018nearest,nuske2014automated}.
Detection approaches in viticulture are presented, for example, by Roscher et al. \cite{roscher2014automated}, Nuske et al. \cite{Nuske11}, and Nyarko et al. \cite{Nyarko18}, who define berries as geometric objects such as circles or convex surfaces and determine them by image analysis procedures such as Hough-transform.
Recent state-of-the-art approaches, especially segmentation \cite{he2017mask}, are mostly based on neural networks.  
One of the earliest works combining grapevine data and neural network analysis was Aquino et al. \cite{aquino2017new}. 
They detect grapes using connected components and determine key features based on them, which are fed as annotations into a three-layer neural network to estimate yield. 
In another work, Aquino et al. \cite{aquino2018automated} deal with counting individual berries, which are first classified into berry candidates using pixel classification and morphological operators. 
Afterward, a neural network classifies the results again and filters out the false positives.

The two studies by Zabawa et al. \cite{zabawa2019detection,zabawa2020counting} serve as the basis for our paper. 
In \cite{zabawa2019detection}, Zabawa et al. use a neural network which performs a semantic segmentation with the classes \texttt{berry}, \texttt{berry-edge} and \texttt{background}, which enables the identification of single berry instances. The masks generated in that work serve as input for our approach. 
The paper \cite{zabawa2020counting} based on \cite{zabawa2019detection} extends identification to counting berries by discarding the class edge and counting the berry components with a connected component algorithm. We also use the counting procedure applied in that work for the analyses of our experiments.

\paragraph{Given prior information about regions to be transferred.}

A significant problem in fruit yield estimation is the overlapping of the interesting fruit regions by other objects, like in our case, the leaves.
Several works are already addressing the issue of data with occluded objects or gaps within the data, where actual values are missing, which is typically indicated by special values like, e.g., not-a-number. 
The methodologies can be divided into two areas: 1) there is prior information available about where the covered positions are, and 2) there is no prior information.
In actual data gaps, where the gap positions can be easily identified a priori, data imputation approaches can be used to complete data.
This imputation is especially important in machine learning since machine learning models generally require complete numerical data. The imputation can be performed using constant values like a fixed constant, mean, median or k-nearest neighbor imputation \cite{batista2002study} or calculated using a random number like the empirical distribution of the feature under consideration \cite{enders2001primer,von2012maximum,rubin2004multiple,rubin1996multiple}. 
Also possible are multivariate imputations, which additionally measures the uncertainty of the missing values \cite{van1999flexible,kim2009unified,robins2000inference}. 
Data imputation is also possible using deep learning. 
Lee et al. \cite{lee2019collagan}, for example, introduce CollaGAN in which they convert the image imputation problem to a multi-domain image-to-image translation task.

In case there are no data gaps, but the image areas that are occluded or need to be changed are known, inpainting is a commonly used method. 
The main objective is to generate visually and semantically plausible appearances for the occluded regions to fit in the image. 
Conventional inpainting methods \cite{bertalmio2003simultaneous,barnes2009patchmatch} work by filling occluded pixels with patches of the image based on low level features like SIFT descriptors \cite{lowe2004distinctive}. 
The results of these methods do not look realistic if the areas to be filled are near foreground objects or the structure is too complex. 
An alternative is deep learning methods that learn a direct end-to-end mapping from masked images to filled output images. 
Particularly realistic results can be generated using Generative Adversarial Networks (GANs) \cite{iizuka2017globally,liu2018image,dekel2018sparse}. 
For example, Yu et al. \cite{yu2018generative} deal with generative image inpainting using contextual attention. 
They stack generative networks to ensure further the color and texture consistence of generated regions with surroundings. 
Their approach is based on rectangular masks, which do not generalize well to free-form masks.
This task is solved by Yu et al. \cite{yu2019free} one year later by using guidance with gated convolution to complete images with free-form masks. 
Further work introduces mask-specific inpainting that fills in pixel values at image locations defined by masks. Xiong et al. \cite{xiong2019foreground} learn a mask of the partially masked object from the unmasked region. 
Based on the mask, they learn the edge of the object, which they subsequently use to generate the non-occluded image in combination with the occluded input image.

\paragraph{No prior information about regions to be transferred.}

Methods that do not involve any prior knowledge about gaps and occluded areas can be divided into two-step and one-step approaches. Two-step approaches first determine the occluded areas, which then are used, for example, as a mask to inpaint the occluded areas. 
Examples are provided by Yan et al. \cite{yan2019visualizing}, which visualize the occluded parts by determining a binary mask of the visible object using a segmentation model and then creating a reconstructed mask using a generator.
The resulting mask is fed into coupled discriminators together with a 3D-model pool in order to decide if the generated mask is real or generated compared to the masks in the model pool.
Ostyakov et al. \cite{ostyakov2018seigan} train an adversarial architecture called SEIGAN to first segment a mask of the interesting object, then paste the segmented region into a new image and lastly fill the masked part of the original image by inpainting.
Similar to our approach, SeGAN introduced by Ehsani et al. \cite{ehsani2018segan} uses a combination of a convolutional neural network and a cGAN \cite{mirza2014conditional,isola2017image} to first predict a mask of the occluded region and, based on this, generate a non-occluded output.

%In contrast to previous works, we introduce a one-step approach. We implicitly specify a region in the image using a mask that should not change, while all other image regions are potentially changeable. Since we do not further constrain whether and where exactly the region outside the mask should change, we call it uninformed inpainting. We compute the inpainted output and the corresponding mask in one step by applying a cGAN.

% %%%%%%%%%%%%%%%%%%%%%%%%%%%%%%%%%%%%%%%%%%%%%%%%%%%%%%%%%%%%%%%%%%%%%%%%%%%%%%%%%%%%%%%%%%%%%%%%%%%%%%%%%%%%%%%%%%%%
% %%%%%%%%%%%%%%%%%%%%%%%%%%%%%%%%%%%%%%%%%%%%%%%%%%%%%%%%%%%%%%%%%%%%%%%%%%%%%%%%%%%%%%%%%%%%%%%%%%%%%%%%%%%%%%%%%%%%

\section*{Framework}

In our work, we regard the revealing of the occluded berries as a transfer between two image domains.
We first detail this and show how we model this transfer for our data.
Then we will lay out the cGAN and the framework we use for this task.
Finally, we show how we train this network.

\subsection*{Domain-transfer framework}
\label{sec:domain_transfer}

\begin{figure}[th!]
	\centering
    \includegraphics[
    width=\textwidth]{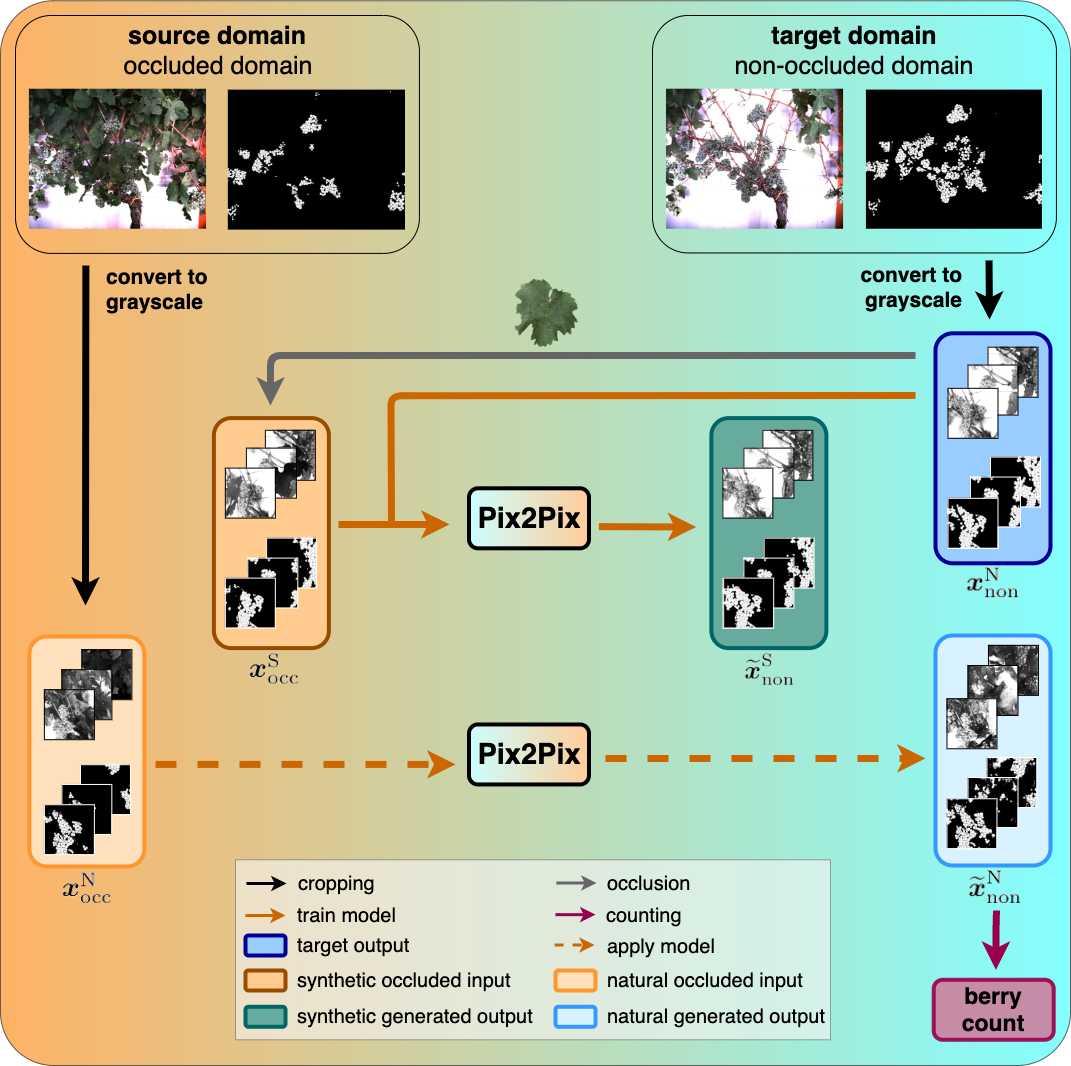}
	\caption{Domain-transfer framework. We transfer images from the source domain with occluded berries to the target domain with revealed berries using the Pix2Pix cGAN.
	We train and test the model on synthetic data and subsequently apply it to natural data. Finally, a berry counting is performed on the generated outputs. Further evaluation steps will be performed in our experiments.}
	\label{fig:framework}
\end{figure}

On a high level, the task of revealing the occluded berries can be described as generating a new impression of an existing image.
We model this generative task as a transfer of an existing image from one domain, the source domain, to another domain, the target domain.
In our work, we regard images where berries are occluded by various objects as the source domain and call it \emph{occluded domain}.
Accordingly, our target domain contains images of defoliated plants, and we call it \emph{non-occluded domain}.
Therefore by performing this domain-transfer, we aim to reveal hidden berries.
Samples of both domains are shown at the top of \figref{fig:framework}.

This task can typically be learned by a cGAN, like Pix2Pix in our case.
We train this network using aligned pairs of images from the occluded domain and the non-occluded domain and indicate them with $\d{x}_{\text{occ}}$ and $\d{x}_{\text{non}}$ respectively.
The first ones are used as the network input and the latter ones, being the desired output, as the training target. 
Due to computational limitations, we use cropped patches from the original data and convert them to grayscale to develop an efficient approach that is independent of the berry color. 
In practice, we accompany the images of each domain with a corresponding semantic mask, that indicates per image pixel the content based on the classes \texttt{berry}, \texttt{berry-edge} and \texttt{background}. 
This mask supports the discriminability of relevant information like the berries from the surrounding information in the image and the generation of separated berries, supporting the later counting step.
After training, we use the cGAN to generate images, $\widetilde{\d{x}}_{\text{non}}$, that we further process with a berry counting method.

Since we only have limited amounts of data available for training and testing, we resort to a dataset consisting of synthetic images for the occluded domain and natural images for the non-occluded domain that we describe in detail in Sec. \enquote{\nameref{sec:data}}. %\secref{sec:data}.
In addition, we test our trained model on fully natural data to analyze the generalizability of the model.
For the training set, the non-occluded domain contains natural images, whereas the images from the occluded domain are derived from the former domain, where berries are artificially occluded with leave templates.
To differentiate the different datasets of images, we further qualify the natural images with index $\text{N}$ and the synthetic images with index $\text{S}$, which results in the two occluded domain groups: $\d{x}^\text{N}_{\text{occ}}$ and $\d{x}^\text{S}_{\text{occ}}$. 
The generated images are accordingly indicated by $\widetilde{\d{x}}^\text{N}_{\text{non}}$ and $\widetilde{\d{x}}^\text{S}_{\text{non}}$.
We therefore train the model with input images $\d{x}^\text{S}_{\text{occ}}$ and use $\d{x}^\text{N}_{\text{non}}$ as target images. 
Finally, we apply the model on natural images $\d{x}^\text{N}_{\text{occ}}$ and compute the berry counts of the generated output images, $\widetilde{\d{x}}^\text{N}_{\text{non}}$.

%%%%%%%%%%%%%%%%%%%%%%%%%%%%%%%%%%%%%%%%%%%%%%%%%%%%%%%%%%%%%%%%%%%%%%%%%%%%%%%%%%%%%%%%%%%%%%%%%%%%%%%%%%%

\subsection*{Conditional Generative Adversarial Networks}
\label{sec:GAN}

\begin{figure}[t]
	\centering
    \includegraphics[%trim=0 100 0 0, clip,
    width=\textwidth]{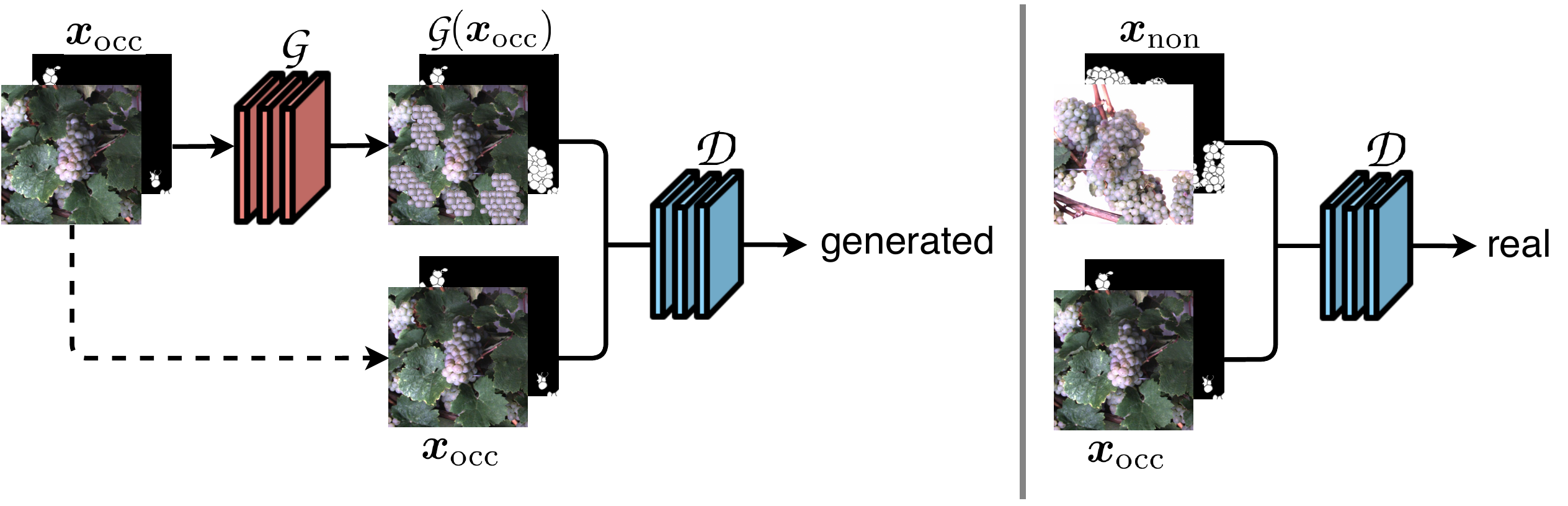}
	\caption{Our network structure based on the Pix2Pix framework \protect\cite{isola2017image}. An input image $\d{x}_{\text{occ}}$ of the occluded domain is transferred to a non-occluded domain using a generator network \ical{G}. The discriminator network \ical{D} distinguish whether the output of \ical{G} looks real or generated.}
	\label{fig:network2}
\end{figure}

The core of our framework is the cGAN that we use to generate images with berries being revealed.
Specifically, we use the Pix2Pix \cite{isola2017image} network and training method, which is illustrated in simplified form in \figref{fig:network2}.

%ADD

The model consists of two networks, the generator, and the discriminator. 
The generator network \ical{G} takes images with occluded berries as an input and is intended to generate images with revealed berries $\ical{G}\,(\d{x}_{\text{occ}}) = \widetilde{\d{x}}_{\text{non}}$ that cannot be distinguished from real images $\d{x}_{\text{non}}$ of the non-occluded domain. 
The adversarially trained discriminator network \ical{D}, on the other side, tries to discriminate between generated images $\widetilde{\d{x}}_{\text{non}}$ and real images $\d{x}_{\text{non}}$. 
The generator used in Pix2Pix is based on a U-Net \cite{ronneberger2015u}, the discriminator \ical{D} on a PatchGAN.
 
% For better clarity, an overview of the used variables can be found in \tabref{table:variables}.
% \begin{table*}[ht]
% \centering
% \begin{tabular}{c|cc}%p{0.08\linewidth}|p{0.49\linewidth}|p{0.49\linewidth}} %0.39
% \hline
% & \textbf{occluded domain} & \textbf{non-occluded domain} \\
% \hline
% \textbf{real} & \cellcolor{gray!18}{$\d{x}_{\text{occ}}$} & \cellcolor{gray!18}{$\d{x}_{\text{non}}$}\\
% \textbf{generated} &  \cellcolor{gray!8}{$\widetilde{\d{x}}_{\text{occ}}$} &  \cellcolor{gray!18}{$\widetilde{\d{x}}_{\text{non}}$}\\
% \hline
% \end{tabular}
% \caption{Denotation. Dark grey correspond to variables that serve as input to the network or are generated as output. Light grey is a non typically output of a cGAN. It is generated in the later experiments based on the dark grey variables and serves for analysis purposes.}
% \label{table:variables}
% \end{table*}

% %%%%%%%%%%%%%%%%%%%%%%%%%%%%%%%%%%%%%%%%%%%%%%%%%%%%%%%%%%%%%%%%%%%%%%%%%%%%%%%%%%%%%%%%%%%%%%%%%%%%%%%%%%%%%%%%%%%%

% Model Training

As described by Goodfellow et al. \cite{goodfellow2014generative}, both parts of GANs are trained simultaneously using a min-max approach.
The goal of the discriminator during training is to be able to distinguish as good as possible between real and generated images.
For this, the discriminator uses a mini-batch of input images $\d{x}_{\text{non}}$ and computes the discriminator loss $\ical{l}_{\ical{D}_{\text{real}}}$. 
Additionally, it uses generated images $\widetilde{\d{x}}_{\text{non}}$ obtained from the generator \ical{G} and computes the corresponding loss $\ical{l}_{\ical{D}_{\text{gen}}}$.
For both computations, the mean squared error (MSE) loss $\ical{l}_\text{MSE}$ is used. 
The overall loss $\ical{l}_\ical{D}$ of the discriminator is calculated as:
\begin{equation}
    \ical{l}_\ical{D} = \frac{1}{2} \cdot (\ical{l}_{\ical{D}_{\text{fake}}} + \ical{l}_{\ical{D}_{\text{real}}})
\end{equation} 
The objective is to maximize this loss, as this means that the discriminator can distinguish between generated and real images with ease. 
The weights of the discriminator network are then updates with respect to this loss.

When generating new images, the generator tries to trick the discriminator at the same time, which is the adversarial part of the network.
Compared to the maximization of the discriminator loss, the objective of the generator is to minimize the generator loss $\ical{l}_G$.
This is calculated from a combination of %$\ical{l}_\text{MSE}$  
MSE loss computed by $\ical{D}\,(\ical{G}\,(\d{x}_{\text{occ}}))$ referred to the reference label \texttt{generated} and a $\ical{l}_1$ loss, which avoids blurring. 
The $\ical{l}_1$ loss is computed using real and generated images, $\d{x}_{\text{non}}$ and $\widetilde{\d{x}}_{\text{non}}$, from the non-occluded domain. The generator loss $\ical{l}_\ical{G}$ is then used to update the generator's weights. 
\begin{equation}
    \ical{l}_\ical{G} = \ical{l}_\text{MSE}(\ical{D}\,(\ical{G}\,(\d{x}_{\text{occ}}))) + \lambda \cdot \ical{l}_1(\d{x}_{\text{non}}, \widetilde{\d{x}}_{\text{non}}) 
\end{equation}
The weighting factor $\lambda$ adjusts the scale of the losses to each other and is, in our case, $\lambda = 100$.
 
The minimization of the generator loss $\ical{l}_\ical{G}$ results in either a strong generator or a very weak discriminator. If the loss becomes maximal, the opposite possibilities can occur. 
The objective is to balance both adversarial goals at the end of the training in the best possible way by realizing both at the same time.

% % %%%%%%%%%%%%%%%%%%%%%%%%%%%%%%%%%%%%%%%%%%%%%%%%%%%%%%%%%%%%%%%%%%%%%%%%%%%%%%%%%%%%%%%%%%%%%%%%%%%%%%%%%%%%%%%%%%%%
% % %%%%%%%%%%%%%%%%%%%%%%%%%%%%%%%%%%%%%%%%%%%%%%%%%%%%%%%%%%%%%%%%%%%%%%%%%%%%%%%%%%%%%%%%%%%%%%%%%%%%%%%%%%%%%%%%%%%%

\section*{Data}
\label{sec:data}

% vine= Rebe, Stock; cane: Bogen; Shoot: Trieb; grape bunch: Traube; berry: einzelne Beere; canopy= Laubwand; defoliation= Entblätterung

\subsection*{Study site}
%ADD
\begin{figure}[t]
	\centering
	\setlength\arrayrulewidth{2pt}
	 \begin{tabular}{c|c}
	{
	 \captionsetup{justification=centering}
	 \begin{minipage}{0.47\textwidth}
		\begin{minipage}{0.38\textwidth}
			\subfloat[Before \\defoliation (SMPH)]{
                \includegraphics[
                width=\textwidth]{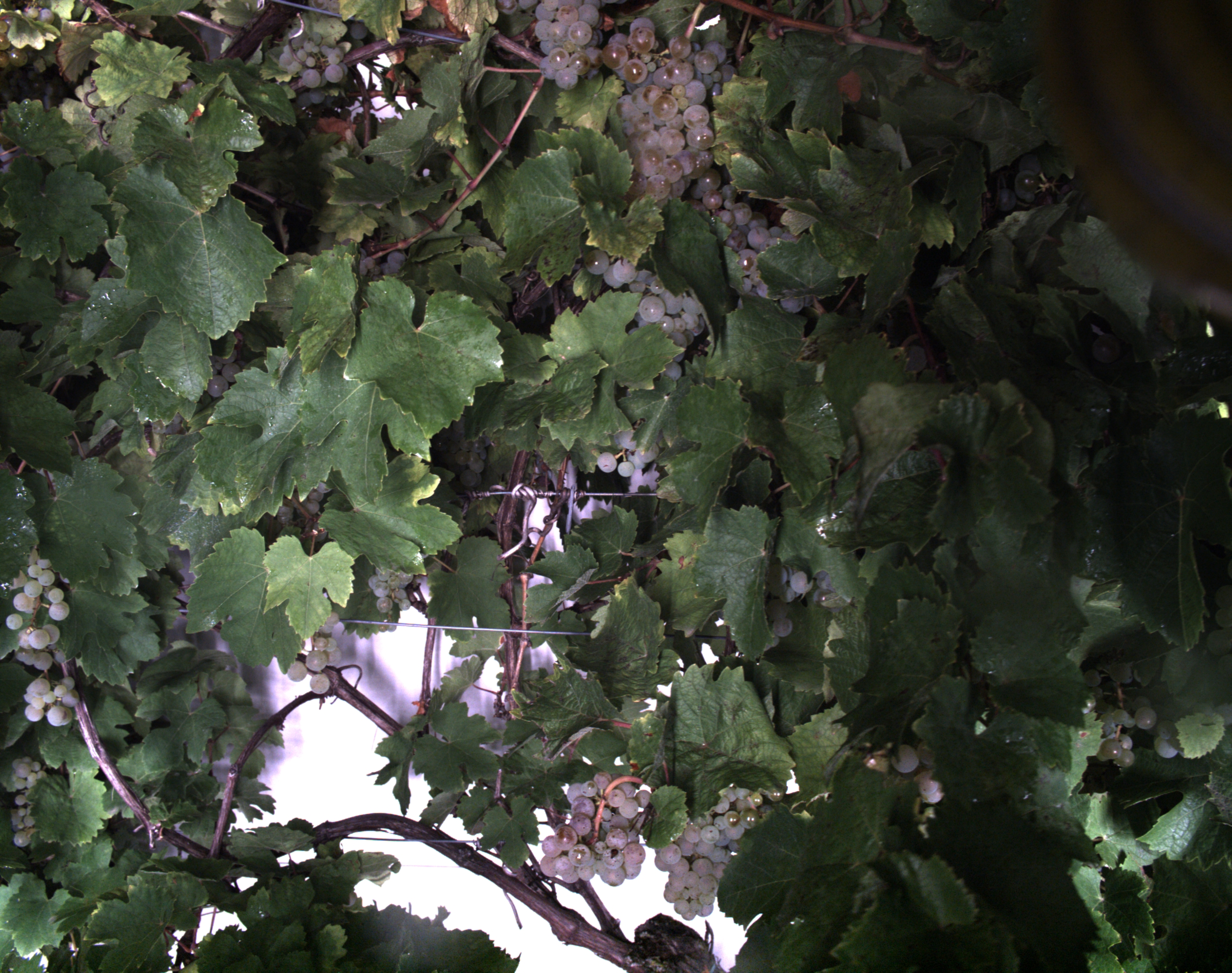}}
                
            \subfloat[After defoliation (SMPH)]{
                 \includegraphics[
                 width=\textwidth]{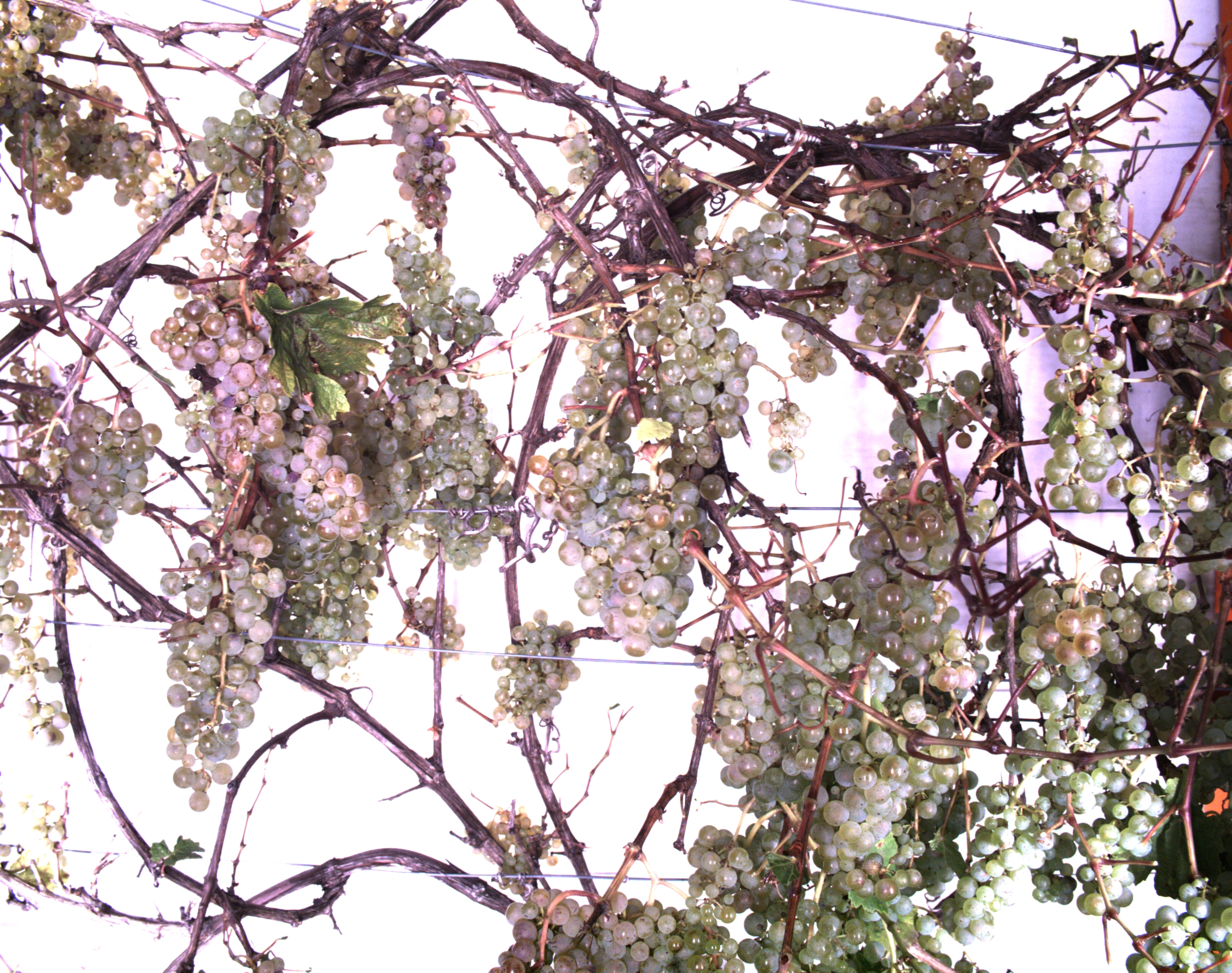}}
% 		 \captionsetup{justification=centering}
		\end{minipage}
		\hspace{2pt}
		\begin{minipage}{0.56\textwidth}
			\subfloat[Semi minimal \\pruned hedge \\(SMPH)]{
                \includegraphics[trim=0 20 0 0, clip,
                width=\textwidth]{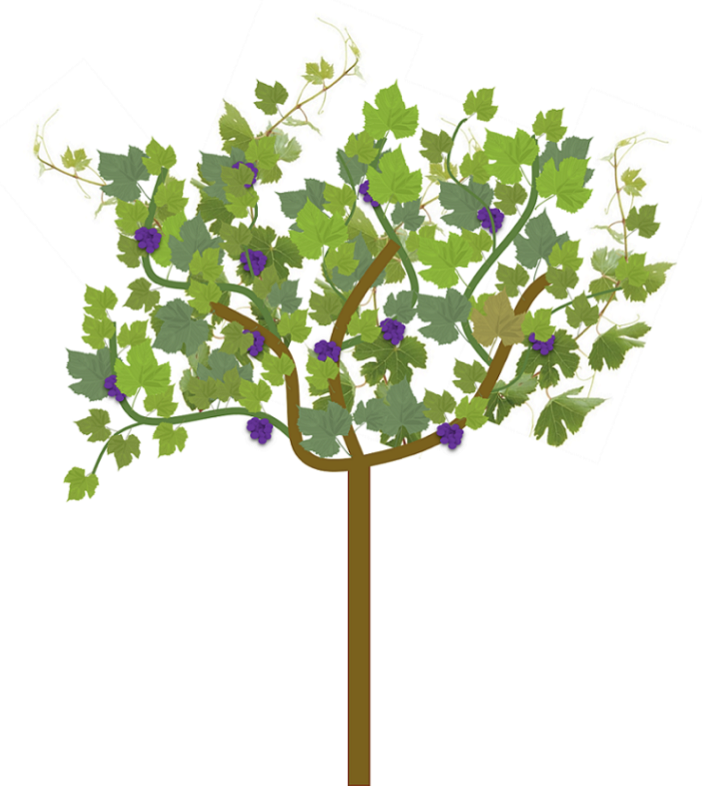}
                \label{fig:MS}}
% 		 \captionsetup{justification=centering}
		\end{minipage}
	\end{minipage}
	} & {
			 \captionsetup{justification=centering}
	 \begin{minipage}{0.47\textwidth}
		\begin{minipage}{0.38\textwidth}
			\subfloat[Before \\defoliation \\(VSP)]{
                \includegraphics[
                width=\textwidth]{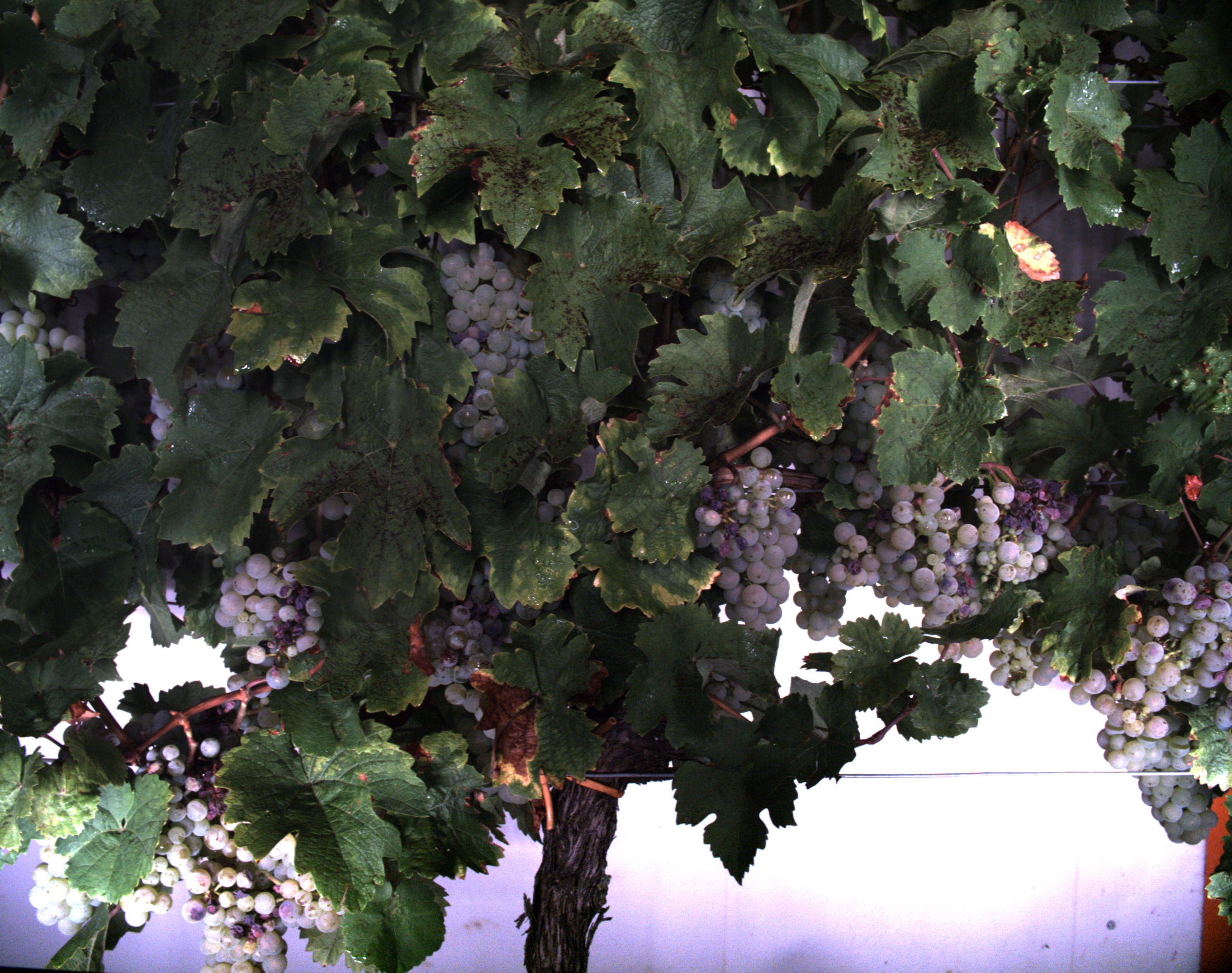}}
                
            \subfloat[After \\defoliation \\(VSP)]{
                \includegraphics[
                width=\textwidth]{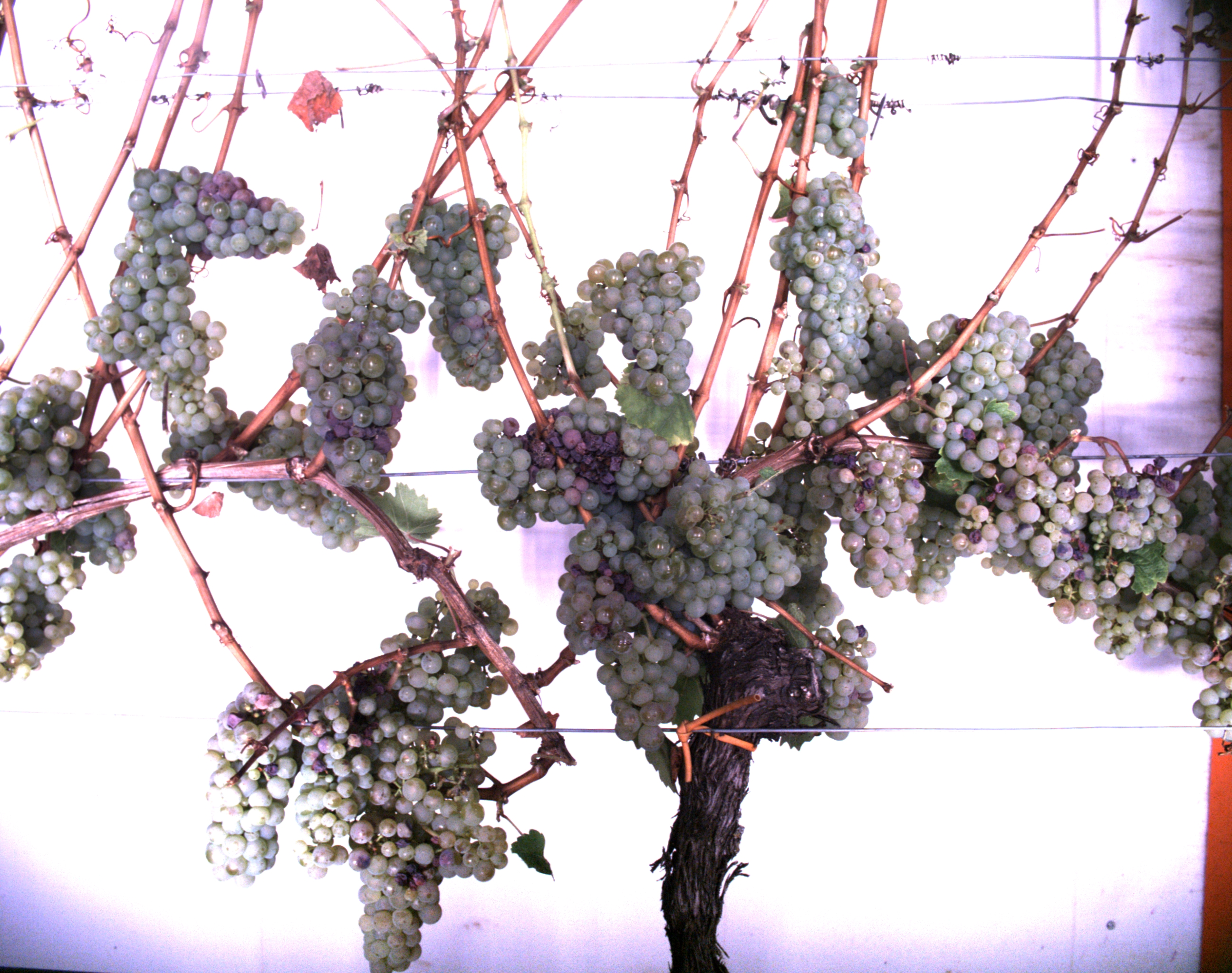}}
		 \captionsetup{justification=centering}
		\end{minipage}
		\hspace{2pt}
		\begin{minipage}{0.56\textwidth}
			\subfloat[Vertical shoot\\positioned system\\(VSP)]{
                \includegraphics[trim=0 20 0 0, clip,
                width=\textwidth]{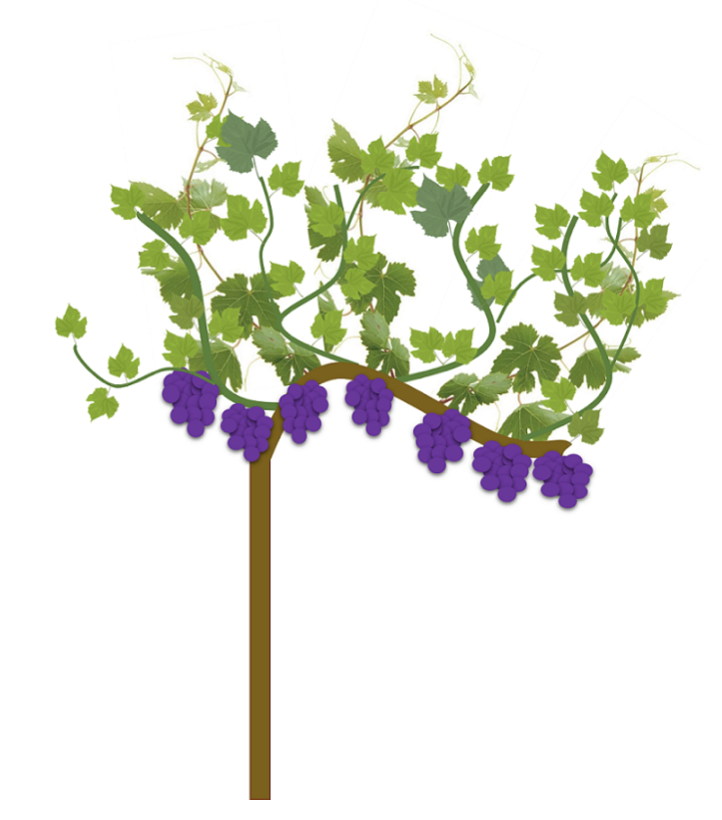}
                \label{fig:SS}}
                
		 \captionsetup{justification=centering}
		\end{minipage}
	
	\end{minipage}
	}
	\end{tabular}
		
	\caption{Acquired images of the Phenoliner \protect\cite{kicherer2017phenoliner} for two different kind of cuttings SMPH ((a) - (c)) and VSP ((d) - (f)). (a) and (d) show example images before defoliation in the occluded domain. (b) and (e) show example images after defoliation in the non-occluded domain.}
	\label{fig:Examples_DomainAandB}
\end{figure}

The data, we use in this work, was acquired at the experimental fields of JKI Geilweilerhof located in Siebeldingen, Germany. 
It was acquired using the Phenoliner \cite{kicherer2017phenoliner}, a reconstructed grape harvester that can be used as a phenotyping platform to acquire geo-referenced sensor data directly in the field.
A description of the on-board camera setup can be found in Zabawa et al. \cite{zabawa2020counting}. 
The images were acquired in two different training systems of the cultivar Riesling (DEU098\_VIVC10077\_Riesling\_Weiss\_DEU098-2008-085): (1) Vertical shoot positioned (VSP) vines (\figref{fig:SS}) and (2) vines trained as semi minimal pruned hedges (SMPH) (\figref{fig:MS}) were chosen due to diverse difficulties in image analysis \cite{zabawa2020counting}. 
The acquisition took place in September 2019 and 2020, before harvest at the plant growth stage BBCH89, and in each year the images were taken one day before (\figref{fig:Examples_DomainAandB} (a,d)) and right after defoliation (\figref{fig:Examples_DomainAandB} (b,e)). In 2019 50 cm and 2020, respectively, 100 cm of the grapevine canopy have been defoliated.

In our framework, we use three different types of inputs:
\begin{itemize}
  \item \textbf{Natural data}: Images acquired in the vineyard before and after defoliation. For our studies, we use grayscale images. We denote this dataset with $\mathcal X^{\text{N}}$.
  \item \textbf{Synthetic data}: Images acquired in the vineyard after defoliation. Images with occluded berries are synthetically generated. We denote this dataset with $\mathcal X^{\text{S}}$.
  \item \textbf{Semantic segmentation masks (berry masks)}: So-called berry masks obtained by a semantic segmentation approach presented in \cite{zabawa2019detection}. Each pixel in these images is assigned to the class \texttt{berry}, \texttt{berry-edge}, or \texttt{background}. We denote this data as $\mathcal X_{\text{A}}$.
\end{itemize}
The use of the mentioned grayscale images is indicated by the index $\text{L}$ and with index $\text{A}$ we denote the use of the berry masks. 
Moreover, we define $\mathcal X_{\text{LA}}$ as the input where the grayscale image and the berry mask are stacked to form a multichannel 2D input.
In the following, the used data is explained in more detail.

% %%%%%%%%%%%%%%%%%%%%%%%%%%%%%%%%%%%%%%%%%%%%%%%%%%%%%%%%%%%%%%%%%%%%%%%%%%%%%%%%%%%%%%%%%%%%%%%%%%%%%%%%%%%%%%%%%%%%

\subsection*{Natural data}
\label{sec:realData}
% general things about the data
We convert the acquired RGB images into grayscale images in order to develop an efficient approach that is independent of the berry color. 
Covering the whole variability of possible berry colors is complex and not feasible in our case. For example, in the case of green berries, the color also does not serve to differentiate them from leaves. 

% alignment
Since the Phenoliner platform revisits the vine row for each data collection of the two domains, the images depicting the same scene are acquired at different times and from different positions, leading to differences in translation, rotation and scale.
Moreover, the defoliation of vines causes a movement of the branches and grape bunches, and additional environmental changes between the two acquisition time points can result in different scenes in the aligned patches.

However, to obtain aligned image pairs for a qualitative evaluation, we manually align images from both domains. 
%Since our approach requires image pairs of the two domains that represent a one-to-one mapping of the contained scene, we apply a spatial transformation to align them.
% Images that show the same scene but occur in different domains are referred to as paired images.
For this, we compute a 4-parameter Helmert transformation \cite{helmert1880mathematischen} between the two domains, where we manually define corresponding keypoints per image pair to calculate the parameters. 
We apply this transformation to images from the non-occluded domain to register them to the occluded domain. 

\begin{figure}[t]
	\centering
	\begin{minipage}{0.65\textwidth} %55
		\subfloat[Occluded domain.]{
          \includegraphics[
    width=0.33\textwidth]{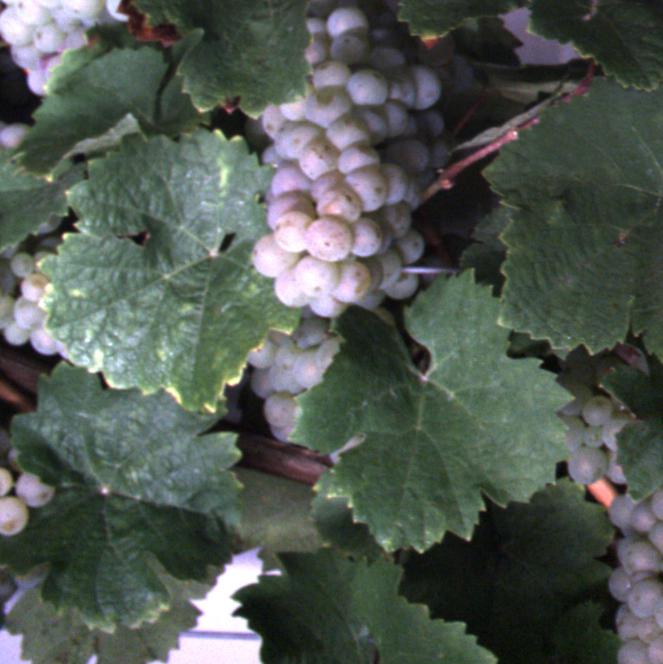}
     \includegraphics[
    width=0.33\textwidth]{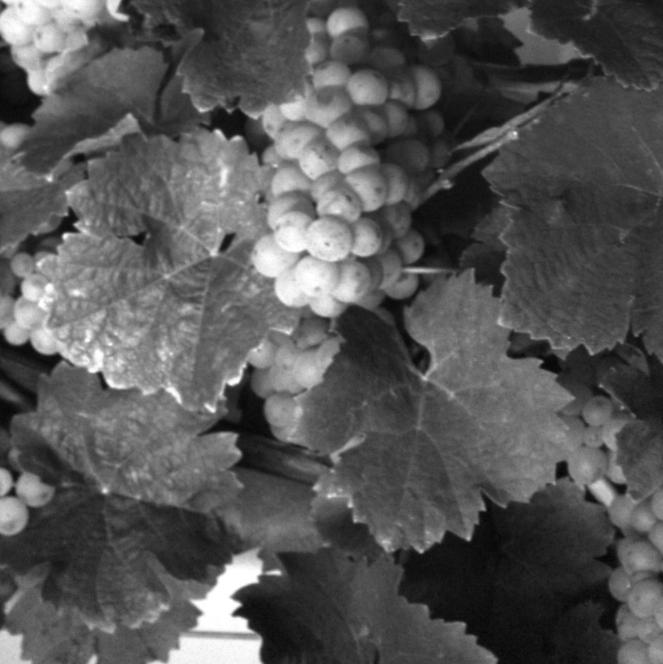}
     \includegraphics[
    width=0.33\textwidth]{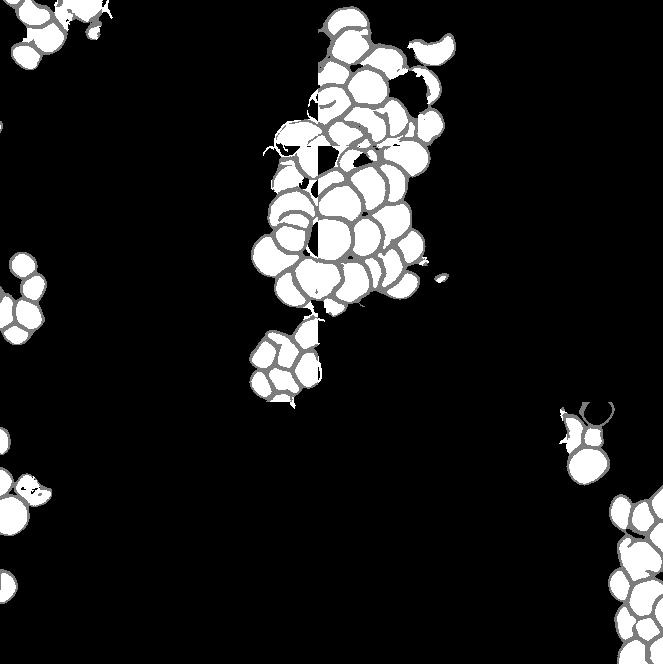}
    \label{fig:occludedDomain}}
% 	\captionsetup{justification=centering}
  	\end{minipage}
  
	\begin{minipage}{0.65\textwidth} %55
	\subfloat[Non-occluded domain.]{
			\includegraphics[
                width=0.33\textwidth]{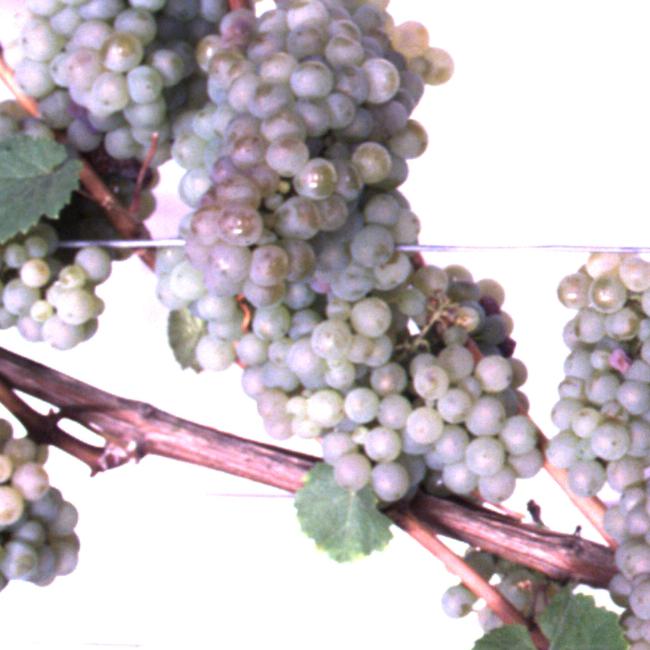}
          \includegraphics[
                width=0.33\textwidth]{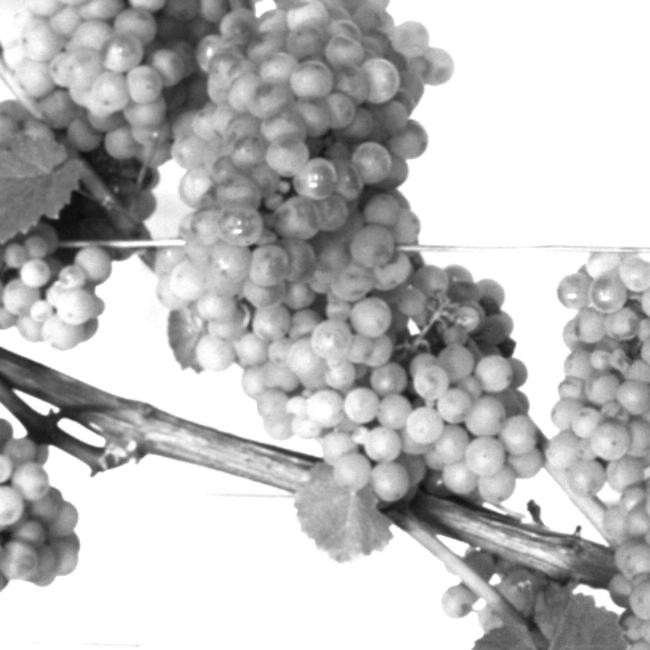}
            \includegraphics[
                width=0.33\textwidth]{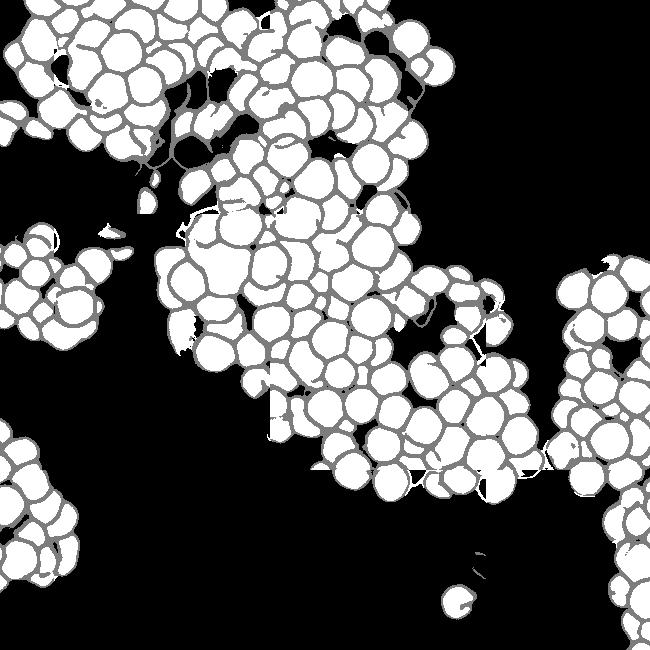}
         \label{fig:NonOccludedDomain}
% 		 \captionsetup{justification=centering}
    }
	\end{minipage}

	\caption{Example patches extracted from images of the occluded and non-occluded domain. One row shows the same patch in RGB, grayscale and berry mask format. One column represents a patch pair $\lbrace \d{x}_{\text{occ}}, \d{x}_{\text{non}} \rbrace$.}
	\label{fig:examplePatches}
\end{figure}

Due to computational limitations, we use a sliding window of size $\SI{656}{\px} \times \SI{656}{\px}$ and stride $\SI{162}{\px}$ to extract patches from the grayscale images.
\figref{fig:examplePatches} illustrates one RGB patch, the grayscale patch, and the corresponding berry mask, which is explained in the following subsection, for both domains.
We denote the aligned patch pair $\d{x}^{\text{N}} = \lbrace \d{x}^{\text{N}}_{\text{occ}}, \d{x}^{\text{N}}_{\text{non}} \rbrace$, where $\d{x}^{\text{N}} \in \mathcal{X}^{\text{N}}$.

%ADD

% %%%%%%%%%%%%%%%%%%%%%%%%%%%%%%%%%%%%%%%%%%%%%%%%%%%%%%%%%%%%%%%%%%%%%%%%%%%%%%%%%%%%%%%%%%%%%%%%%%%%%%%%%%%%%%%%%%%%

\subsection*{Semantic segmentation mask (berry mask)}
\label{sec:maskSegmentation}

Besides the acquired images, we use a berry mask, obtained with a semantic segmentation approach, presented by Zabawa et al. \cite{zabawa2019detection}. 
The identification of regions containing berries and the detection of single berry instances is performed with a convolutional neural network. 
The network uses a MobilenetV2 \cite{Sandler18} encoder and a DeepLabV3+ decoder \cite{Chen18}. 
The network assigns each image pixel to one of the classes \texttt{background}, \texttt{berry-edge}, or \texttt{berry}, which corresponds to the grayscale values $0$, $127$ and $255$.
In contrast to a standard semantic segmentation without distinguishing between different instances, we use the additional class \texttt{berry-edge} to ensure the separation of single berries, which allows the counting of berries using a connected component approach. 

For our task of generating a likely scenario behind leaves, the berry mask supports the discriminability of relevant information like the berries from the surrounding information in the image, and the generation of separated berries. 
In addition, since the berry masks contain a masking of existing berries, it provides further knowledge about which areas in the images do not show occlusions and should be preserved in the revelation process and where potentially occlusions might appear, which are areas that are unmasked.

Since we are interested in scenes in the image that depict berries, we only integrate patch pairs in training and testing, whose berry mask of non-occluded domain contains more than 1/24 background pixels and mask of the occluded domain contains at least one pixel whose class differs from the background class.

% %%%%%%%%%%%%%%%%%%%%%%%%%%%%%%%%%%%%%%%%%%%%%%%%%%%%%%%%%%%%%%%%%%%%%%%%%%%%%%%%%%%%%%%%%%%%%%%%%%%%%%%%%%%%%%%%%%%%

\subsection*{Synthetic data}
\label{sec:synthData}
One challenge for our application is that the amount of paired data from both domains containing both, occluded and non-occluded regions of berries, is limited for training a reliable model and for evaluation.
We, therefore, resort to generate artificially modified images, where berries are artificially occluded, based on natural images of defoliated plants.
This allows us to generated a large dataset to ease the described lack of natural images of both domains.
We denote this synthetic dataset with $\mathcal X^{\text{S}}$.
The natural patches $\d{x}^{\text{N}}_{\text{non}}$ of the non-occluded domain serve as a basis. 
We create paired patches $\lbrace \d{x}^S_{\text{non}}, \d{x}^{\text{S}}_{\text{occ}} \rbrace$ where $\d{x}^{\text{S}}_{\text{non}} = 
\d{x}^{\text{N}}_{\text{non}}$.

%ADD
\begin{figure}[t]
	\centering
    \includegraphics[%trim=50 50 141 30, clip,
    width=0.13\textwidth]{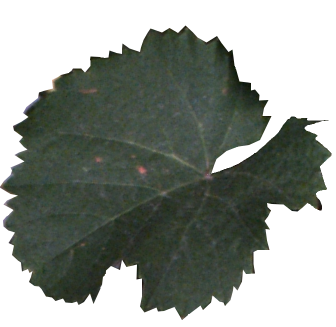}
  \includegraphics[%trim=50 50 141 30, clip,
    width=0.13\textwidth]{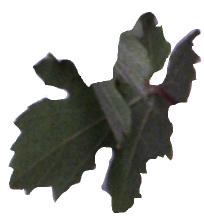}
     \includegraphics[%trim=50 50 141 30, clip,
    width=0.14\textwidth]{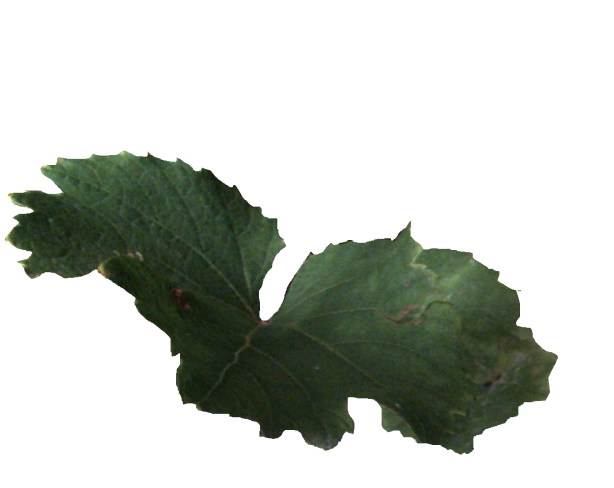}
     \includegraphics[%trim=50 50 141 30, clip,
    width=0.13\textwidth]{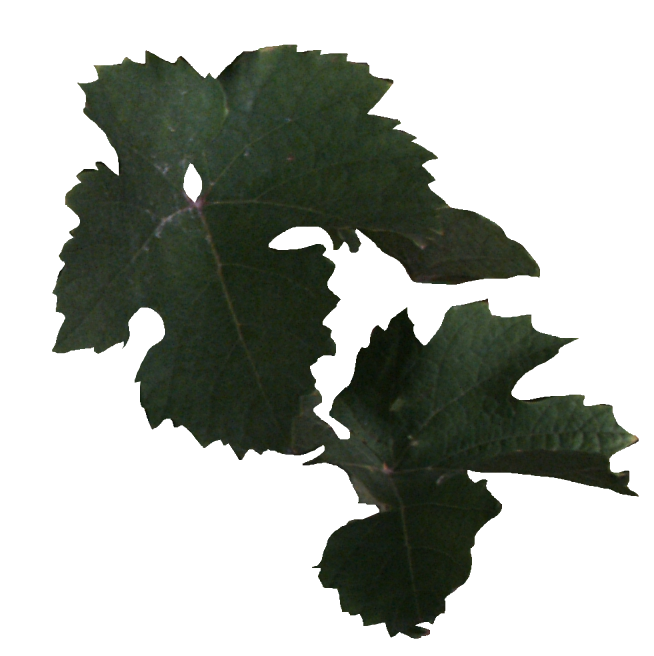}
     \includegraphics[%trim=50 50 141 30, clip,
    width=0.09\textwidth]{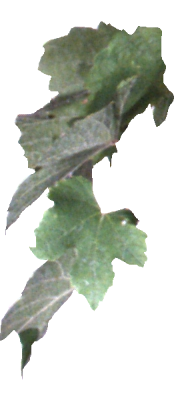}
     \includegraphics[%trim=50 50 141 30, clip,
    width=0.13\textwidth]{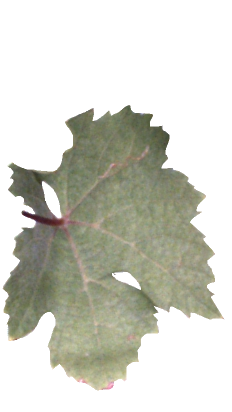}
     \includegraphics[%trim=50 50 141 30, clip,
    width=0.14\textwidth]{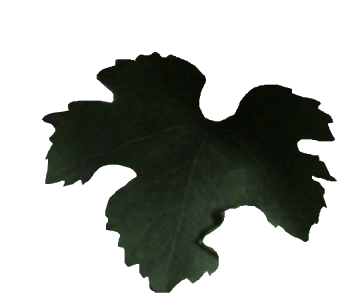}
	\caption{Exemplary leaves used for data augmentation.}
	\label{fig:Leaves}
\end{figure}

\begin{figure}[t]
% 	\centering
\begin{minipage}{\textwidth}
  \begin{minipage}{0.5\textwidth} %0.55
    \centering
    \subfloat[Data composition.]{
    \includegraphics[trim=0 0 0 0, clip,
        width=0.96\textwidth]{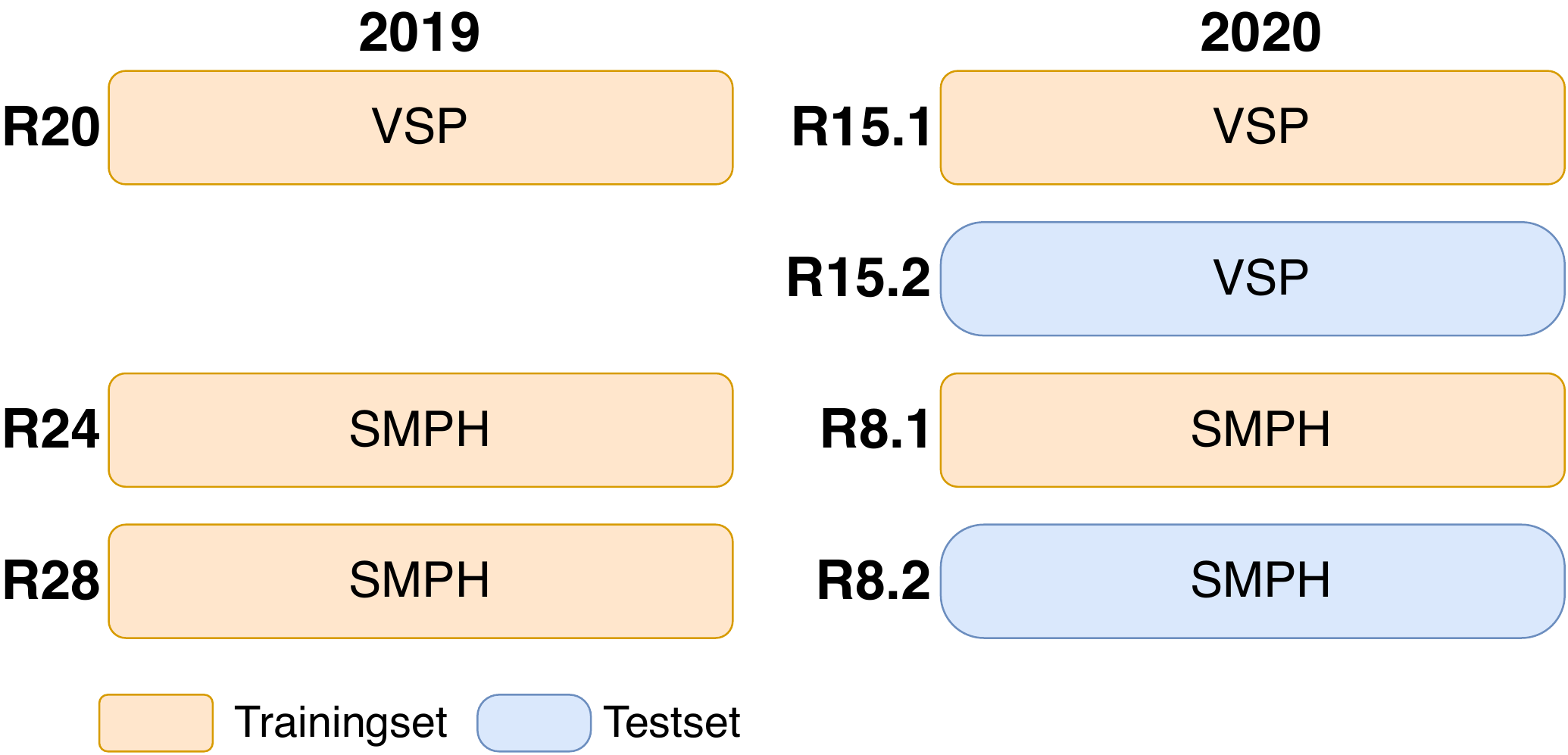}
        \label{fig:visualDataComposition}
    }
  \end{minipage}
%   \hfill
  \begin{minipage}{0.4\textwidth} %[b]
      \begin{minipage}{\textwidth} %[b]
        \centering
        % \begin{tabular*}[t]
            %\centering
            % {\tabulinesep=1.2mm
            % \vspace{17mm}
            % \subfloat[Amount of real data.]{
            % % \captionsetup{skip=10mm}
            %     \begin{tabu}{|c||c|c|c|}%p{0.08\linewidth}|p{0.49\linewidth}|p{0.49\linewidth}} %tabular %0.39
            %         \hline
            %          & Training & Testing &  Sum\\
            %         \hline
            %         \multirow{2}{*}{SMPH} & 2.801 &  1.921&  4.722 \\
            %         %  \cline{8-12}
            %          & $59.32\%$ & $40.68\%$ & $100\%$\\
            %          \cline{2-4}
            %          \hline
            %          \multirow{2}{*}{VSP} & 1.774 &  1.520 & 3.294\\
            %          &  $53.86\%$ & $46.14\%$ & $100\%$\\
            %         \cline{2-4}
            %         \hline
            %     \end{tabu}%}% tabular
            % \label{table:amountOfRealData}
            % }
            %caption{Definitions of the used datasets. The table shows which kind of data is used for which experiment.}
        % \end{tabular*}
        \end{minipage}
        
        \begin{minipage}{\textwidth} %[b]
        \centering
       
            \subfloat[Amount of synthetic data.]{
                \begin{tabu}{|c||c|c|c|}
                    \hline
                     & Training & Testing &  Sum\\
                    \hline
                    \multirow{2}{*}{SMPH} & 25.209
                    &  5.763 &  30.972\\
                     & $81.39\%$ & $18.61\%$ & $100\%$\\
                     \cline{2-4}
                     \hline
                     \multirow{2}{*}{VSP} & 15.996 & 4.560 &  20.556\\
                     & $77.82\%$ & $22.18\%$ & $100\%$\\
                    \cline{2-4}
                    \hline
                \end{tabu}
            \label{table:amountOfSynthData}
            }
        \end{minipage}

    \end{minipage}
    \captionof{figure}{Dataset composition.	(a) shows the visual division of training and test data for the two data sets semi minimal pruned hedge (SMPH)	(see \figref{fig:MS}) and vertical shoot positioned system (VSP)	(see \figref{fig:SS}) which were collected in the years 2019 and 2020. The data sets are taken from different rows R. Each row corresponds to one of the two sets. In 2020 the same rows were run twice. Once left (R15.1 and R8.1) and once right (R15.2 and R8.2) of the row.
	%$15\%$ of the total data (2019 and 2020) is used for testing and is taken from data collected in 2020. The rest of collected data is also used for training. 
	Orange marked areas are used for training, blue marked ones for testing. The table shown in (b) indicates the corresponding numbers of training and test data patches.}
	\label{fig:dataComposition}
  \end{minipage}
\end{figure}

To generate $\d{x}^{\text{S}}_{\text{occ}}$, we apply artificial data modification on both training and test data. We artificially occlude the patches using 24 different wine leaves (\figref{fig:Leaves}) with various shapes extracted from the natural dataset and use them as occluding objects in the patches. 
We use 18 leaves for augmenting the training set and 6 leaves to augment the test set. 
On the basis of one image patch $\d{x}^{\text{S}}_{\text{non}}$, we create up to 9 corresponding synthetically augmented versions of $\d{x}^{\text{S}}_{\text{occ}}$ for the training set, resulting in 9 aligned image patch pairs.
During the procedure, a leaf is randomly selected from the set of leaves and rotated by a randomly chosen angle $\alpha \in \lbrace -50, -30, -10, 0, 10, 30, 50, 70 \rbrace$.
Converted to grayscale, it randomly overlays the grayscale patch and occludes parts of the visible berries. 
These steps are also performed for patches $\d{x}^{\text{S}}_{\text{non}}$ of the test set. However, here only three new patch pairs are created.
After applying artificial data modification, the proportion of test data amounts to approx. 18\% - 23\% of the extracted patches depending on the type of defoliation, see \tabref{table:amountOfSynthData}.
The split of the data into training and test data is illustrated visually and numerically in \figref{fig:dataComposition}. 
The test data is taken from the dataset collected in 2020, see \figref{fig:visualDataComposition}. 

%ADD

%We randomly overlay the grayscale patch $L$ with the leaf. Thus, the leaf occludes parts of the visible berries. %To create more data, we apply data augmentation on the data
%We repeat these steps three times for each patch $\d{x}^{\text{S}}_{\text{non}}$ of the test set, so that three new patch pairs are created. %So take three randomly chosen leaves, rotate each leaf randomly by an angle $\alpha$ and put it on the grayscale patch of $\d{x}^{\text{S}}_{\text{non}}$ to create the corresponding $\d{x}^{\text{S}}_{\text{occ}}$. Less than for training to reduce the occurrence of systematics. 
%Notice that $\d{x}^{\text{S}}_{\text{occ}} = \d{x}^{\text{ES}}_{\text{occ}}$.

\begin{figure}[t]
	\centering
      \includegraphics[%trim=3 3 10 10, clip,
    width=0.8\textwidth]{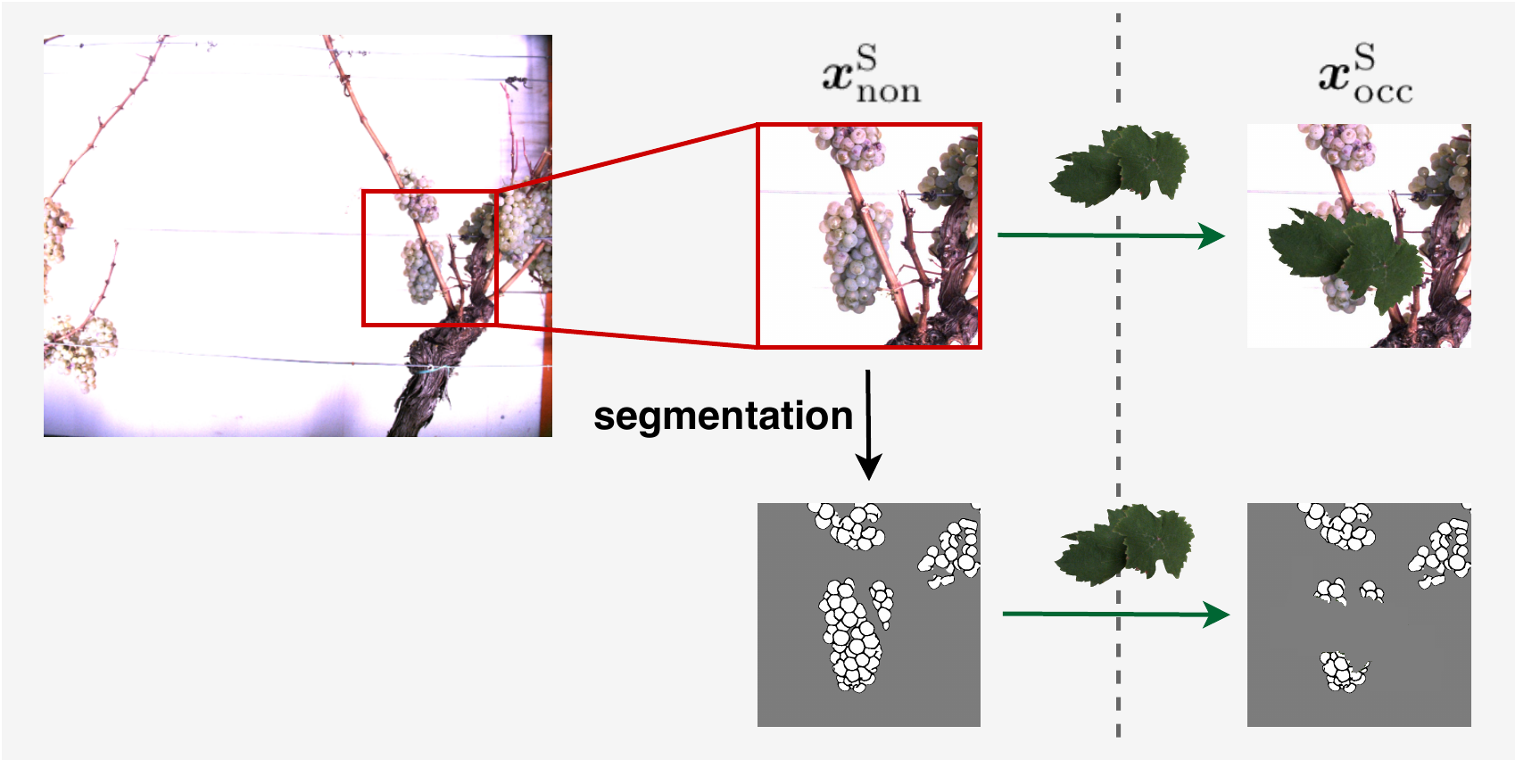}
    
	\caption{Visualization of synthetic data creation. The visualization indicates the use of an artificial leaf to calculate the corresponding mask of $\d{x}^{\text{S}}_{\text{occ}}$ of the occluded domain instead of using the segmentation mask prediction based on the occluded RGB-image.}
	\label{fig:synthDataVsModifiedSynthData}
\end{figure}

For each synthetic grayscale image, we calculate a corresponding berry mask.
However, depending on the used procedure, the appearance of the berry mask differs.
In our work, we create the masks for the two domains as illustrated in \figref{fig:synthDataVsModifiedSynthData}. 
%\figref{fig:synthDataVsModifiedSynthData} shows the difference between the derived synthetic datasets $\mathcal X^{\text{S}}$ and $\mathcal X^{\text{MS}}$ by means of the berry mask of $\d{x}^{\text{S}}_{\text{occ}}$ and $\d{x}^{\text{MS}}_{\text{occ}}$ corresponding to the same grayscale image. 
The mask of the non-occluded patch $\d{x}^{\text{S}}_{\text{non}}$ is based on the segmentation step, described in Sec. \hyperref[sec:maskSegmentation]{\enquote{Semantic segmentation mask}}, %\secref{sec:maskSegmentation}, 
which needs RGB images as input.
We compute the mask of $\d{x}^{\text{S}}_{\text{occ}}$ by overlaying the pixels of the non-occluded mask of $\d{x}^{\text{S}}_{\text{non}}$ that are covered with a leaf in the RGB, or respectively grayscale patch.
These pixels in the berry mask are assigned to the class \texttt{background}.
The leaf pixels adjacent to berry pixels are changed to \texttt{berry-edge} pixels. 
In this way, the overlapped berries have a closed contour. 
By adding these edges, the synthetic data thus has the same characteristics as berry masks derived from the natural data.
With this step, we create two corresponding masks, $\d{x}^{\text{S}}_{\text{occ}}$ and $\d{x}^{\text{S}}_{\text{non}}$, which match exactly in the non-occluded pixel.

Another way to define the occluded mask is a direct computation as for $\d{x}^{\text{S}}_{\text{non}}$ using the segmentation step to create a predicted mask of the patch.
%In contrast to the first procedure, we compute the mask of the second synthetic dataset $\d{x}^{\text{S}}_{\text{occ}}$ directly using the corresponding grayscale image patch, see \figref{fig:synthData}. 
Since the berry mask is an estimation, the class of individual non-occluded pixels may differ between $\d{x}^{\text{S}}_{\text{non}}$ and $\d{x}^{\text{S}}_{\text{occ}}$.
For a simplified analysis, we have chosen the first option.

%ADD

Overall, for dataset $\prescript{\text{VSP}}{}{\mathcal X}^{\text{S}}$, we obtain 20.556 synthetic patch pairs, and for dataset $\prescript{\text{SMPH}}{}{\mathcal X}^{\text{S}}$, we obtain 30.972 synthetic patch pairs (\tabref{table:amountOfSynthData}).

% %%%%%%%%%%%%%%%%%%%%%%%%%%%%%%%%%%%%%%%%%%%%%%%%%%%%%%%%%%%%%%%%%%%%%%%%%%%%%%%%%%%%%%%%%%%%%%%%%%%%%%%%%%%%%%%%%%%%

\subsection*{Challenges}
Various challenges occur in the data, which influence our training and thus our results. 
Since our reference masks are not manually derived but are estimations, uncertainties can occur.
For example, not all visible berries are entirely shown in the images of the non-occluded domain. 
Therefore, it can happen that either berries are missed or only partly detected in the mask. 
Additionally, the estimated contour in the berry mask may not be closed and parts of the berry region may be classified as background.  
Thus, these errors in the reference could be learned in the model. 
Furthermore, there are images in the non-occluded domain, which contain leaves despite defoliation. 
In an ideal case, the model learns to ignore these faults in defoliation. 
Other challenges are the varying sharpness of the patches. 
This can be caused by resizing the data, shadows, or the varying distance of the berries to the camera. 
Furthermore, the illumination varies within the data, e.g., due to the coverage by surrounding objects like branches or leaves or the distance of the berries to the camera. 
Also worth noting are the different growth stages of the grapes in 2019 and 2020, so the grapes have different sizes due to different berry sizes.

% %%%%%%%%%%%%%%%%%%%%%%%%%%%%%%%%%%%%%%%%%%%%%%%%%%%%%%%%%%%%%%%%%%%%%%%%%%%%%%%%%%%%%%%%%%%%%%%%%%%%%%%%%%%%%%%%%%%%

\section*{Experiments}
\subsection*{Experimental setup}

Our experiments are designed to apply a domain-transfer using cGANs (Sec. \enquote{\nameref{sec:GAN}}) %(\secref{sec:GAN}) 
to (1) learn a distribution by which we can generate a likely scenario of how occluded grapes could look like depending on the input, and (2) improve the counting of grapevine berries in images.
To address the challenge of limited amount of natural data $\mathcal X^{\text{N}}$, we perform four experiments based on a synthetic dataset $\mathcal X^{\text{S}}$. %in Sec. \hyperref[sec:Experiment1]{\enquote{Experiment 1}}%, Sec. \hyperref[sec:Experiment2]{\enquote{Experiment 2}}, Sec. \hyperref[sec:Experiment3]{\enquote{Experiment 3}} and Sec. 
%to Sec. \hyperref[sec:Experiment4]{\enquote{Experiment 4}}. %\secref{sec:Experiment1}, \secref{sec:Experiment2}, \secref{sec:Experiment3} and \secref{sec:Experiment4}. 
In Sec. \hyperref[sec:Experiment5]{\enquote{Experiment 5}}, %\secref{sec:Experiment5}, 
we show the applicability to natural data $\mathcal X^{\text{N}}$ based on the models and results learned in earlier experiments.

For our experiments, we define five different datasets, which are listed in \tabref{table:Datasets}.
In addition to the natural data, described in Sec. \enquote{\nameref{sec:realData}}, %\secref{sec:realData}, 
we introduce a synthetic dataset in Sec. \enquote{\nameref{sec:synthData}}. %\secref{sec:synthData}. 
\emph{Datasets 1-5} will again be divided into the different types of defoliation SMPH and VSP. 
For our experiments, we also distinguish the set of input channels used. 
In most datasets, we use a combination of grayscale image (L) and berry mask (A), denoted as LA. 
In \emph{Dataset 2}, we only use the berry mask without grayscale information.

 \begin{table*}[t]
    \centering
    {\tabulinesep=1.2mm
    \begin{tabu}{|c||c|c|c|c|c|c|c||c|c|c|c|c|}
        \hline
         \multirow{2}{*}{Definition} &  \multirow{2}{*}{R} & \multirow{2}{*}{MS} &  \multirow{2}{*}{S} &  \multirow{2}{*}{SMPH} &  \multirow{2}{*}{VSP} &  \multirow{2}{*}{LA} &  \multirow{2}{*}{A} & \multicolumn{5}{c|}{Experiment}\\
         \cline{9-13}
         & & & & & & & & 1 & 2 & 3 & 4 & 5 \\
        \hline
        \emph{Dataset 1} & & \xmark & & & \xmark & \xmark & & \xmark & \xmark & \xmark & \xmark &\\
        \emph{Dataset 2} & & \xmark & & & \xmark & & \xmark & \xmark & & & & \\
        \emph{Dataset 3} & & \xmark & & \xmark & & \xmark & & & & & \xmark  & \\
        \emph{Dataset 4} & \xmark & & & & \xmark & \xmark & & & & & & \xmark \\
        \emph{Dataset 5} & \xmark & & & \xmark & & \xmark & & & & & & \xmark \\
        
        \hline
    \end{tabu}}
    \caption{Definitions of the used datasets. The table shows which kind of data is used for which experiment.}
\label{table:Datasets}
\end{table*}

We resize all image patches to a uniform size of $\SI{286}{\px} \times \SI{286}{\px}$ with nearest neighbor interpolation. 
During training, we follow the procedure of \cite{isola2017image} and add small variations to the data in each epoch by randomly cropping patches of size $\SI{256}{\px}\times \SI{256}{\px}$ from the given patches. Additionally, patches are randomly flipped vertically, and the values within the patches are scaled and shifted to range [-1,1]. For testing, only scaling and shifting of the values to range [-1,1] is carried out. The network output is scaled back to value range [0,255] to obtain a visually evaluable output.

To train the models, we use an Intel Core i7-6850K 3.60 GHz processor and two Geforce GTX 1080Ti with 11 GB RAM. The models are trained over 600 epochs.
We use the Adam optimizer, where the learning rate is constant at 0.0004 for the first 300 epochs and is reduced linearly towards 0 for the last 300 epochs.

\subsection*{Model evaluation}
\label{sec:modelEvaluation}
\subsubsection*{Data post-processing}
\label{sec:dataPostProcessing}

After the test phase, the generated masks do not only contain the values 0, 127 and 255. There are also mixed pixels that are not clearly assigned to one of the three classes. We use thresholding to ensure that only the values 0, 127 and 255 appear in the mask. We use the following class assignment. 
\begin{description}
  \item[$\bullet$] Pixel values in the interval [0,50] are set to value 0 and assigned to class \texttt{background}. 
  \item[$\bullet$] Pixel values in the interval (50,180] are set to value 127 and assigned to class \texttt{berry-edge}. 
  \item[$\bullet$] Pixel values in the interval (180,255] are set to value 255 and assigned to class \texttt{berry}. 
\end{description}

\subsubsection*{Evaluation metrics}
\label{sec:evalMetrics}

In the following, we describe several evaluation metrics used for our experiments.
The first metric we use is the \emph{area} $F_c$, that we define as the number of pixels within a mask that correspond to a class $c$ with $c \in \lbrace \texttt{background}, \texttt{berry-}\\ \texttt{edge}, \texttt{berry} \rbrace$. 
With area $F_c$ and the generated area $\widetilde{F}_c$, which is based on the generated mask of the cGAN, we calculate the \emph{intersection over union} IoU by dividing the area of overlap by the area of union.

\begin{equation}
    \text{IoU}_c = \frac{F_c \cap \widetilde{F}_c} {F_c \cup \widetilde{F}_c}
\end{equation}
The IoU compares the similarity between two arbitrary shapes.

% %%%%%%%%%%%%%%%%%%%%%%%%%%%%%%%%%%%%%%%%%%%%%%%%%%%%%%%%%%%%%%%%%%%%%%%%%%%%%%%%%%%%%%%%%%%%%%%%%%%%%%%%%%%%%%%%%%%%

The second metric we use is the \emph{pearson product-moment correlation coefficient}. 
It gives a measure of the degree of linear relationship between two variables. 
The correlation coefficient is obtained by the correlation coefficient matrix \m Q which is calculated by means of the covariance matrix \m C,

\begin{equation}
    \m Q_{i,j} = \frac{\m C_{i,j}}{\sqrt{\m C_{i,i} \cdot \m C_{i,j}}}
\end{equation}

where $i$ and $j$ indicate the row and column index, respectively. 
The values of \m Q are between -1 and 1, inclusive. 
The correlation coefficient $\rho$ between two variables can then be expressed by $\rho = \m Q_{0,1}$.
We use the correlation to compare the generated images $\widetilde{\d{x}}_{\text{non}}$ from the model with the input $\d{x}_{\text{occ}}$ as well as the target output $\d{x}_{\text{non}}$.

% %%%%%%%%%%%%%%%%%%%%%%%%%%%%%%%%%%%%%%%%%%%%%%%%%%%%%%%%%%%%%%%%%%%%%%%%%%%%%%%%%%%%%%%%%%%%%%%%%%%%%%%%%%%%%%%%%%%%

The \emph{coefficient of determination}, also denoted by $R^2$, indicates the relationship between a predicted value with respect to a reference value. 
It provides a measure of how well observed references are replicated by the model. 
In our case, we use the $R^2$ value for the comparison between the predicted number of berries generated by the model and the reference number from the berries manually counted in the non-occluded domain. Plots, as illustrated in \figref{fig:R2_counting_xNon}, represent the generated distribution of the model compared to the reference. Please note, that the gray line represents the reference values. The optimal generated samples are distributed along this line, reflected in a $R^2$ value equal to 1.

% %%%%%%%%%%%%%%%%%%%%%%%%%%%%%%%%%%%%%%%%%%%%%%%%%%%%%%%%%%%%%%%%%%%%%%%%%%%%%%%%%%%%%%%%%%%%%%%%%%%%%%%%%%%%%%%%%%%%

The \emph{counting} is based on the procedure described in the work of Zabawa et al. \cite{zabawa2020counting}. 
The counting is performed based on the masks, which are predicted with the convolutional neural network presented in their work. 
The classes \texttt{background} and \texttt{berry-edge} are discarded and the counting is solely performed with pixels of the class \texttt{berry}. 
Before counting the number of connected components of the berry mask, we introduce geometrical and qualitative filter stages to improve the count. 
First, we discard elements that are smaller than 25 pixels, since these artifacts are too small to represent berries.
Secondly, we exploit the knowledge that berries are roughly round by removing objects with a minor-major-axis ratio below 0.3 and an insufficient area. 
The actual area of each component is compared to the expected area based on a radius which is computed as the mean of the minor and major axis of the component. 
Lastly, we check how well each object is surrounded by an edge, since most high confidence predictions are well surrounded by an edge. 
For further details, we refer the reader to \cite{zabawa2019detection}.

% %%%%%%%%%%%%%%%%%%%%%%%%%%%%%%%%%%%%%%%%%%%%%%%%%%%%%%%%%%%%%%%%%%%%%%%%%%%%%%%%%%%%%%%%%%%%%%%%%%%%%%%%%%%%%%%%%%%%

\begin{figure}[t]
\captionsetup[subfigure]{justification=centering}
	\centering
	\subfloat[$\d{x}_{\text{occ}} - \widetilde{\d{x}}_{\text{occ}}$]{
        \includegraphics[trim= 11 10 0 0, clip, 
        width=0.245\textwidth]{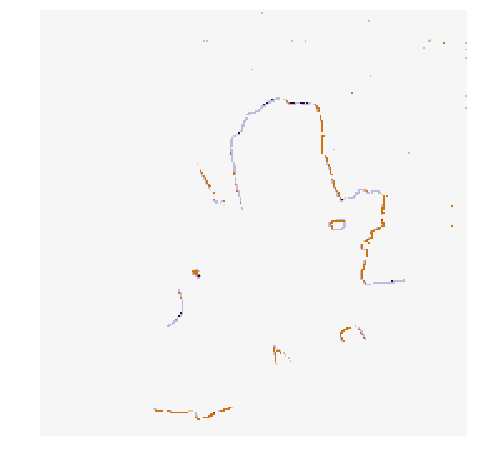} %2525
         \label{fig:generationMapping_A}}
%  	\captionsetup{justification=centering}
    \subfloat[$\d{x}_{\text{non}} - \widetilde{\d{x}}_{\text{non}}$ \\ (3 Classes)]{
        \includegraphics[trim=11 10 0 0, clip, 
        width=0.245\textwidth]{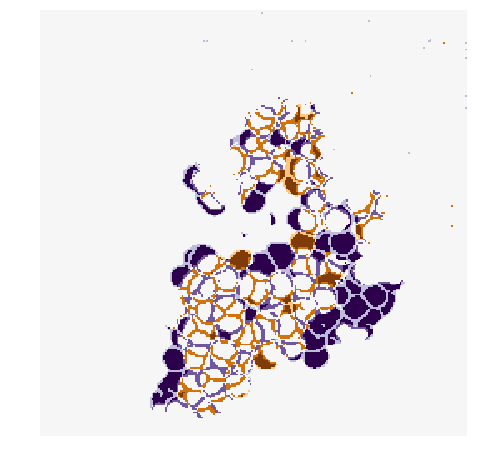}
         \label{fig:generationMapping_B}}
        \includegraphics[trim=340 10 0 0, clip,
        width=0.092\textwidth]{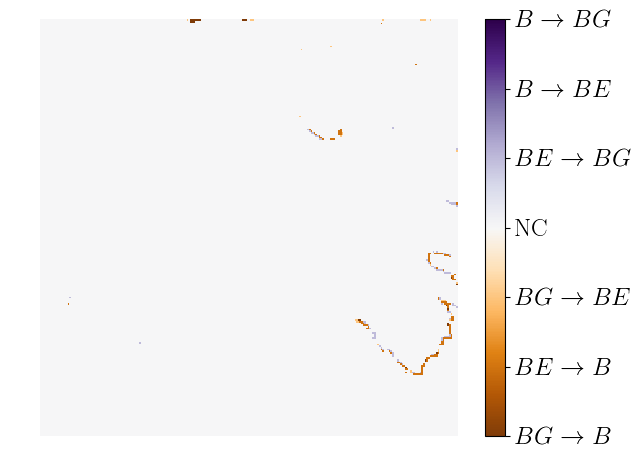}%094
% 	\captionsetup{justification=centering}
    \subfloat[$\d{x}_{\text{non}} - \widetilde{\d{x}}_{\text{non}}$ \\ (2 Classes)]{
        \includegraphics[trim=10 10 0 0, clip,
        width=0.245\textwidth]{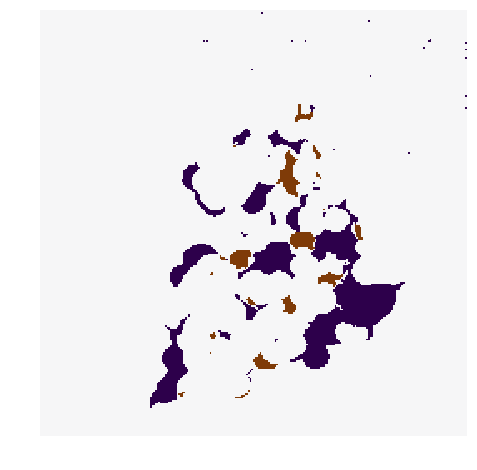}
         \label{fig:generationMapping_C}}
        \includegraphics[trim=340 10 0 0, clip,
        width=0.092\textwidth]{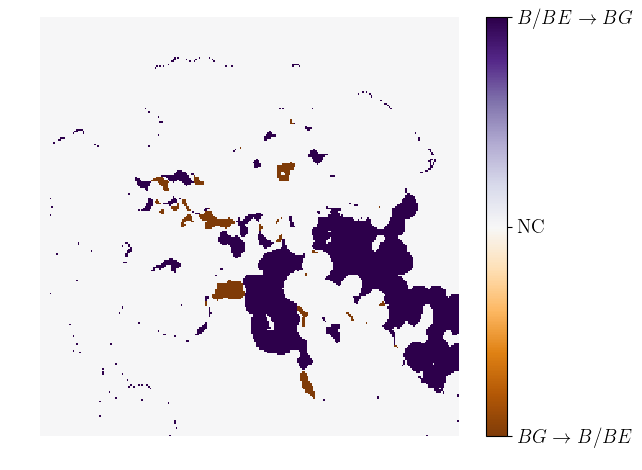}

	\caption{Example representation of generation maps in (a) the occluded domain, and (b) and (c) the non-occluded domain for the berry mask (b) and a binary mask (c) where classes \texttt{berry} (B) and \texttt{berry-edge} (BE) are combined to one class. \texttt{Background} is indicated by BG and if there is no class change it is indicated by NC for no change.}
	\label{fig:generationMapping2}
\end{figure}

Another metric we use for a visual comparison is the \emph{generation map}.
Generation mapping is used to visualize the differences between two masks. 
In our case the distances are calculated between (1) the input mask $\d{x}_{\text{occ}}$ and the generated mask $\widetilde{\d{x}}_{\text{occ}}$ (\figref{fig:generationMapping_A}), (2) the target output mask of $\d{x}_{\text{non}}$ and the generated mask of $\widetilde{\d{x}}_{\text{non}}$ (\figref{fig:generationMapping_B}) and lastly (3) the target output mask of $\d{x}_{\text{non}}$ and the generated mask of $\widetilde{\d{x}}_{\text{non}}$ including only 2 classes, where \texttt{berry} and \texttt{berry-edge} are considered as one class (\figref{fig:generationMapping_C}). 
We denote this mask as binary mask.

%ADD

The different colors allow us to make a statement about the area in which, for example, berries are generated where none are present in the reference.
The colors can be analyzed as follows:
For \figref{fig:generationMapping_A} and \figref{fig:generationMapping_B}, at pixel positions with a medium orange and medium blue discoloration, either the class berry is predicted to be an edge or the class edge is predicted to be a berry. 
These two cases are acceptable for our task, since we do not want to map the reference, but generate images, which provide very highly probable results with a distribution that matches the input.
The other pixel values are to be avoided, since at these positions for a light and dark orange discoloration the classes \texttt{berry} and \texttt{berry-edge} are generated, where in the reference \texttt{background} occurs. 
At the positions with a light and dark blue discolouration the class \texttt{background} is generated, where in the reference the class \texttt{berry} or \texttt{berry-edge} is present. 
The generation map, where only two classes are included, highlights the non acceptable pixel regions in the generated map.

% %%%%%%%%%%%%%%%%%%%%%%%%%%%%%%%%%%%%%%%%%%%%%%%%%%%%%%%%%%%%%%%%%%%%%%%%%%%%%%%%%%%%%%%%%%%%%%%%%%%%%%%%%%%%%%%%%%%%

\subsection*{Experiment 1 -- Comparison of generation quality based on LA and A data}
\label{sec:Experiment1}

With the first experiment, we want to analyze how the grayscale channel influences achieving our goals (i) to reproduce the hidden berries and thus (ii) improve the berry counting per image.
With the help of the grayscale channel L, it is possible to derive information about the presence of objects such as berries, leaves, and branches.
Theoretically, this information helps to identify positions in the image where berries might be generated, for example, behind leaves or branches. 
In practice, however, in the non-occluded reference, a part of the berries is not present, since a proportion of berries is still occluded due to leaves or bigger branches not being cut away. 
This makes training more difficult, since it is generally learned that new berries should not be generated at the position of branches that have not been cut away. 
This implies, we cannot expect to make new berries visible in the generated output $\widetilde{\d{x}}_{\text{non}}$ while testing, that are never present in the reference data $\d{x}_{\text{non}}$ of the training set. 

To get further insights into this, we will analyze whether ignoring the L channel will lead to a generation of berries in areas such as branches.
Moreover, we will investigate if using channel A only is better suited on natural data, because information such as color, exposure, and lighting conditions have no influence.
Thus, this experiment should determine whether the L channel adds value to the experiments and what this added value looks like.

%ADD
\begin{figure*}[t]
\renewcommand{\tabcolsep}{5pt}
\centering
    % \begin{tabular}{ccc}
	
% 	{
	\subfloat[Example 1]{
        \begin{minipage}{0.30\textwidth}
            \begin{minipage}{0.31\textwidth}
                \centering
        		$\d{x}_{\text{occ}}$
    		\end{minipage}
    		\begin{minipage}{0.31\textwidth}
        		\centering
        		 $\d{x}_{\text{non}}$
    		\end{minipage}
            \begin{minipage}{0.31\textwidth}
                \centering
                $\widetilde{\d{x}}_{\text{non}}$
    		\end{minipage}
    		
			\includegraphics[width=0.31\textwidth]{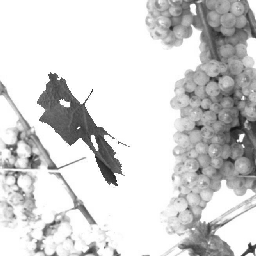}
			\includegraphics[width=0.31\textwidth]{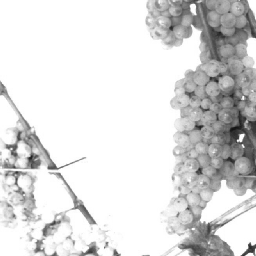}
			\includegraphics[width=0.31\textwidth]{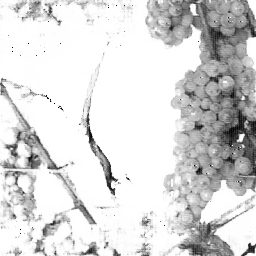}

			\includegraphics[width=0.31\textwidth]{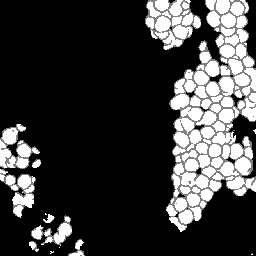}
			\includegraphics[width=0.31\textwidth]{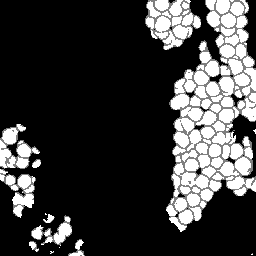}
			\includegraphics[width=0.31\textwidth]{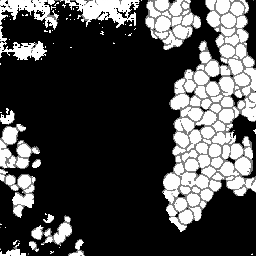}

    		\includegraphics[width=0.31\textwidth]{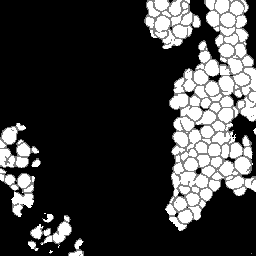}
            \includegraphics[width=0.31\textwidth]{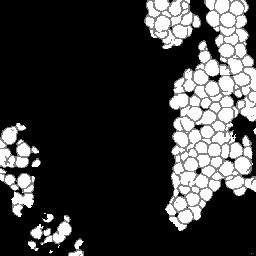}
    		\includegraphics[width=0.31\textwidth]{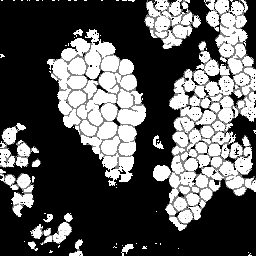}
    	\end{minipage}
    	\label{fig:visualComparisonLA2vsA2_a}}
    \hspace{3pt}
	\subfloat[Example 2]{
    	 \begin{minipage}{0.30\textwidth}
            \begin{minipage}{0.31\textwidth}
                    \centering
            		$\d{x}_{\text{occ}}$
        		\end{minipage}
        		\begin{minipage}{0.31\textwidth}
            		\centering
            		 $\d{x}_{\text{non}}$
        		\end{minipage}
                \begin{minipage}{0.31\textwidth}
                    \centering
                    $\widetilde{\d{x}}_{\text{non}}$
    		\end{minipage}
    		
			\includegraphics[width=0.31\textwidth]{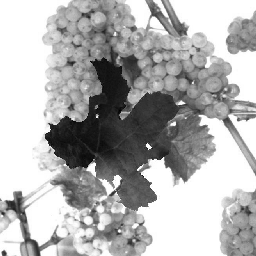}
			\includegraphics[width=0.31\textwidth]{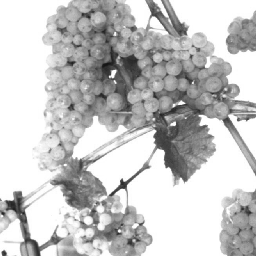}
			\includegraphics[width=0.31\textwidth]{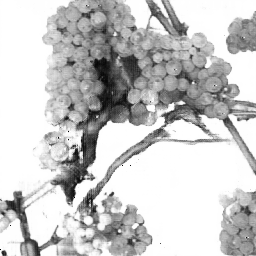}

			\includegraphics[width=0.31\textwidth]{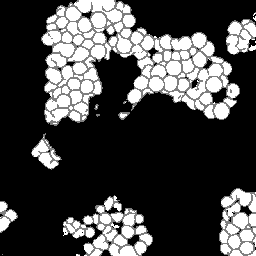}
			\includegraphics[width=0.31\textwidth]{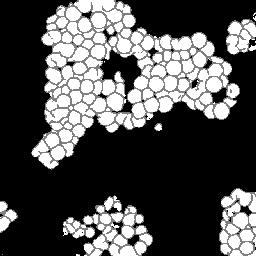}
			\includegraphics[width=0.31\textwidth]{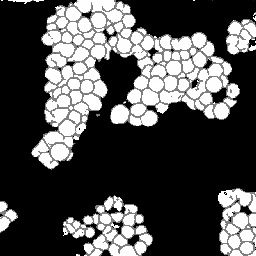}

			\includegraphics[width=0.31\textwidth]{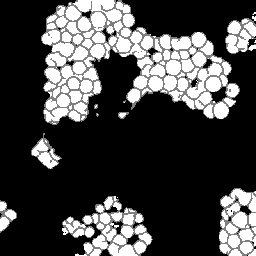}
			\includegraphics[width=0.31\textwidth]{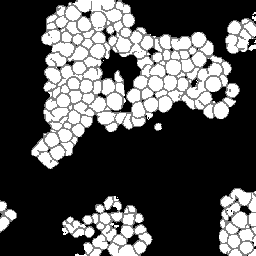}
			\includegraphics[width=0.31\textwidth]{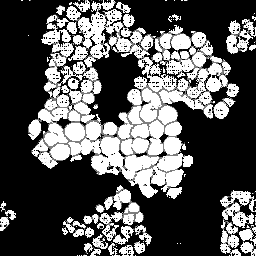}
    	\end{minipage}
    	\label{fig:visualComparisonLA2vsA2_b}}
    \hspace{3pt}
	\subfloat[Example 3]{
    	 \begin{minipage}{0.30\textwidth}
            \begin{minipage}{0.31\textwidth}
                \centering
        		$\d{x}_{\text{occ}}$
    		\end{minipage}
    		\begin{minipage}{0.31\textwidth}
        		\centering
        		 $\d{x}_{\text{non}}$
    		\end{minipage}
            \begin{minipage}{0.31\textwidth}
                \centering
                $\widetilde{\d{x}}_{\text{non}}$
		    \end{minipage}
		
			\includegraphics[width=0.31\textwidth]{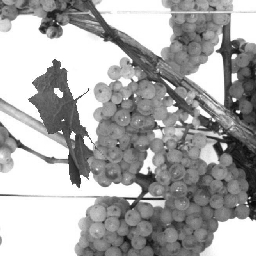}
			\includegraphics[width=0.31\textwidth]{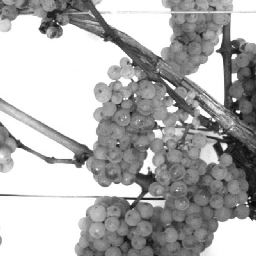}
			\includegraphics[width=0.31\textwidth]{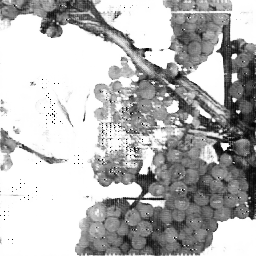}
			
			\includegraphics[width=0.31\textwidth]{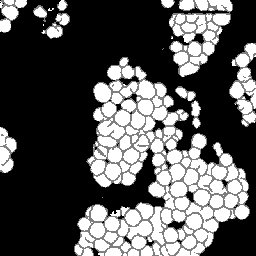}
			\includegraphics[width=0.31\textwidth]{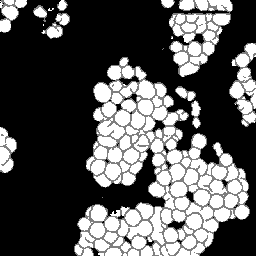}
			\includegraphics[width=0.31\textwidth]{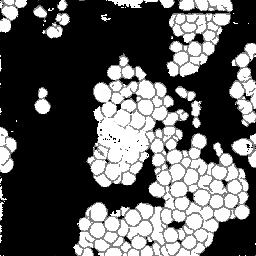}
			
			\includegraphics[width=0.31\textwidth]{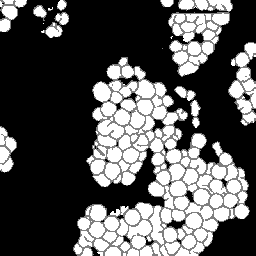}
			\includegraphics[width=0.31\textwidth]{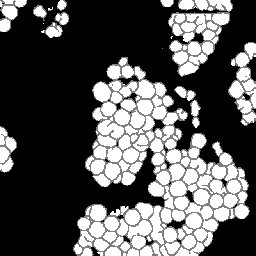}
			\includegraphics[width=0.31\textwidth]{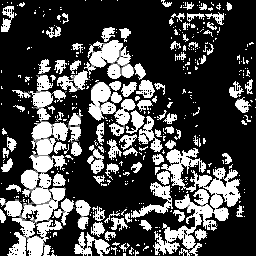}
    	\end{minipage}
    	\label{fig:visualComparisonLA2vsA2_c}}
	\caption{Visual representation of generated test results $\widetilde{\d{x}}_{\text{non}}$ in the non-occluded domain of \emph{Dataset 1} including LA channel and \emph{Dataset 2} including only A channel in comparison to reference target output $\d{x}_{\text{non}}$ of the non-occluded domain and input $\d{x}_{\text{occ}}$ of the non-occluded domain. The first row shows the L channel of $\mathcal{X}_{\text{LA}}$. The second row illustrates the mask A of $\mathcal{X}_{\text{LA}}$. The last row represents the corresponding result of $\mathcal{X}_{\text{A}}$.}
    \label{fig:visualComparisonLA2vsA2}
	\end{figure*}

\paragraph{Used data, model and evaluation metrics.}

For this experiment, we train a cGAN model on each of the training sets of \emph{Dataset 1} and \emph{Dataset 2}. 
The evaluation is based on the corresponding test sets. 
Since we want to determine the value of the L channel with this experiment, we limit the used data exclusively to defoliation type VSP. 
SMPH type shows proportionally similar outcomes to the VSP results.

To compare the two datasets $\mathcal X_{\text{A}}$ and ${\mathcal X}_{\text{LA}}$, we use the described metrics in Sec. \enquote{\nameref{sec:evalMetrics}}. %\secref{sec:evalMetrics}.
We compare the correlation and the IoU in the occluded domain between the input $\d{x}_{\text{occ}}$ and the generated input $\widetilde{\d{x}}_{\text{occ}}$, as well as in the non-occluded domain between the target output $\d{x}_{\text{non}}$ and the generated output $\widetilde{\d{x}}_{\text{non}}$ for both datasets.
The generated input $\widetilde{\d{x}}_{\text{occ}}$ is computed by taking the generated output $\widetilde{\d{x}}_{\text{non}}$ and occlude the same pixels in the berry mask which are occluded in the input by a synthetic leaf.

\paragraph{Results.}

\figref{fig:visualComparisonLA2vsA2} shows three example results to visually compare $\mathcal{X}_{\text{A}}$ and $\mathcal{X}_{\text{LA}}$. 
The first two columns of an example show the reference of the two domains, where the third column represents the generated output $\widetilde{\d{x}}_{\text{non}}$.
The first row shows the grayscale channel of LA, the second row shows the mask channel of LA, and the bottom row shows the mask channel of A.

Using data without the L channel leads to higher generalizability regarding different varieties such as color, lighting conditions, and occlusions.
Remarkable for the mask of A (row 3) is that for input patches containing many berries, proportionally too large and therefore too few berries are generated in $\widetilde{\d{x}}_{\text{non}}$ of the test results. 
This applies to the entire dataset and is demonstrated by \figref{fig:visualComparisonLA2vsA2_a} and \figref{fig:visualComparisonLA2vsA2_b}. 
Generated berries in $\widetilde{\d{x}}_{\text{non}}$ of $\mathcal{X}_{\text{LA}}$ adapt better to existing berries in mask $\d{x}_{\text{occ}}$ than of $\mathcal{X}_{\text{A}}$. 
Furthermore, it turns out that the model trained on $\mathcal{X}_{\text{A}}$ has problems in generating patches with many berries. 
The berries are not only too big, but also in general berries are difficult to represent in their shape, as seen in \figref{fig:visualComparisonLA2vsA2_b} and \figref{fig:visualComparisonLA2vsA2_c}.

Another positive aspect of $\mathcal{X}_{\text{LA}}$ is the already mentioned point that background information of the grayscale patch is included in the generation of new berries. 
The model learns to recognize where background is present in the patch and thus does not generate new berries in $\widetilde{\d{x}}_{\text{non}}$ in contrast to the model trained on $\mathcal{X}_{\text{A}}$. 
This is particularly obvious in Example 1, see \figref{fig:visualComparisonLA2vsA2_a}, where a whole grape is generated in the center of the mask. 
In the reference input and output of L, it is visible that on this position, background occurs.

%ADD
\begin{figure}[t]
	\centering
	\subfloat[$x_{\text{occ}}$ vs. $\widetilde{\d{x}}_{\text{occ}}$]{
    \includegraphics[
        width=0.5\textwidth]{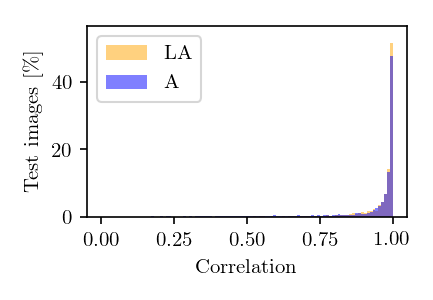}
        \label{fig:correlationLAvsA_realA_fakeA}}
    \subfloat[$x_{\text{non}}$ vs. $\widetilde{\d{x}}_{\text{non}}$]{
    \includegraphics[
        width=0.5\textwidth]{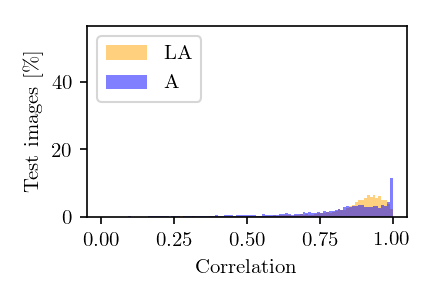}
      \label{fig:correlationLAvsA_realB_fakeB}}
      
	\caption{Comparison of LA and A data with respect to (a) the correlation between input patch $\d{x}_{\text{occ}}$ and generated input patch $\widetilde{\d{x}}_{\text{occ}}$ in the occluded domain, and (b) the correlation between target output $\d{x}_{\text{non}}$ and generated output $\widetilde{\d{x}}_{\text{non}}$ in the non-occluded domain. Shown is the percentage of test images that are assigned to a specific range of correlation. One bar corresponds to the range of 0.01.}
	\label{fig:correlation_real_fake}
\end{figure}

% correlation
In the following, we will take a look at the objective metrics described above. 
If we compare them regarding the $\mathcal{X}_{\text{LA}}$ and $\mathcal{X}_{\text{A}}$ input, we notice that the results for correlation between $\d{x}_{\text{occ}}$ and $\widetilde{\d{x}}_{\text{occ}}$ are similar (see  \figref{fig:correlationLAvsA_realA_fakeA}). 
For $\mathcal{X}_{\text{A}}$, there are more generated patches with a correlation smaller than 0.8 and, therefore, less with a higher correlation.
The correlation histogram between $\d{x}_{\text{non}}$ and $\widetilde{\d{x}}_{\text{non}}$, shown in \figref{fig:correlationLAvsA_realB_fakeB}, shows different distributions for the dataset. 
While the correlation histogram of LA, presented in orange, shows a left-skewed distribution, the amount of test patches of A increases on average with increasing correlation. 
At a correlation in the interval of (0.99, 1], represented by the right bar, the distribution shows a striking peak. However, there is a larger proportion of values below a correlation of 0.85. Even in the interval (0.85, 0.99], the percentages of patches for LA are higher than for A. This means that with the goal likely to generate results with a distribution that matches the input, rather than the exact image content of each image, \emph{dataset 1} leads to better results on average.

\begin{figure}[t]
% \captionsetup[subfigure]{justification=centering}
	\centering
    \subfloat[Counting in occluded domain][Counting in occluded \\domain.]{
    \includegraphics[trim=8 0 8 0, clip,
            width=0.26\textwidth]{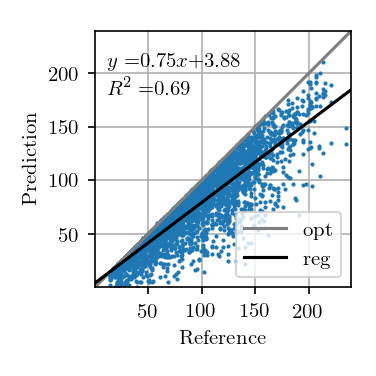}
    \label{fig:R2_Exp4_xOcc_a}}
    \subfloat[Counting in non-occluded domain with input A.][Counting in non-occluded domain with \\input A.]{
    \includegraphics[trim=8 0 8 0, clip,
            width=0.26\textwidth]{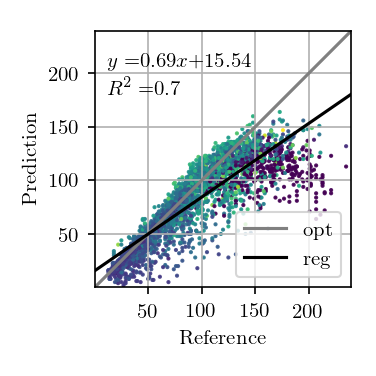}
            \label{fig:R2_Exp4_xNon_A}}
    \includegraphics[trim=134 0 5 8, clip,
            width=0.0675\textwidth]{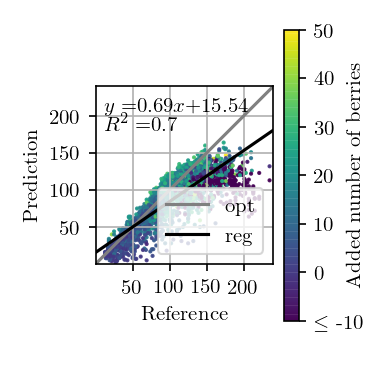}
    \subfloat[Counting in non-occluded domain][Counting in non-occluded domain with \\input LA.]{
    \includegraphics[trim=8 0 8 0, clip,
            width=0.26\textwidth]{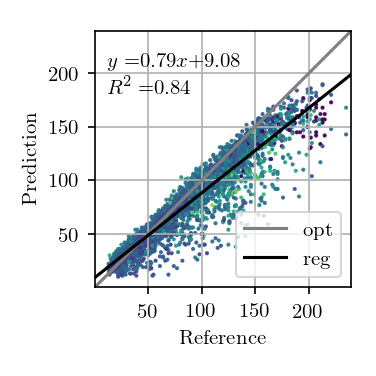}
    \label{fig:R2_Exp4_xNon_b}}
    \includegraphics[trim=134 0 5 8, clip,
            width=0.0675\textwidth]{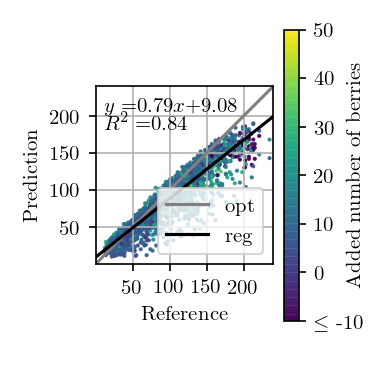}
  
	\caption{Counting of the berries in (a) the input of the occluded domain and of generated berries in ((b), (c)) the non-occluded domain of VSP defoliated data visualized by a $R^2$-Plot. 
	Shown is the relation between the generated output $\widetilde{\d{x}}_{\text{non}}$ in relation to reference $\d{x}_{\text{non}}$ for data with input channel (b) A and (c) LA. The reference is represented by the grey line. The black line represents the regression line adapted to the predictions with the corresponding regression equation at the top left of the plot. The colouration of the data points in the plots (b) and (c) indicates the added number of berries compared to the non-occluded domain.
    }
	\label{fig:R2_counting_xNon}
\end{figure}

% counting
\figref{fig:R2_Exp4_xNon_A} and \figref{fig:R2_Exp4_xNon_b} present a counting comparison of the different models in the non-occluded domain using a $R^2$-Plot. 
Additionally, \figref{fig:R2_Exp4_xOcc_a} shows the counting results without domain-transfer, i.e. no additional generated berries. 
Counting applied to the target $\d{x}_{\text{non}}$ in the non-occluded domain serves as the counting reference and is represented by the diagonal gray line. 
We observe that the results with input LA give the best matched results with respect to the reference. This is indicated visually as well as by the $R^2$ value of the different models which is the highest for our approach in the non-occluded domain with input LA. 
As in the visual evaluation, the counting plot for input A in \figref{fig:R2_Exp4_xNon_A} shows that the model indicates problems generating berries with a larger number of berries per patch. 
Also in the LA results, we observe that, especially with a reference number of more than 150 berries, the model does not reach the reference. 
This is explained by the fact that there are relatively few patches in the dataset with a number greater than 150 compared to the number of patches containing less than 160 berries. 
More detailed analyses of the berry counting can be found in Sec. \hyperref[sec:Experiment4]{\enquote{Experiment 4}}. %experiment 4 \secref{sec:Experiment4}.
Based on the observations and results from this experiment, we decide to use \emph{Dataset 1} including the L channel for our further experiments.

%ADD

% %%%%%%%%%%%%%%%%%%%%%%%%%%%%%%%%%%%%%%%%%%%%%%%%%%%%%%%%%%%%%%%%%%%%%%%%%%%%%%%%%%%%%%%%%%%%%%%%%%%%%%%%%%%%%%%%%%%%

\subsection*{Experiment 2 -- Real vs. generated results in the occluded domain} 
\label{sec:Experiment2}

In this experiment, we investigate whether the regions showing berries in the occluded domain stay unchanged in the transferred non-occluded domain.
Furthermore, we verify that new berries are generated exclusively in the occluded area, and thus, the model detects where the appearance of berries is very likely.

\paragraph{Used data, model and evaluation metrics.}

For this experiment, we use the synthetic \emph{Dataset 1} of the VSP defoliation, which is described in Sec. \enquote{\nameref{sec:synthData}}. %\secref{sec:synthData}. 
For evaluation, we use different masks: The first mask is the so-called generated input mask $\widetilde{\d{x}}_{\text{occ}}$, for which we take the generated output $\widetilde{\d{x}}_{\text{non}}$ of the test set and overlay it with the leaf used for data augmentation of the synthetic input $\d{x}_{\text{occ}}$. 
The other mask is the so-called baseline mask $\d{x}_{\text{non},\text{leaf}}$ of this experiment. 
For this purpose, we use the target output $\d{x}_{\text{non}}$ and overlay it likewise with the leaf used for data augmentation of the synthetic input $\d{x}_{\text{occ}}$. 
Thus, only the non-occluded pixels of $\d{x}_{\text{occ}}$ will remain visible in $\widetilde{\d{x}}_{\text{occ}}$ and $\d{x}_{\text{non},\text{leaf}}$. 
The evaluation is then performed on the pairs $\lbrace \d{x}_{\text{occ}}, \d{x}_{\text{occ},\text{leaf}} \rbrace$ and $\lbrace \d{x}_{\text{occ}}, \widetilde{\d{x}}_{\text{occ}} \rbrace$. 

\begin{figure}[h!]
	\centering
	    \vspace{-14pt}
	\rotatebox{90}{\large \parbox{3cm}{$\d{x}_{\text{occ,L}}$\\ }}
	\hspace{2pt}
    \includegraphics[
    width=0.143\textwidth]{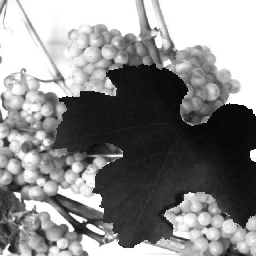}
    \hspace{3pt}
     \includegraphics[
    width=0.143\textwidth]{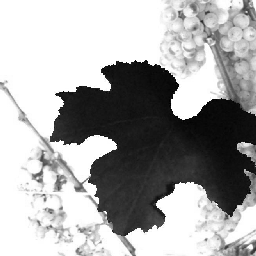}
    \hspace{3pt}
    \includegraphics[
    width=0.143\textwidth]{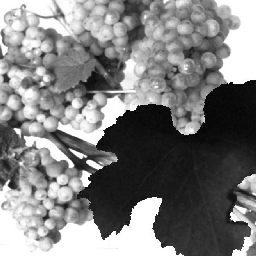}
    \hspace{3pt}
  \includegraphics[
    width=0.143\textwidth]{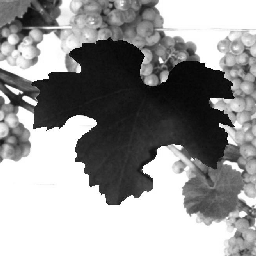}
    \hspace{3pt}
    \includegraphics[
    width=0.143\textwidth]{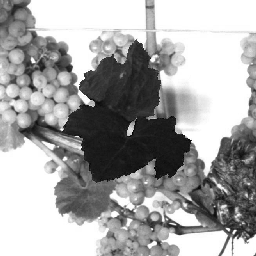}
    \hspace{1pt}
    \includegraphics[trim=335 10 0 0, clip,
    width=0.065\textwidth]{bar}
    \vspace{-14pt}
	\rotatebox{90}{\large \parbox{3cm}{$\d{x}_{\text{occ}}- \d{x}_{\text{non,\text{leaf}}}$\\ }}
    \includegraphics[trim=10 10 0 0, clip,
    width=0.16\textwidth]{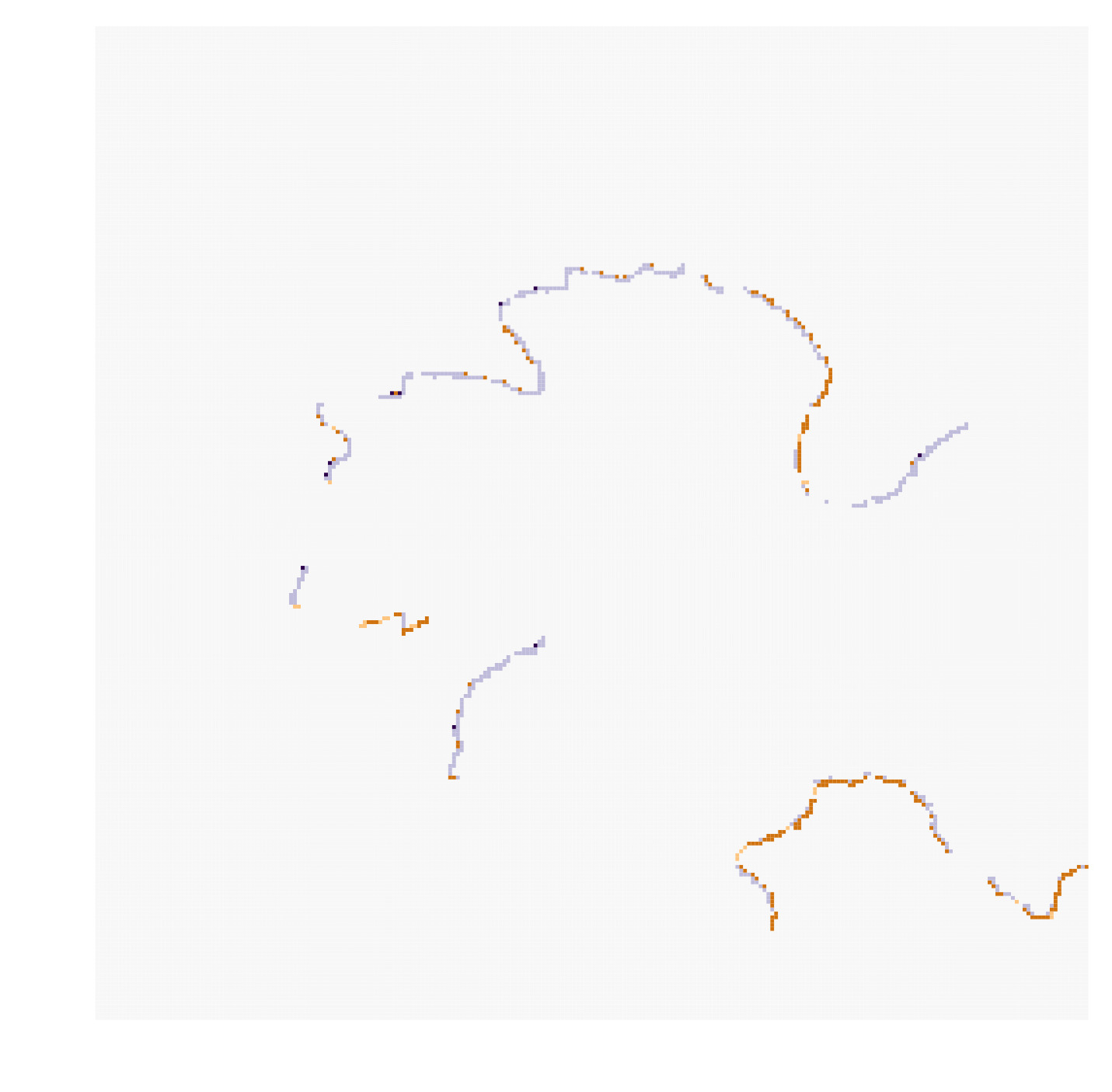}
     \includegraphics[trim=10 10 0 0, clip,
    width=0.16\textwidth]{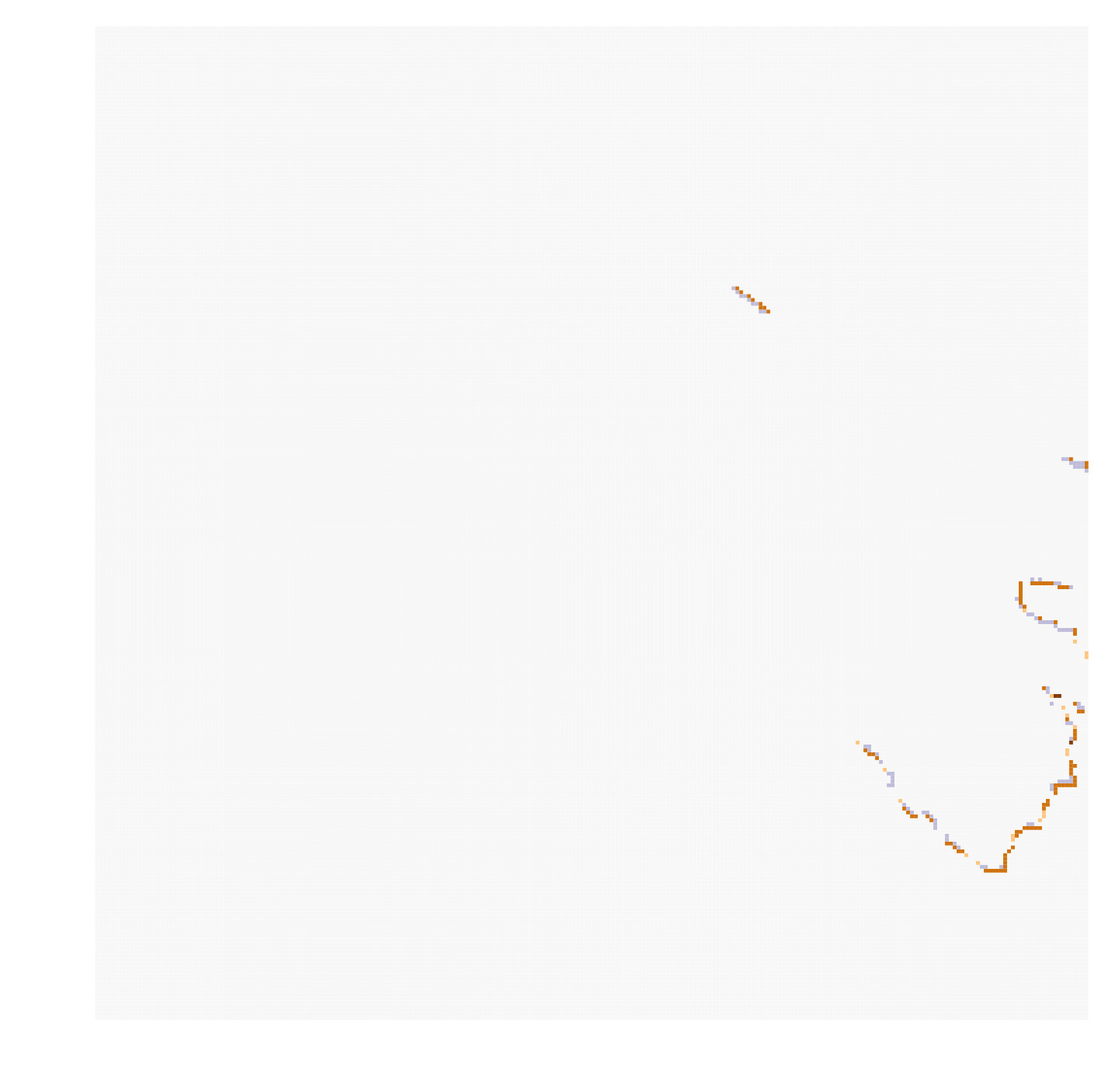}
    \includegraphics[trim=10 10 0 0, clip,
    width=0.16\textwidth]{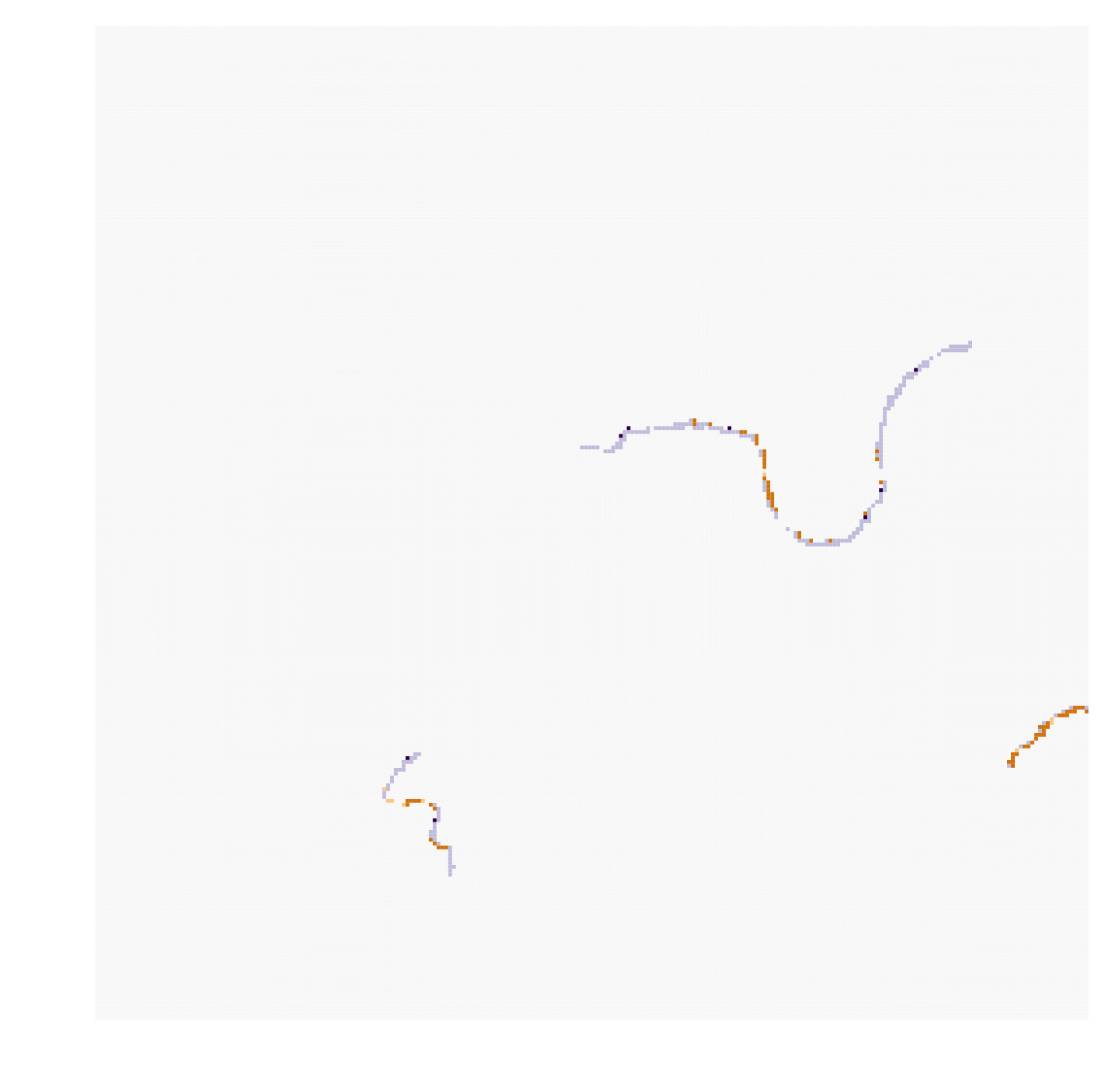}
  \includegraphics[trim=10 10 0 0, clip,
    width=0.16\textwidth]{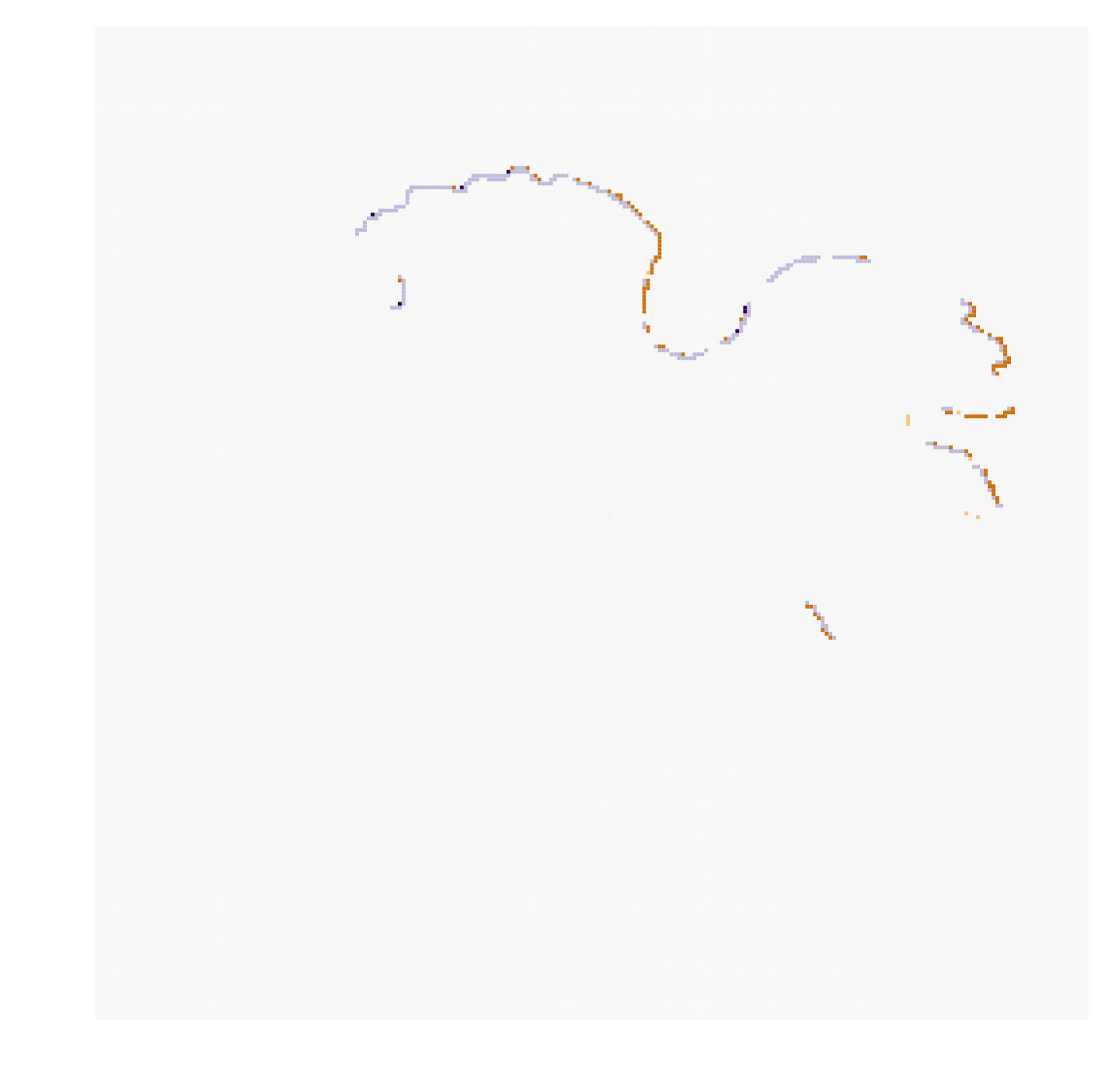}
    \includegraphics[trim=10 10 0 0, clip,
    width=0.16\textwidth]{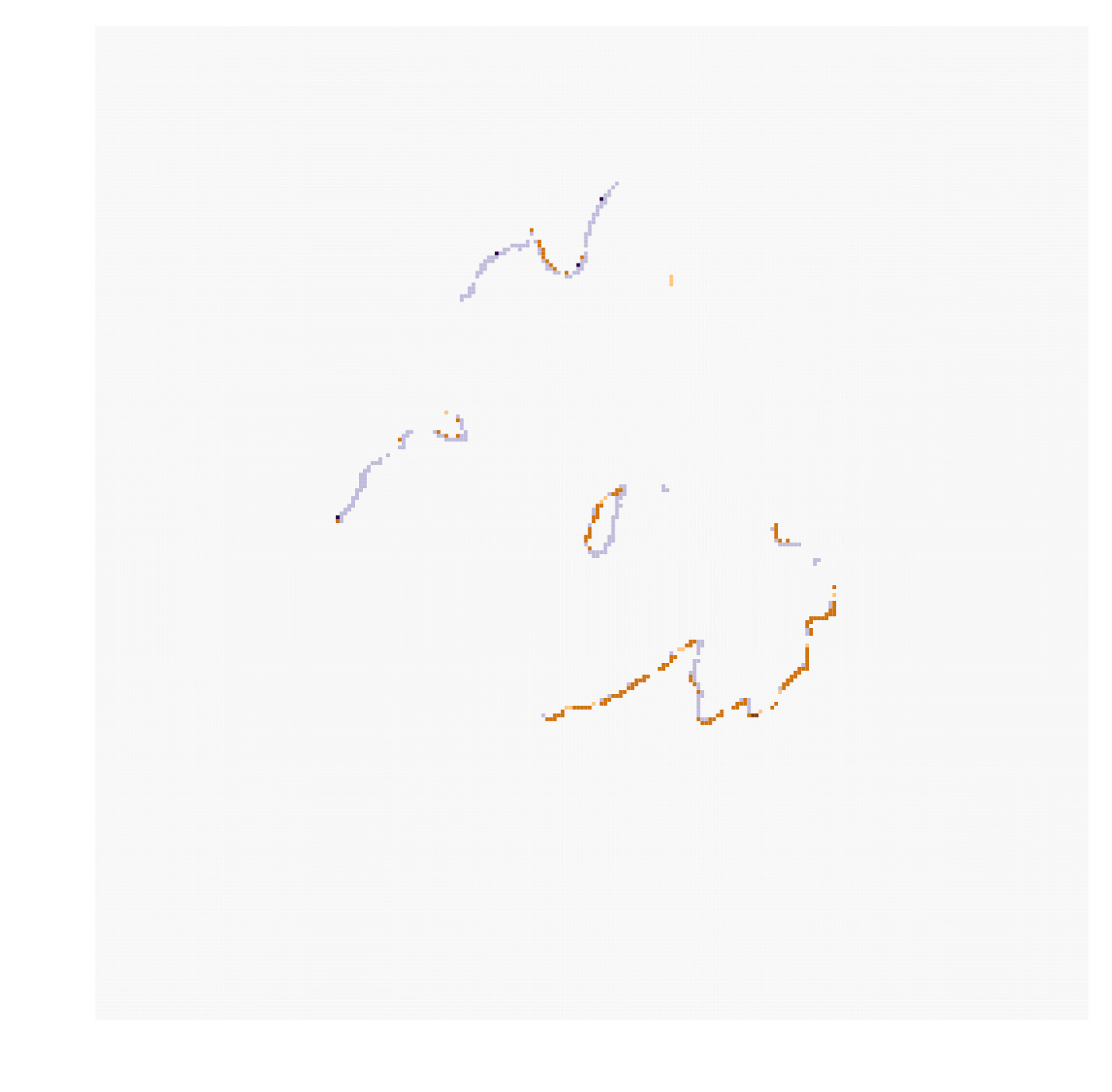}
    \includegraphics[trim=335 10 0 0, clip,
    width=0.065\textwidth]{bar}
    \vspace{-14pt}
    \rotatebox{90}{\large \parbox{3cm}{$\d{x}_{\text{occ}} - \widetilde{\d{x}}_{\text{occ}}$\\ }}
    \includegraphics[trim=10 10 0 0, clip,
    width=0.16\textwidth]{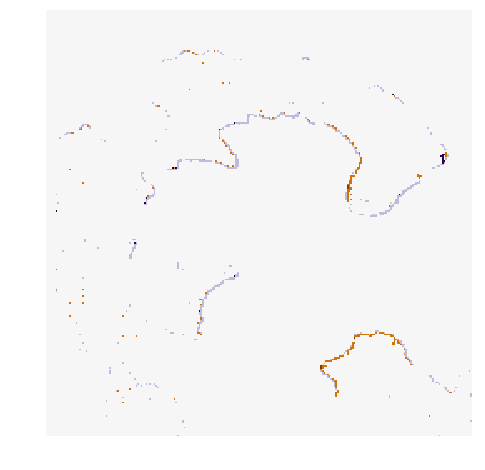}
     \includegraphics[trim=10 10 0 0, clip,
    width=0.16\textwidth]{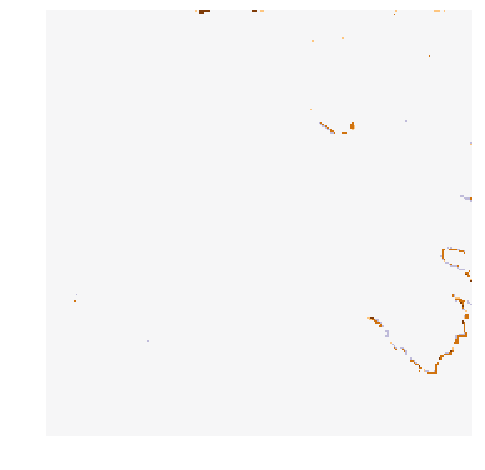}
    \includegraphics[trim=10 10 0 0, clip,
    width=0.16\textwidth]{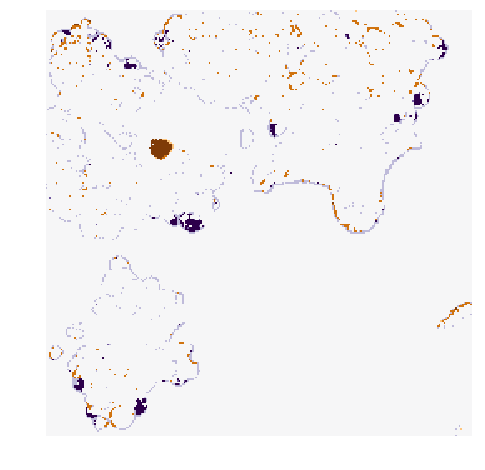}
  \includegraphics[trim=10 10 0 0, clip,
    width=0.16\textwidth]{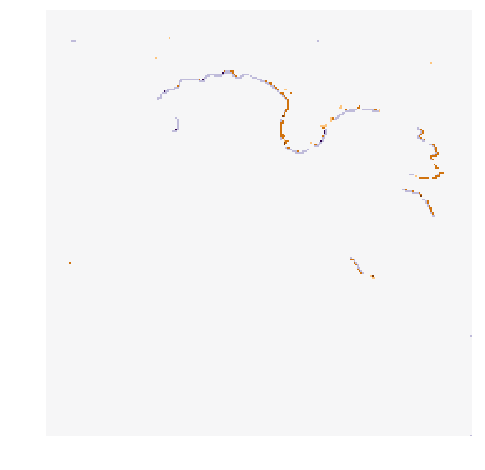}
    \includegraphics[trim=10 10 0 0, clip,
    width=0.16\textwidth]{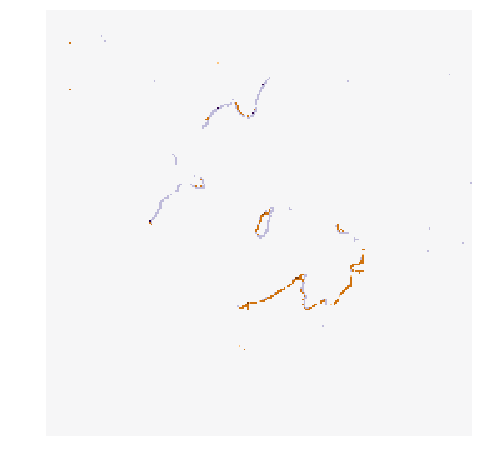}
    \includegraphics[trim=335 10 0 0, clip,
    width=0.065\textwidth]{bar.png}
    \vspace{-14pt}
    \rotatebox{90}{\large  \parbox{3cm}{ $\d{x}_{\text{non}} - \widetilde{\d{x}}_{\text{non}}$ \\ (3 Classes) } }
    \includegraphics[trim=10 10 0 0, clip,
    width=0.16\textwidth]{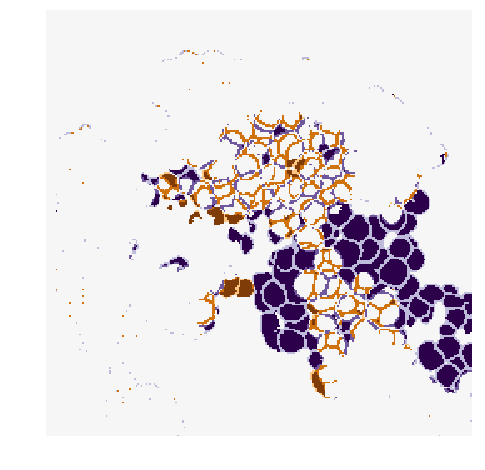}
    \includegraphics[trim=10 10 0 0, clip,
    width=0.16\textwidth]{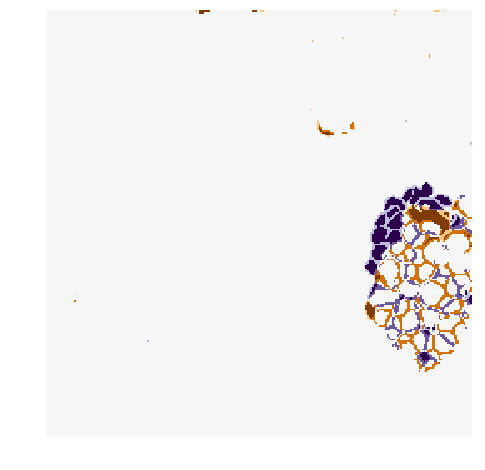}
    \includegraphics[trim=10 10 0 0, clip,
    width=0.16\textwidth]{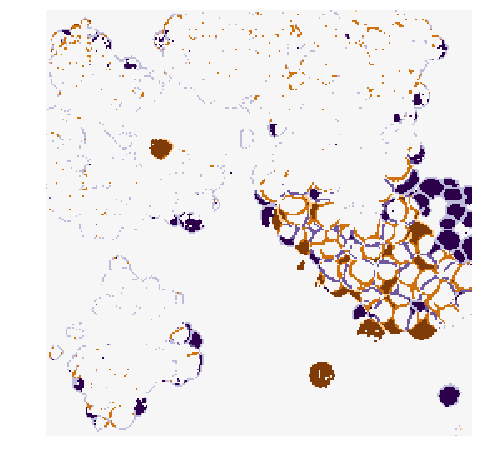}
    \includegraphics[trim=10 10 0 0, clip,
    width=0.16\textwidth]{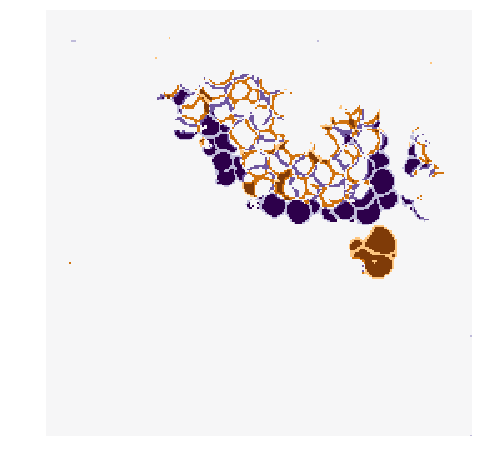}
    \includegraphics[trim=10 10 0 0, clip,
    width=0.16\textwidth]{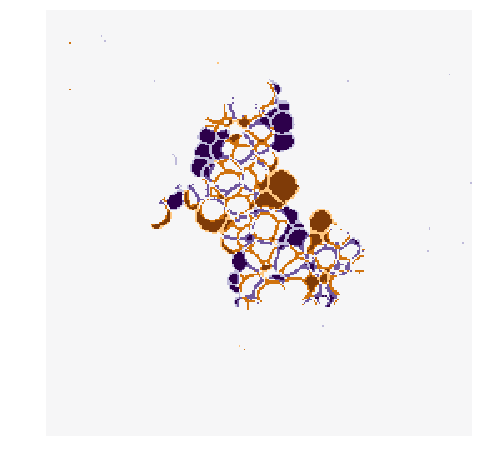}
    \includegraphics[trim=335 10 0 0, clip,
    width=0.065\textwidth]{bar}
    % \vspace{-15pt}
    
    \rotatebox{90}{\large \parbox{3cm}{ $\d{x}_{\text{non}} - \widetilde{\d{x}}_{\text{non}}$ \\ (2 Classes) } }
    \includegraphics[trim=10 10 0 0, clip,
    width=0.16\textwidth]{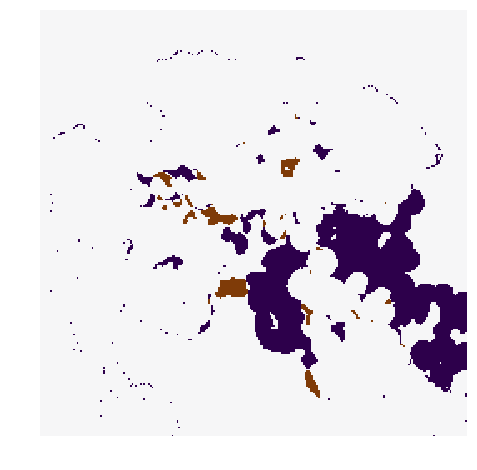}
    \includegraphics[trim=10 10 0 0, clip,
    width=0.16\textwidth]{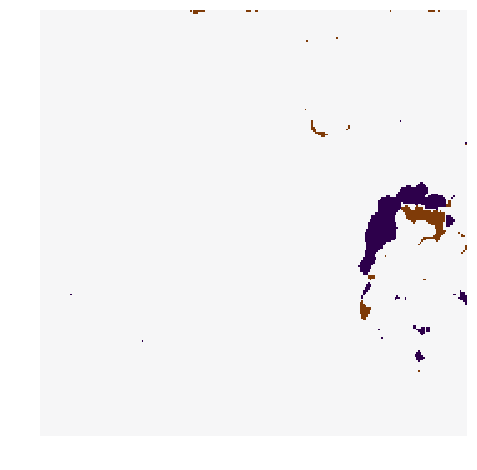}
    \includegraphics[trim=10 10 0 0, clip,
    width=0.16\textwidth]{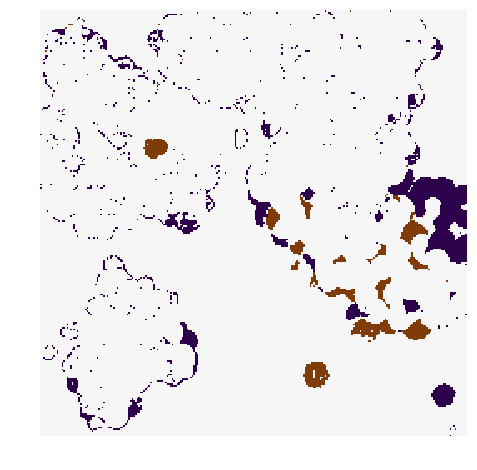}
    \includegraphics[trim=10 10 0 0, clip,
    width=0.16\textwidth]{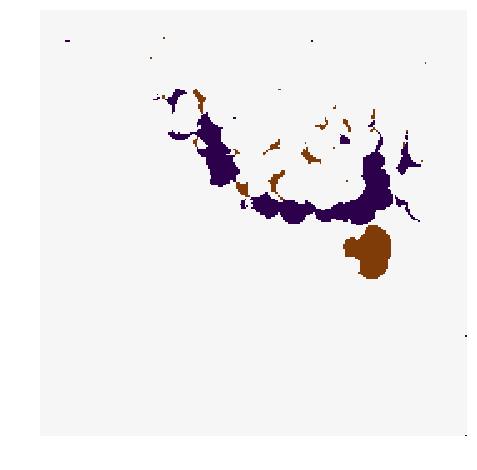}
    \includegraphics[trim=10 10 0 0, clip,
    width=0.16\textwidth]{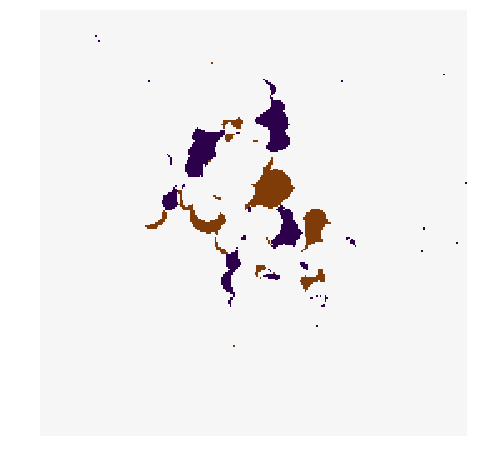}
    \includegraphics[trim=335 10 0 0, clip,
    width=0.064\textwidth]{bar_2Ch}
    
	\caption{Generation maps between target berry masks and generated output berry masks, as described in Sec. \enquote{\nameref{sec:evalMetrics}}. %\secref{sec:evalMetrics}. 
	The first row illustrates the respective grayscale input image.
	The second row shows the input $\d{x}_{\text{occ}}$ compared to the target output occluded by the leaf used for creating the input $\d{x}_{\text{non},\text{leaf}}$. The third row shows the maps in the occluded domain. The input $\d{x}_{\text{occ}}$ is compared to the generated output occluded by the leaf used for creating the input $\widetilde{\d{x}}_{\text{occ}}$. Second and third row occur in the occluded domain. The fourth and fifth row show the maps of the non-occluded domain. The target output $\d{x}_{\text{non}}$ is compared to the generated output $\widetilde{\d{x}}_{\text{non}}$. In the fourth row all three classes are included. The last row illustrate the same, but including only 2 classes. Classes \texttt{berry-edge} (BE) and \texttt{berry} (B) are combined in one class. \texttt{Background} is indicated by BG and if there is no class change it is indicated by NC for no change.}
	\label{fig:Heatmap_A_realA_fakeA}
\end{figure}

We use IoU and correlation as comparative metrics for this experiment.
Additionally, we create generation maps which show the differences between the masks within each of the pairs $\lbrace \d{x}_{\text{occ}}, \d{x}_{\text{occ},\text{leaf}} \rbrace$ and $\lbrace \d{x}_{\text{occ}}, \widetilde{\d{x}}_{\text{occ}} \rbrace$, as illustrated in \figref{fig:Heatmap_A_realA_fakeA}. 
For this experiment, the first three rows are of interest to us. 
The first row shows the respective grayscale patch of the generation maps. 
The second row shows the differences within the pair $\lbrace \d{x}_{\text{occ}}, \d{x}_{\text{occ},\text{leaf}} \rbrace$. Row three shows the differences within the pair $\lbrace \d{x}_{\text{occ}}, \widetilde{\d{x}}_{\text{occ}} \rbrace$. 
The columns indicate different patch examples.

\paragraph{Results.}

%ADD

The reference correlation within the pair $\lbrace \d{x}_{\text{occ}}, \d{x}_{\text{occ},\text{leaf}} \rbrace$ is above 0.98 for all test patches.
With our method, we manage to achieve a correlation of over 0.98 within the pair $\lbrace \d{x}_{\text{occ}}, \widetilde{\d{x}}_{\text{occ}} \rbrace$ for about 65\% of the test images, see \figref{fig:correlationLAvsA_realA_fakeA} (orange). 
The remaining 35\% are largely distributed over a correlation within the interval [0.75, 0.98). %[0.75,0.85) ?? warum 85?
The correlation strongly correlates with the IoU calculation of the \texttt{berry} area.
The low correlations are either due to artifacts in the generated masks or to test images with a high number of berries. In this case, the model do not transfer all non-occluded pixels one to one into the non-occluded domain. 
The effect of the amount of berries in the patch is shown in the generation maps in \figref{fig:Heatmap_A_realA_fakeA} $\lbrace$column 1, row 3$\rbrace$ and
$\lbrace$column 3, row 3$\rbrace$.
These observations can be traced back to the test results, which not only show the class values 0, 127, and 255 within the mask, but also pixel with values in between. 
This means that the model does not clearly assign the respective pixel to a class. 
At this point, we apply data post-processing to our generated data, as described in Sec. \enquote{\nameref{sec:dataPostProcessing}}. %\secref{sec:dataPostProcessing}. 
Pixel in areas of class boundaries are particularly affected here, which is why the differences arise in these areas.

For the patch examples in columns 2, 4 and 5, the generation maps of the pairs $\lbrace \d{x}_{\text{occ}}, \d{x}_{\text{occ},\text{leaf}} \rbrace$ are almost identical to the generation maps of the pairs $\lbrace \d{x}_{\text{occ}}, \widetilde{\d{x}}_{\text{occ}} \rbrace$. 
Such maps correspond to correlation values close to 1. 
It is noticeable that in all five examples, the border of the leaf used for data augmentation is highlighted in the generation maps. 
The colouring occurs at transitions between the leaf and the adjacent \texttt{berry-edge} and \texttt{berry} pixel.
At these positions, an additional edge was added when the synthetic occluded input mask was created, which has a width of about three pixels. 
The masks $\widetilde{\d{x}}_{\text{non}}$ and $\d{x}_{\text{non}}$, on which the masks $\widetilde{\d{x}}_{\text{occ}}$ and $\d{x}_{\text{occ},\text{leaf}}$ used in this experiment are based on, show a continuation of the depicted grape branches exactly at these transitions. 
This results in variations between the paired masks at this location.

The key findings from this experiment are that despite individual deviations, the visible part of the mask of the occluded domain is safely transferred to the non-occluded domain and stays unchanged. 
We assume that the model will make no result-altering changes.

% %%%%%%%%%%%%%%%%%%%%%%%%%%%%%%%%%%%%%%%%%%%%%%%%%%%%%%%%%%%%%%%%%%%%%%%%%%%%%%%%%%%%%%%%%%%%%%%%%%%%%%%%%%%%%%%%%%%%

\subsection*{Experiment 3 -- Real vs. generated results in the non-occluded domain}
\label{sec:Experiment3}

In this experiment, we investigate the similarity of our generated output $\widetilde{\d{x}}_{\text{non}}$ compared to the target output $\d{x}_{\text{non}}$ in the non-occluded domain. 

%ADD
\begin{figure}[t]
% \captionsetup[subfigure]{justification=centering}

	\centering
	\subfloat[]{
     \includegraphics[
    width=0.495\textwidth]{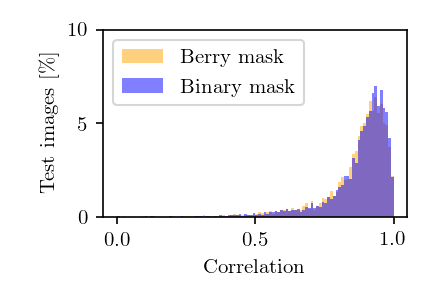}
    \label{fig:Corr_fakeB_alles_2ClVs3Cl}}
    \subfloat[]{
     \includegraphics[
    width=0.495\textwidth]{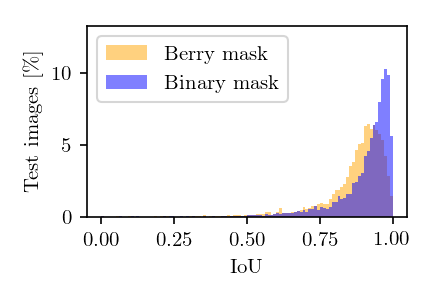}
    \label{fig:IoU_fakeB_2ClVs3Cl}}
	\caption{The graphs show a comparison within the non-occluded domain between berry mask and binary mask including only two classes for the metrics (a) correlation between $\d{x}_{\text{non}}$ and $\widetilde{\d{x}}_{\text{non}}$ and (b) IoU of the \texttt{berry} pixel in $\d{x}_{\text{non}}$ and $\widetilde{\d{x}}_{\text{non}}$.}
	\label{fig:Corr_IoU_fakeB_alles_2ClVs3Cl}
\end{figure}

\paragraph{Used  data,  model  and  evaluation  metrics.}
In this experiment, \emph{Dataset 1} is used to train the model.
Since we are aiming only for a highly probable result rather than the exact position and shape of specific berries, for our evaluation, we additionally create a binary mask based on the berry mask, which includes only the classes  \texttt{berry} and \texttt{background}. 
For this, we merge the classes \texttt{berry} and \texttt{berry-edge}.
We compare the mask pair $\lbrace \d{x}_{\text{non}}, \widetilde{\d{x}}_{\text{non}} \rbrace$ of the non-occluded domain in respect to the berry and binary mask.
We evaluate the correlation and IoU within this pair. Furthermore, we create generation maps that illustrate the difference between this pair. 
Exclusively for the berry mask, we calculate the area and diameter of all individual berries in the entire test data set.

\paragraph{Results.}

The correlation (\figref{fig:Corr_fakeB_alles_2ClVs3Cl}) shows a similar left-skewed distribution for berry mask and binary mask. The majority of the test images show a correlation of above 0.8. 
Although our approach does not aim to generate the exact position and shape of berries, the results indicate that the similarity of the generated results and the reference are high.
The IoU in \figref{fig:IoU_fakeB_2ClVs3Cl} also supports this finding. 
The IoU of the binary mask has on average higher values and is closer to the possible maximum than the berry mask.
This indicates a similar position of the grape bunches independent of the berry objects in the generated result compared to the reference. Rather, the berry objects and the corresponding berry-edges do not match one-to-one. 
The generation maps from \figref{fig:Heatmap_A_realA_fakeA} also show this property in the fourth and fifth row. 
The fourth row shows example results for the berry mask, where two cases can be seen.  
\emph{Case 1}: The medium orange and medium blue colours in fourth row illustrate pixels where the classes \texttt{berry} and \texttt{berry-edge} are confused. 
This incorrect generation is acceptable due to the desired property of likely results instead of non exactly matching results. \emph{Case 2}: Dark and light blue, and dark and light orange are incorrectly generated classes that need to be avoided.
In the fifth row, these pixel regions are highlighted by dark blue and dark orange. 
These regions either represent berries where there are no berries in the reference or vice versa. 
Such incorrect generations shift the position and size of the grape bunches. 
In the example maps, however, it can be seen that \emph{Case 1} occurs predominantly. 
It is obvious that berries are predicted in the right areas, but their shape and position do not correspond exactly to the reference. 

%ADD
\begin{figure}[t]
	\centering
    \subfloat[]{
      \includegraphics[%trim=0 0 10 0, clip,
    width=0.49\textwidth]{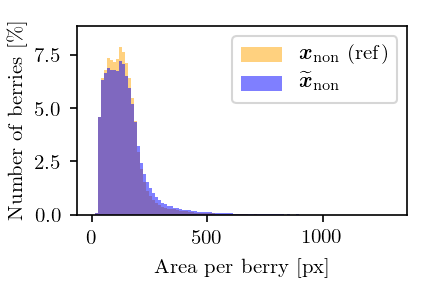}
    \label{fig:Area_RefVsLA}
    }
    \subfloat[]{
      \includegraphics[%trim=0 0 15 0, clip,
    width=0.49\textwidth]{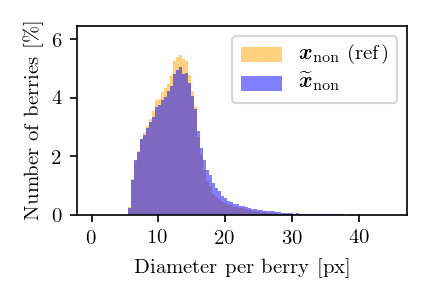}
    \label{fig:Diameter_RefVsLA}
    }
	\caption{Comparison of (a) area and (b) diameter per berry between target output $\d{x}_{\text{non}}$ and generated output $\widetilde{\d{x}}_{\text{non}}$ in the non-occluded domain. Only areas up to $\SI{1300}{\px}$ and diameters up to $\SI{45}{\px}$ are plotted.}
	\label{fig:Diameter_Area_RefVsLA}
\end{figure}

At the transition from image areas with berries to \texttt{background} pixels, the second case occurs where too small or too large grape bunches are produced, because either too few or too many berries are generated.
This is illustrated by the second and fourth column.
The generation maps of the binary masks only highlight the areas that contradict the property of highly probable results.

To further check the similarity between generated and reference data, we consider the distributions for area and diameter within the berry masks shown in \figref{fig:Diameter_Area_RefVsLA}. 
The distributions of the metrics are highly similar between generated result end reference. 
For both metrics, there is a slight tendency towards an increase in area and diameter for the generated berries. 
This means that if the area of the total \texttt{berry} pixel per patch remains the same, there is a possibility that too few berries are predicted.

% % %%%%%%%%%%%%%%%%%%%%%%%%%%%%%%%%%%%%%%%%%%%%%%%%%%%%%%%%%%%%%%%%%%%%%%%%%%%%%%%%%%%%%%%%%%%%%%%%%%%%%%%%%%%%%%%%%%%%

\subsection*{Experiment 4 -- Counting in the non-occluded domain}
\label{sec:Experiment4}

Since the number of berries is of high importance for yield estimation, we investigate the estimation of this number in this experiment. 
We compare the counts based on the input patches in the occluded domain and the target patches in the non-occluded domain with the generated results of our approach.

\begin{figure}[t]
\captionsetup[subfigure]{justification=centering}
	\centering
    \subfloat[Counting in the occluded domain of SMPH.][Counting in the occluded \\domain of SMPH.]{
    \includegraphics[
            width=0.42\textwidth]{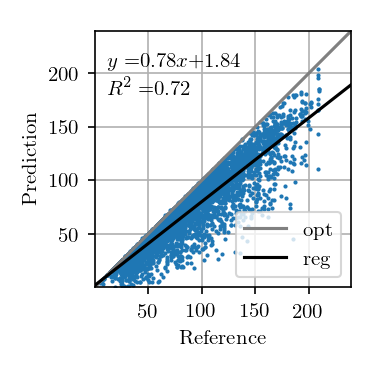}
      \label{fig:R2_Exp4_xOcc_c}}
      \hspace{3pt}
      \subfloat[Counting in non-occluded \\domain of SMPH after applying our approach.]{
    \includegraphics[
            width=0.42\textwidth]{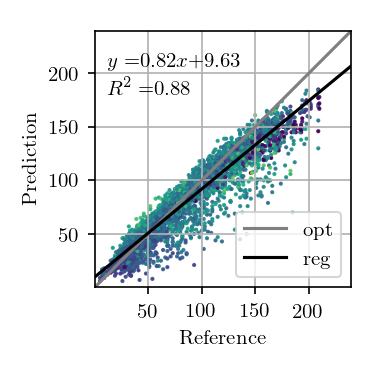}
      \label{fig:R2_Exp4_xNon_d}}
    \includegraphics[trim=182 15 5 5, clip,
            width=0.085\textwidth]{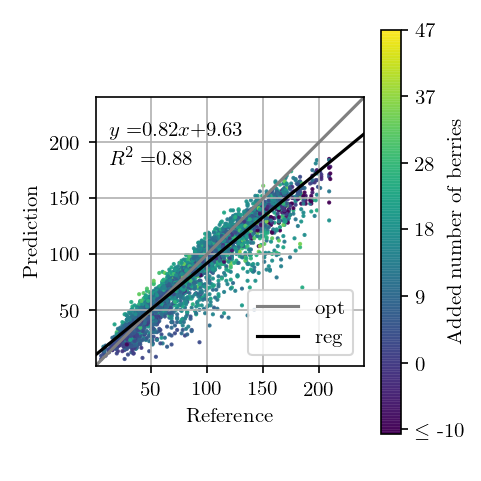}

	\caption{Counting of berries in (a) the input of the occluded domain and of generated berries in (b) the non-occluded domain of SMPH defoliated \emph{Dataset 3} visualized by a $R^2$-Plot. Shown is the relation between input $\d{x}_{\text{occ}}$ bzw. generated output $\widetilde{\d{x}}_{\text{non}}$ in relation to reference $\d{x}_{\text{non}}$. The reference is represented by the grey line. The black line represents the regression line adapted to the predictions with the corresponding regression equation at the top left of the plot. The colouration of the data points in the plots (b) and (c) indicates the added number of berries compared to the non-occluded domain.
	}   
	\label{fig:R2_Exp4}
\end{figure}

\begin{figure}[t]
	\centering
	\subfloat[VSP]{
    \includegraphics[
            width=0.49\textwidth]{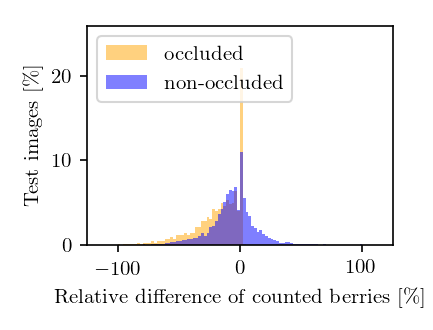}
            }
    \subfloat[SMPH]{
    \includegraphics[
            width=0.49\textwidth]{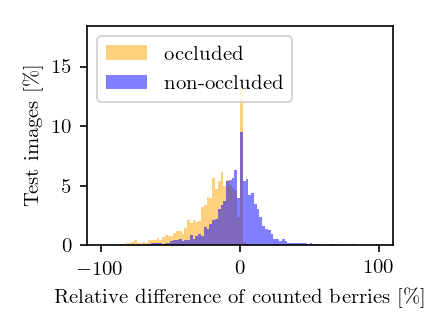}
            }

	\caption{Counting in the occluded domain (orange) and after applying our approach in the non-occluded domain (blue) relative to the reference counting in the non-occluded domain. The plots illustrate the results for (a) VSP defoliation and (b) SMPH defoliation. A negative value means that fewer berries are counted than in the reference and vice versa. Each bar corresponds to a width of 2\%.}
	\label{fig:relativeDifference_Exp4}
\end{figure}

\paragraph{Used  data,  model  and  evaluation  metrics.}
For this experiment, we use the synthetic datasets \emph{Dataset 1} and \emph{Dataset 3} based on VSP and SMPH defoliation. 
Our model is trained on both training sets and evaluated on the corresponding test sets. 
During testing, we consider only the mask of the data patches.  
For the evaluation, we use the $R^2$-Plot to plot the absolute count of the input (\figref{fig:R2_Exp4_xOcc_a}, \figref{fig:R2_Exp4_xOcc_c}) and the absolute count of the generated output of our method (\figref{fig:R2_Exp4_xNon_b}, \figref{fig:R2_Exp4_xNon_d}) with the reference count from the target mask, respectively. 
Furthermore, we examine the distribution of the relative deviations from the reference (see \figref{fig:relativeDifference_Exp4}).

%ADD

\paragraph{Results}
Counting in the occluded domain presented in \figref{fig:R2_Exp4_xOcc_a} and \figref{fig:R2_Exp4_xOcc_c} shows that there is an underestimation of the number of berries compared to the reference.
This can be explained by the occlusion, which covers a part of the berries. 
Our model shows a shift of the number of berries towards the reference for both types of defoliation. 
In both cases, the $R^2$ value increases compared to the $R^2$ value of the occluded domain, which corresponds to a better approximation of the data compared to the reference. 
It is important to mention that not only the sample distribution shifts but also compresses and concentrates along the reference line. 
Our approach thus shows that a better result is obtained than by applying only a factor to the counting. \figref{fig:relativeDifference_Exp4} supports this statement. 
The plots show the relative difference of the counted berries in the occluded domain and our method in the non-occluded domain compared to the reference counting. 
Our method (blue) depicts a normal distribution with a mean near zero. 
If the values of the occluded distribution (orange) were increased by a factor, this would lead to a shift in the distribution, but it would still be more stretched than ours. 
The peaks at value 0 correspond mostly to synthetic images where the synthetic leaf does not cover any berries. 
This is the case, for example, with images that show few berries. 

Both models exhibit problems in the generation of patches that map more than 150 berries. 
This is the case for VSP (\figref{fig:R2_Exp4_xNon_b}) and SMPH (\figref{fig:R2_Exp4_xNon_d}). 
For both types of defoliation, a trend is nevertheless evident above the critical value of 150 berries. 
Even though an underestimation of berries tends to be counted above this value, the count fits the reference better than the count in the occluded domain.
We explain this by the fact that the proportion of training images with a count above the critical value is relatively small in contrast to the number of images with an amount below the critical value.

In the occluded domain, there are data points that differ strongly from the reference. 
Our method reduces the amount of such points and also reduces the deviation of the highly deviating points.

%ADD

% %%%%%%%%%%%%%%%%%%%%%%%%%%%%%%%%%%%%%%%%%%%%%%%%%%%%%%%%%%%%%%%%%%%%%%%%%%%%%%%%%%%%%%%%%%%%%%%%%%%%%%%%%%%%%%%%%%%%

\subsection*{Experiment 5 -- Application to natural data}
\label{sec:Experiment5}

One of the contributions of our work is to investigate the applicability of our approach to natural data. 
In detail, we evaluate whether our model generalizes to natural images when it is trained on synthetic data.

\paragraph{Used  data, model  and  evaluation  metrics.}
We use the synthetic datasets \emph{Dataset 1} and \emph{Dataset 3} to train our model. 
For the test phase, we use the natural datasets \emph{Dataset 4} and \emph{Dataset 5}. 
One dataset each for VSP defoliation and one for SMPH defoliation. 

The differences of the natural dataset to the synthetic dataset are the stronger coverage by a denser leaf canopy, the resulting deviating exposure ratios, and the lower contrast whereby the contours of the leaves are not easily distinguishable from berries. 
Other differences are found in the transformation applied to the natural dataset, since non-occluded areas are not identical in both domains, as already pointed out in the introduction.
Depending on the patch position in the non-occluded domain in the original image, the transformation goes beyond the boundaries of the original image in the occluded domain. 
To achieve a patch size of $\SI{656}{\px} \times \SI{656}{\px}$ which is equivalent to the cropped patch size of the dataset, the appropriate borders of the patch are filled with black pixels.

We perform our evaluation visually, which means we compare the input from the occluded domain with the generated output of our approach in the non-occluded domain. 
Due to the transformation issues, direct numerical comparison and evaluation between target and generated output are not useful for the majority of patches. 
However, we would like to give an impression of the results by means of the visual representation.

%ADD
\begin{figure*}[t]
\renewcommand{\tabcolsep}{5pt}
\centering
    
	\subfloat[Example 1 VSP]{
        \begin{minipage}{0.30\textwidth}
            \begin{minipage}{0.31\textwidth}
                \centering
        		$\d{x}_{\text{occ}}$
    		\end{minipage}
    		\begin{minipage}{0.31\textwidth}
        		\centering
        		 $\d{x}_{\text{non}}$
    		\end{minipage}
            \begin{minipage}{0.31\textwidth}
                \centering
                $\widetilde{\d{x}}_{\text{non}}$
    		\end{minipage}
    		
			\includegraphics[width=0.31\textwidth]{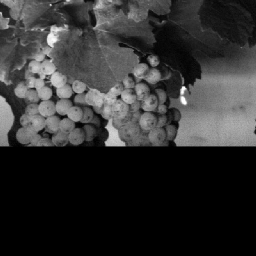}
			\includegraphics[width=0.31\textwidth]{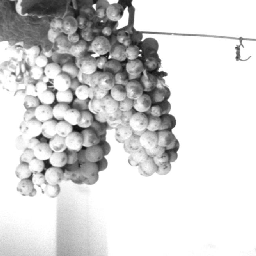}
			\includegraphics[width=0.31\textwidth]{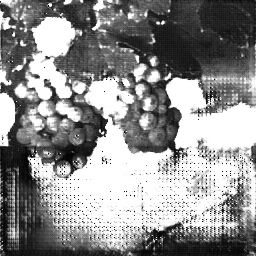}
			
			\includegraphics[width=0.31\textwidth]{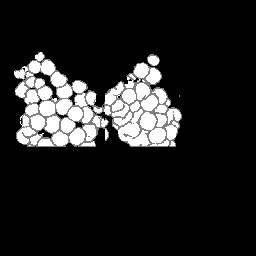}
			\includegraphics[width=0.31\textwidth]{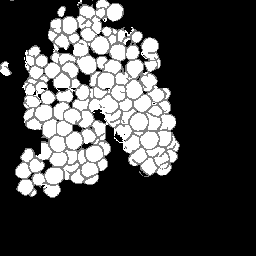}
			\includegraphics[width=0.31\textwidth]{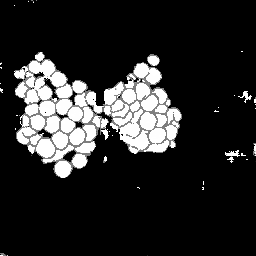}

    	\end{minipage}
    \label{fig:Exp5_ex1_good}}
    \hspace{3pt}
	\subfloat[Example 2 VSP]{
    	 \begin{minipage}{0.30\textwidth}
            \begin{minipage}{0.31\textwidth}
                \centering
        		$\d{x}_{\text{occ}}$
    		\end{minipage}
    		\begin{minipage}{0.31\textwidth}
        		\centering
        		 $\d{x}_{\text{non}}$
    		\end{minipage}
            \begin{minipage}{0.31\textwidth}
                \centering
                $\widetilde{\d{x}}_{\text{non}}$
		    \end{minipage}
		
			\includegraphics[width=0.31\textwidth]{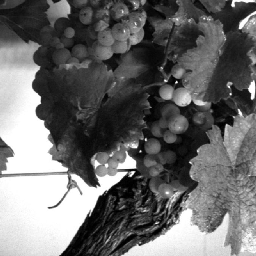}
		    \includegraphics[width=0.31\textwidth]{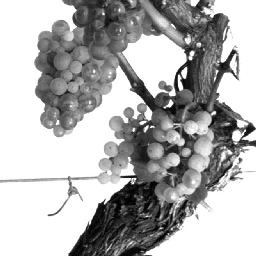}
		    \includegraphics[width=0.31\textwidth]{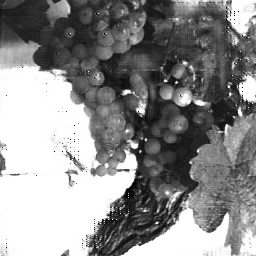}
			
			\includegraphics[width=0.31\textwidth]{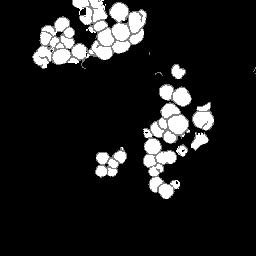}
		    \includegraphics[width=0.31\textwidth]{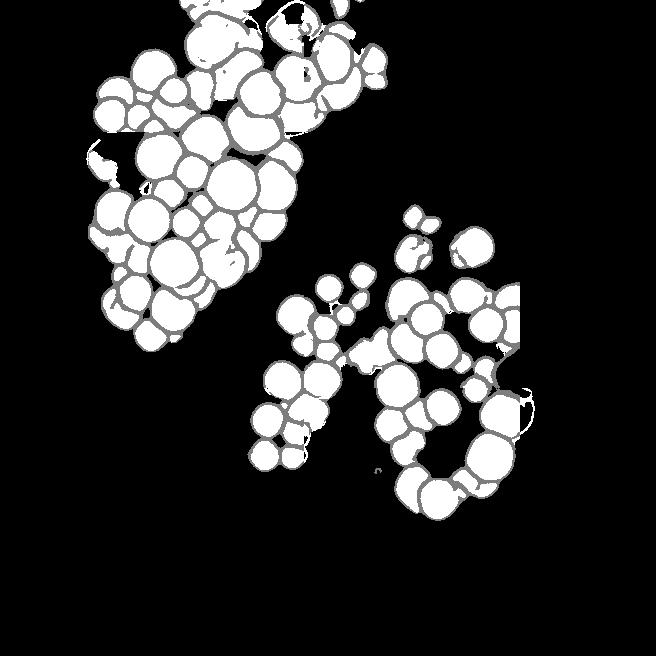}
		    \includegraphics[width=0.31\textwidth]{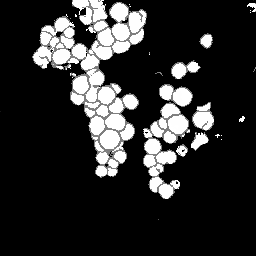}
    	\end{minipage}
    	\label{fig:Exp5_ex2_good}}
    \hspace{3pt}
    \subfloat[Example 3 VSP]{
    	 \begin{minipage}{0.30\textwidth}
            \begin{minipage}{0.31\textwidth}
                \centering
        		$\d{x}_{\text{occ}}$
    		\end{minipage}
    		\begin{minipage}{0.31\textwidth}
        		\centering
        		 $\d{x}_{\text{non}}$
    		\end{minipage}
            \begin{minipage}{0.31\textwidth}
                \centering
                $\widetilde{\d{x}}_{\text{non}}$
		    \end{minipage}
		
		    \includegraphics[width=0.31\textwidth]{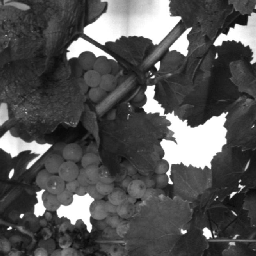}
		   \includegraphics[width=0.31\textwidth]{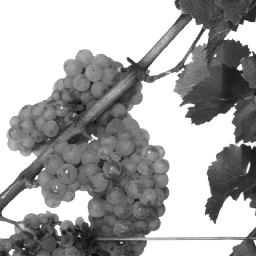}
		   \includegraphics[width=0.31\textwidth]{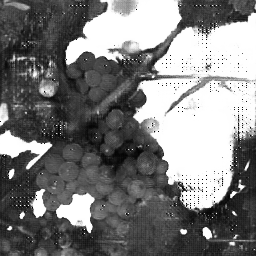}
			
		   \includegraphics[width=0.31\textwidth]{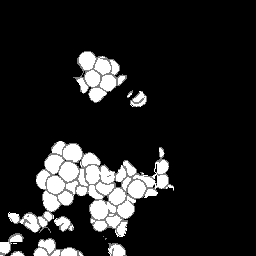}
		   \includegraphics[width=0.31\textwidth]{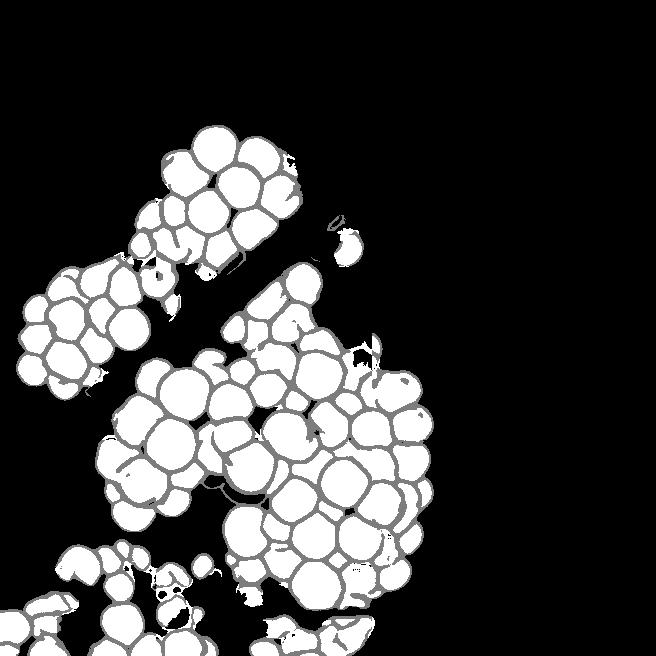}
		   \includegraphics[width=0.31\textwidth]{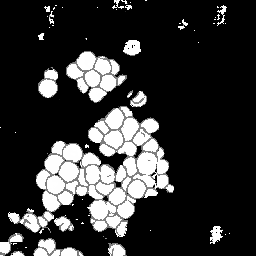}
    	\end{minipage}
    \label{fig:Exp5_ex3_good}}

    \caption{Visual representation of generated result based on natural data input. Examples with a good transformation between the patches. They are present as a minority in the natural dataset.}
    \label{fig:visualComparison_realData_good}
\end{figure*}

\begin{figure*}[t]
\renewcommand{\tabcolsep}{5pt}
\centering
 \subfloat[Example 1 (VSP)]{
     \begin{minipage}{0.30\textwidth}
                \begin{minipage}{0.31\textwidth}
                    \centering
            		$\d{x}_{\text{occ}}$
        		\end{minipage}
        		\begin{minipage}{0.31\textwidth}
            		\centering
            		 $\d{x}_{\text{non}}$
        		\end{minipage}
                \begin{minipage}{0.31\textwidth}
                    \centering
                    $\widetilde{\d{x}}_{\text{non}}$
        		\end{minipage}
        		
    			\includegraphics[width=0.31\textwidth]{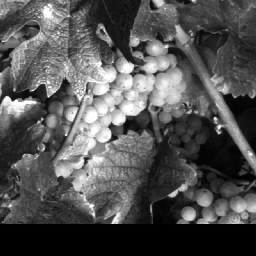}
    			\includegraphics[width=0.31\textwidth]{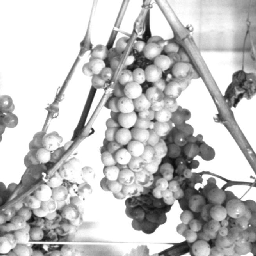}
    			\includegraphics[width=0.31\textwidth]{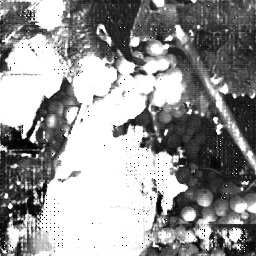}
    			
    			\includegraphics[width=0.31\textwidth]{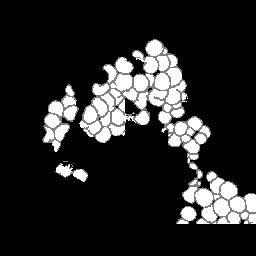}
    			\includegraphics[width=0.31\textwidth]{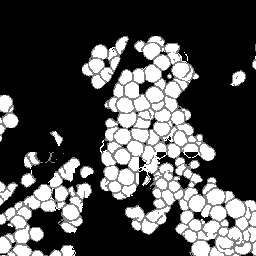}
    			\includegraphics[width=0.31\textwidth]{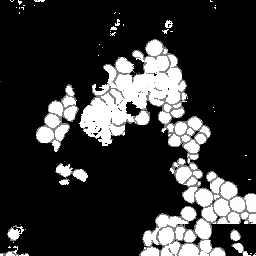}
    
    \end{minipage}
    \label{fig:Exp5_ex1_transform}}
\hspace{3pt}
    \subfloat[Example 2 (SMPH)]{
    	 \begin{minipage}{0.30\textwidth}
            \begin{minipage}{0.31\textwidth}
                \centering
        		$\d{x}_{\text{occ}}$
    		\end{minipage}
    		\begin{minipage}{0.31\textwidth}
        		\centering
        		 $\d{x}_{\text{non}}$
    		\end{minipage}
            \begin{minipage}{0.31\textwidth}
                \centering
                $\widetilde{\d{x}}_{\text{non}}$
		    \end{minipage}
		
			\includegraphics[width=0.31\textwidth]{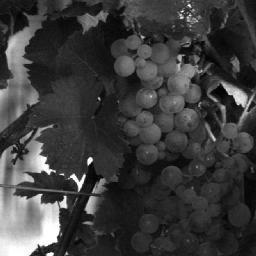}
		   \includegraphics[width=0.31\textwidth]{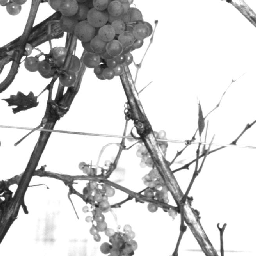}
		   \includegraphics[width=0.31\textwidth]{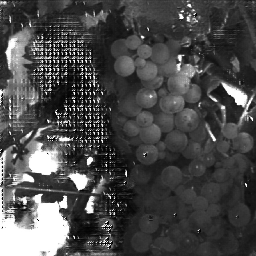}
			
			\includegraphics[width=0.31\textwidth]{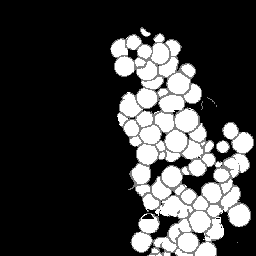}
		   \includegraphics[width=0.31\textwidth]{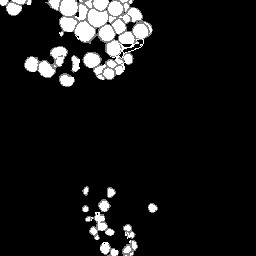}
		   \includegraphics[width=0.31\textwidth]{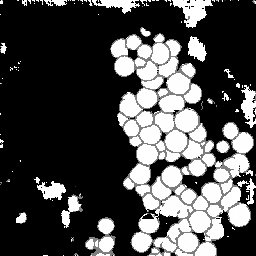}
    	\end{minipage}
    	\label{fig:Exp5_ex2_transform}}
\hspace{3pt}
    \subfloat[Example 3 (VSP)]{
        \begin{minipage}{0.30\textwidth}
            \begin{minipage}{0.31\textwidth}
                \centering
        		$\d{x}_{\text{occ}}$
    		\end{minipage}
    		\begin{minipage}{0.31\textwidth}
        		\centering
        		 $\d{x}_{\text{non}}$
    		\end{minipage}
            \begin{minipage}{0.31\textwidth}
                \centering
                $\widetilde{\d{x}}_{\text{non}}$
    		\end{minipage}
    		
			\includegraphics[width=0.31\textwidth]{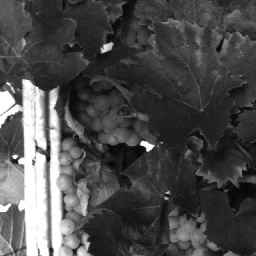}
			\includegraphics[width=0.31\textwidth]{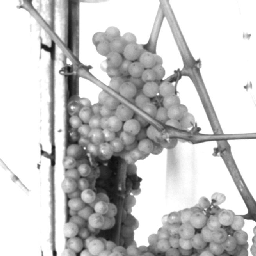}
			\includegraphics[width=0.31\textwidth]{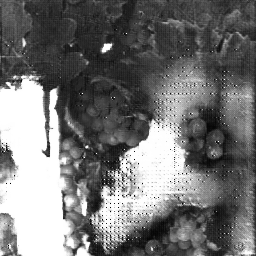}
			
			\includegraphics[width=0.31\textwidth]{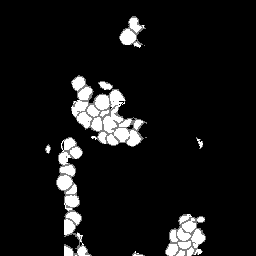}
			\includegraphics[width=0.31\textwidth]{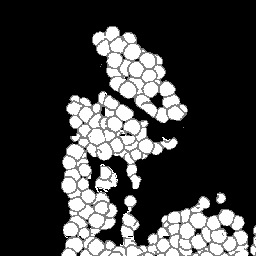}
			\includegraphics[width=0.31\textwidth]{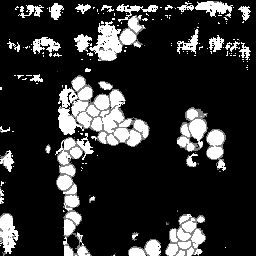}
    \end{minipage}
    \label{fig:Exp5_ex3_artifact}}
    \caption{Visual representation of generated result based on natural data input. (a) and (b) show examples with a insufficient transformation between the patches. (c) shows an example with artifacts in the generated mask.}
    \label{fig:visualComparison_realData_insufficientTrafo_artifacts}
\end{figure*}

\paragraph{Results}
In \figref{fig:visualComparison_realData_good} and \figref{fig:visualComparison_realData_insufficientTrafo_artifacts}, we provide example results of our approach applied to natural data. 
For each example, the first column shows the input $\d{x}_{\text{occ}}$ of the occluded domain, the second column the reference $\d{x}_{\text{non}}$ in the non-occluded domain and the last column our generated output $\widetilde{\d{x}}_{\text{non}}$ in the non-occluded domain. 
The first row visualizes the L channel of a patch and the second row the corresponding mask.

The results show that the canopy is reduced and important areas in the patch are reproduced. 
Generally, our findings from the previously described experiments can be confirmed.
Using our generative approach, \texttt{berry} and \texttt{berry-edge} pixel regions in the input mask are also transferred to the generated output for the natural data. 
For input patches of the occluded domain being similar to the synthetic data (\figref{fig:visualComparison_realData_good}), the results show an expansion of the existing berry region. 
Our approach is also able to deal with transformation problems, as in \figref{fig:Exp5_ex1_good} where the transformation goes beyond the original image boundaries. 
There are examples like seen in \figref{fig:Exp5_ex3_good}, that look similar to the target, or examples that look real compared to the input but do not reflect the target output (\figref{fig:Exp5_ex2_good}).

For the majority of natural data, exact transformations are not available, so this is challenging to evaluate.
In examples like the one in \figref{fig:Exp5_ex1_transform} transformation, rotation and scale fit, but due to defoliation, the orientation of the grape bunch is different in input and output target. 
In the input, the grape bunch is more horizontal. In the target, it is vertical.
Example 2 (\figref{fig:Exp5_ex2_transform}) shows that grape bunches are also completely different in translation due to the different weights attached to the branches.
In this example, the grape bunch that is visible in the input is only partially visible at the top of the patch in the target output. 
The generated output adapts to the input and is also expanded, but is not comparable to the target.

%ADD

Although it is apparent that the model trained only on the synthetic data mentioned above is not yet strong enough to obtain similarly good results for the more complex natural data as for the synthetic data, we consider the results promising. 
We assume that mixing natural and synthetic data or using more complex synthetic training data can improve the results.
We observe checkerboard artifacts that appear in the generated L patches (see \figref{fig:Exp5_ex1_good} and \figref{fig:Exp5_ex3_good}), which could be reduced by improving the generator. 
This could also result in reduced artifacts as they occur in the mask in \figref{fig:Exp5_ex3_artifact}. 
The artifacts occur more in patches that present a dense canopy.

% %%%%%%%%%%%%%%%%%%%%%%%%%%%%%%%%%%%%%%%%%%%%%%%%%%%%%%%%%%%%%%%%%%%%%%%%%%%%%%%%%%%%%%%%%%%%%%%%%%%%%%%%%%%%%%%%%%%%
% %%%%%%%%%%%%%%%%%%%%%%%%%%%%%%%%%%%%%%%%%%%%%%%%%%%%%%%%%%%%%%%%%%%%%%%%%%%%%%%%%%%%%%%%%%%%%%%%%%%%%%%%%%%%%%%%%%%%

\section*{Conclusion and Future Directions}
\label{sec:conclusion}

In this work, we have demonstrated the suitability of a conditional generative adversarial network like Pix2Pix to generate a scenario behind occlusions in grapevine images that is highly probable based on visible information in the images.
Our experiments have shown that our approach has learned patterns that characterize typical berries and clusters without occlusions so that areas where berries are added and other areas where the image remains unchanged can be identified without having to provide prior knowledge about occlusions.
Compared to counting with occluded areas, we show that our approach provides a count that is closer to the manual reference count.
In contrast to applying a factor, our approach directly involves the appearance of the visible berries and thus better adapts to local conditions.

We have trained our conditional adversarial network-based model on synthetic data only in order to overcome the challenge of lacking aligned image pairs.  
We show that the model is also applicable to natural data given that the canopy is not too dense and the variation between natural data and synthetic data is not too high.
To make the model more robust and generalizable to variations between natural and synthetic data, the synthetic data can be designed with more complex changes, for example, by increasing the synthetic occlusion through the use of more leaves. 
In addition, brightness and contrast could be varied, for example, to reduce the dominant white background of the synthetic data and thus make it more difficult for the model to detect the occlusion.
Another promising future direction is to train the model from a combination of synthetic images and a limited amount of natural images. 
In this case, the transformation between the two required domains needs to be accurate enough and suitable data must be selected. Another possibility would involve extensive manual work on the transformation between the domains or more sophisticated techniques such as image warping.
In the future, the checkerboard artifacts that occur in data could be reduced by replacing the transpose convolution layer of the decoder in the U-Net generator with bi-linear up-sampling operations, as described in \cite{odena2016deconvolution}.

% %%%%%%%%%%%%%%%%%%%%%%%%%%%%%%%%%%%%%%%%%%%%%%%%%%%%%%%%%%%%%%%%%%%%%%%%%%%%%%%%%%%%%%%%%%%%%%%%%%%%%%%%%%%%%%%%%%%%
% %%%%%%%%%%%%%%%%%%%%%%%%%%%%%%%%%%%%%%%%%%%%%%%%%%%%%%%%%%%%%%%%%%%%%%%%%%%%%%%%%%%%%%%%%%%%%%%%%%%%%%%%%%%%%%%%%%%%

\section*{ACKNOWLEDGEMENTS}\label{ACKNOWLEDGEMENTS}
% OPTIKO
This project was funded by the European Agriculture Fund for Rural Development with contribution from North-Rhine Westphalia (17-02.12.01 - 10/16 – EP-0004617925-19-001).
% DAS IST VERPFLICHTEND FUER PHENOROP DOKTORANDEN
Furthermore, this work was partially funded by the Deutsche Forschungsgemeinschaft (DFG, German Research Foundation) under Germany’s Excellence Strategy – EXC 2070 – 390732324, and
% NOVISYS
partially by the German Federal Ministry of Education and Research (BMBF, Bonn, Germany) in the framework of the project novisys (FKZ 031A349).

% \section*{References}

% %%%%%%%%%%%%%%%%%%%%%%%%%%%%%%%%%%%%%%%%%%%%%%%%%%%%%%%%%%%%%%%%%%%%%%%%%%%%%%%%%%%%%%%%%%%%%%%%%%%%%%%%%%%%%%%%%%%%

\bibliographystyle{spmpsci}      % mathematics and physical sciences
\bibliography{References.bib}   % name your BibTeX data base

% \begin{thebibliography}{}
% %
% % and use \bibitem to create references. Consult the Instructions
% % for authors for reference list style.
% % Format for Journal Reference
% Author, Article title, Journal, Volume, page numbers (year)
% Author, Book title, page numbers. Publisher, place (year)
% \end{thebibliography}

\end{document}